\documentclass[a4paper,fleqn]{cas-dc}

\usepackage[numbers]{natbib}
\usepackage{soul}
\usepackage{bm}
\soulregister\cite7 
\soulregister\ref7 
\usepackage{hyperref}

\begin{document}
\let\WriteBookmarks\relax
\let\printorcid\relax
\def\floatpagepagefraction{1}
\def\textpagefraction{.001}

\shorttitle{Dynamic Spatiotemporal Network for Video Salient Object Detection}    

\shortauthors{J. Liu et al.}  

\title [mode = title]{DS-Net: Dynamic Spatiotemporal Network for Video Salient Object Detection}  

\tnotemark[1] 

\tnotetext[1]{This work is supported by Key Laboratory of Artificial Intelligence, Ministry of Education, under grant AI2020006, and National Science Foundation of China under Grant 61701341.} 

%

\author[1,2]{Jing Liu}
\cormark[1]
\ead{jliu_tju@tju.edu.cn}
\author[1]{Jiaxiang Wang}
\author[1]{Weikang Wang}
\author[1]{Yuting Su}







\affiliation[1]{organization={The School of Electrical and Information Engineering, Tianjin University},
            city={Tianjin},
            postcode={300072}, 
            country={China}}






\affiliation[2]{organization={Key Laboratory of Artificial Intelligence, Ministry of Education},
            city={Shanghai},
            postcode={200240}, 
            country={China}}

\cortext[1]{Corresponding author.}



\begin{abstract}
As moving objects always draw more attention of human eyes, the temporal motion information is always exploited complementarily with spatial information to detect salient objects in videos. Although efficient tools such as optical flow have been proposed to extract temporal motion information, it often encounters difficulties when used for saliency detection due to the movement of camera or the partial movement of salient objects. In this paper, we investigate the complimentary roles of spatial and temporal information and propose a novel \textit{dynamic spatiotemporal network} (DS-Net) for more effective fusion of spatiotemporal information. We construct a symmetric two-bypass network to explicitly extract spatial and temporal features. A \textit{dynamic weight generator} (DWG) is designed to automatically learn the reliability of corresponding saliency branch. And a top-down \textit{cross attentive aggregation} (CAA) procedure is designed so as to facilitate dynamic complementary aggregation of spatiotemporal features. Finally, the features are modified by spatial attention with the guidance of coarse saliency map and then go through decoder part for final saliency map. Experimental results on five benchmarks VOS, DAVIS, FBMS, SegTrack-v2, and ViSal demonstrate that the proposed method achieves superior performance than state-of-the-art algorithms. The source code is available at \textit{\href{https://github.com/TJUMMG/DS-Net}{https://github.com/TJUMMG/DS-Net.}}
\end{abstract}



\begin{keywords}
\sep Video salient object detection 
\sep Convolutional neural network 
\sep Spatiotemporal feature fusion 
\sep Dynamic aggregation
\end{keywords}

\maketitle

\section{Introduction}
\label{Introduction}

Video Salient Object Detection (VSOD) aims at discovering the most visually distinctive objects in a video. The task of video salient object detection is not only to locate the salient region, but also to focus on the segmentation of salient object, that is, to separate the salient object pixels from the background pixels. Therefore, the results of video salient object detection can be applied to a variety of subsequent computer vision tasks, such as person re-identification~\cite{6619304}, visual tracking~\cite{8354237} and video compression~\cite{1331443}.
\begin{figure}[t]
\centering
\includegraphics[width=1\columnwidth]{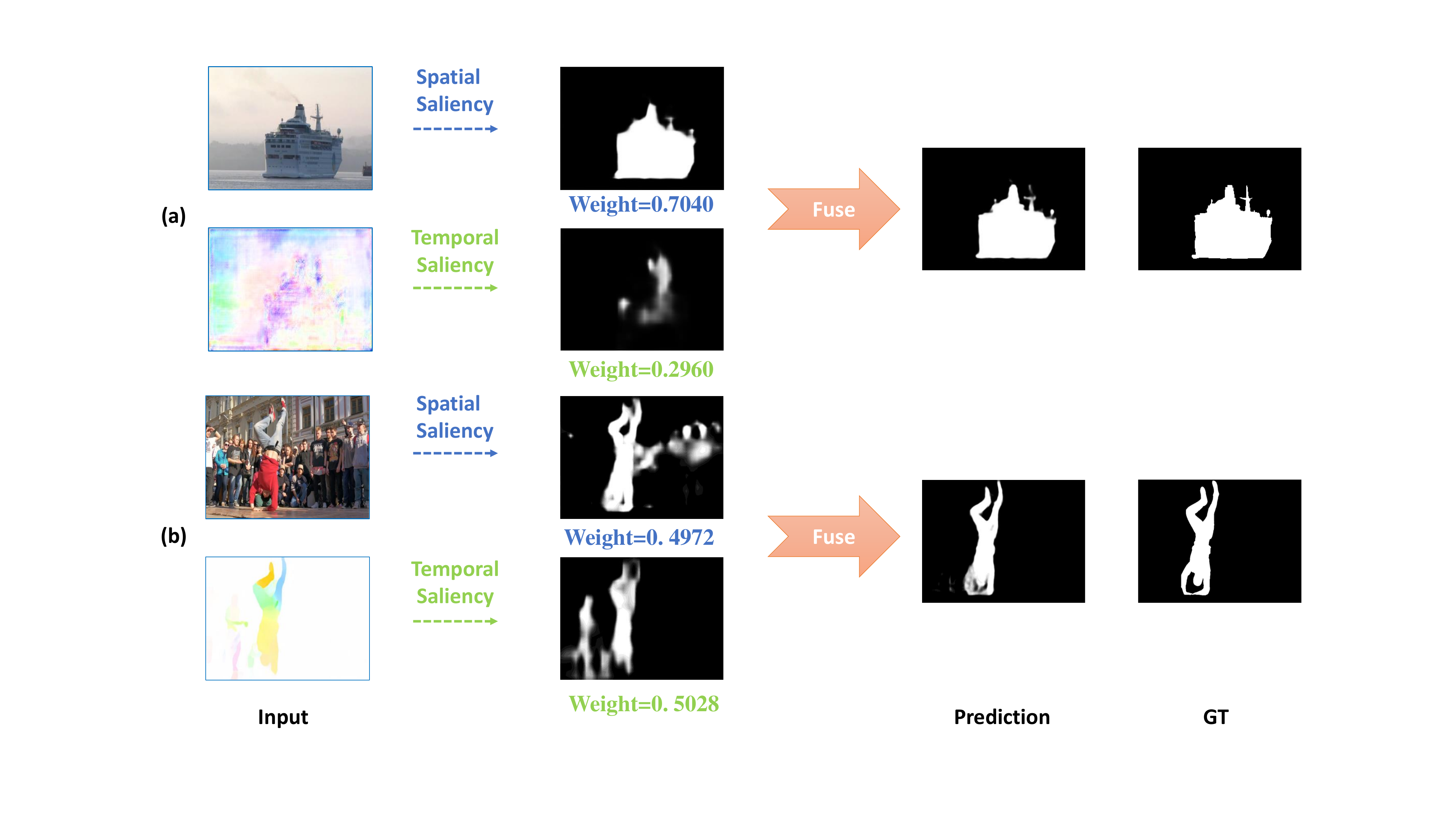} 
\caption{Illustration of the proposed algorithm for two common yet challenging cases. (a) Noisy optical flow with high spatial contrast. (b) Reliable optical flow with complex spatial background. The DS-Net is capable of efficiently aggregating reliable spatial and temporal information.}
\label{example}
\end{figure}

Compared with image salient object detection, video salient object detection is much more challenging because it needs to consider the time continuity between video frames, camera shake, object movement and other factors. In recent years, with the development of convolution neural network (CNN), a variety of learning based salient object detection networks have been proposed~\cite{inbook, 8578440, 8954050, 8047320, LIU2022103425}, which have greatly improved the performance of the algorithms in this field. In video salient object detection task, the motion information between video frames plays a significant role, since the researches have shown that human eyes pay more attention to moving objects in a video~\cite{730558}. So it is essential to extract motion information in video sequences. At present, a large number of algorithms extract the optical flow~\cite{9010060, 8578440}, which reflects the motion of pixels between two video frames. And the motion information contained in optical flow is often used to guide static saliency for better saliency map. However, in cases where the camera moves, the movement of the salient object is too small, or only part of the salient object is moving, the motion information contained in the optical flow is less correlated with salient objects. Therefore, indiscriminately aggregating the less reliable motion information with spatial information can hardly improve but may even ruin the saliency results.

To address the dynamic reliability of static and motion saliency cues, dynamic aggregation of spatial and temporal information according to their reliability is desired. In this paper, we propose a dynamic spatiotemporal network (DS-Net). The proposed network is composed of four components: symmetric spatial and temporal saliency network, dynamic weight generation, top-down cross attentive aggregation, and final saliency prediction. Symmetric spatial and temporal saliency network explicitly extract saliency-relevant features in static frame and optical flow, respectively.
Given the multi-scale spatial and temporal features, a novel \textit{dynamic weight generator} (DWG) is designed to learn the dynamic weight vectors representing the reliability of corresponding features, which are then used for top-down \textit{cross attentive aggregation} (CAA) for each feature. The spatial and temporal saliency maps are also adaptively fused to obtain a coarse saliency map based on the dynamic weight vector. Finally, the coarse saliency map is applied to the progressively dynamically aggregated features via spatial attention to further eliminate background noises and then outputs the final saliency map. 

As shown in Fig.~\ref{example}, the temporal information captured by optical flow may varies in its reliability for salient object detection. When it fails to reflect the motion of foreground objects, as shown in (a), the proposed DS-Net can adjust the aggregation process so that the spatial branch plays a dominant role. On the contrary, when video frame contains complex background noises while the temporal information is more highly correlated with the foreground object, as shown in (b), the proposed DS-Net performs equally well by emphasizing on the temporal branch via cross attentive aggregation. Experimental results on five large-scale video salient object detection datasets validate the effective of the proposed algorithm. In short, the contributions of this paper are summarized as follows:

\begin{enumerate}[1)]
	\item To tackle the dilemma of dynamic reliability of static and motion saliency cues, we propose a dynamic spatiotemporal network (DS-Net) which automatically learns the reliability of spatial and temporal branch and complementarily aggregates the corresponding features in a cross attentive way.
    \item A dynamic weight generator is designed to fuse multi-scale features and explicitly learn the weight vector representing the reliability of input features, which can be used for aggregating multi-scale spatiotemporal features as well as coarse saliency maps.
	\item A top-down cross attentive aggregation procedure with non-linear cross thresholding is designed to progressively fuse the complementary spatial and temporal features according to their reliability.
\end{enumerate}

The rest of this paper is organized as follows. Section~\ref{relatedwork} reviews existing video salient object detection algorithms and spatiotemporal fusion methods. Section~\ref{algorithm} describes the proposed algorithm thoroughly. Experiment setting as well as evaluation results are discussed in Section~\ref{experiment}. Finally, conclusive remarks are made in Section~\ref{conclusion}.

\section{Related Work}
\label{relatedwork}
\subsection{Video Salient Object Detection}
\label{relatedwork:VSOD}

Image salient object detection has been explored for decades and achieved excellent performance~\cite{9008371, 8798692, 8768013, ZHOU2021102975, LI2022103461, ZHOU2020102818}. Compared with image salient object detection, video salient object detection task is more challenging due to the motion of objects.

Traditional video salient object detection methods usually employed handcrafted features to locate the salient regions~\cite{8361895, 8705313, 8832185}. In recent years, the rapidly developing convolution neural network greatly improves the performance of video salient object detection algorithms. Wang \emph{et al.}~\cite{8047320} firstly applied the fully convolution neural (FCN) network to video salient object detection and achieved remarkable performance. After that, more and more deep learning based methods have been proposed. Li \emph{et al.}~\cite{8578440} extended the FCN-based image saliency model by the flow guided feature warp and recurrent feature refinement module. Afterwards, to get better spatiotemporal feature fusion performance, Song \emph{et al.}~\cite{inbook} directly input sequences into two parallel dilated bi-directional ConvLSTMs to learn spatiotemporal information and obtained significant performance improvement. Following previous works, Fan \emph{et al.}~\cite{8954050} used parallel dilated convolution layers to exploit multi-scale information and equips convLSTM with a saliency-shift-aware attention mechanism. Some works also investigate the guidance role of motion features to static features. Li \emph{et al.}~\cite{9010060} proposed a novel motion guided network taking single static frame and optical flow as input, which can achieve excellent performance. Recently, some works started to focus on designing lightweight VSOD networks. Gu \emph{et al.}~\cite{DBLP:conf/aaai/GuWW0CL20} proposed a novel self-attention mechanism based on non-local block and utilized pyramid structure to get motion cues in different scales and different speed, which achieved state-of-art performance as well as the highest fps among all other learning based VSOD algorithms.

\subsection{Fusion of Spatiotemporal Information}
\label{relatedwork:st-fusion}
The optimization of fusion and extraction strategy of spatiotemporal information is always of the first importance in video saliency detection. In the early days, conventional algorithms mainly relied on techniques including low-rank and sparse decomposition~\cite{6288171}, local gradient flow~\cite{7164324}, object proposals~\cite{8082546} and so on. At present, there are generally three main categories of spatiotemporal fusion strategies in learning based video saliency detection methods: direct temporal fusion, recurrent fusion, and bypass fusion.

Representative models for direct temporal fusion are~\cite{7775087,8047320}, which used FCN to extract spatial features of different frames and directly concatenate them together to fuse spatiotemporal information. This simple connection fails to effectively reflect the correlation between spatial and temporal features. For recurrent fusion, the convolutional memory units such as ConvLSTM and ConvGRU are usually used to integrate sequence information in a recurrent manner. Representative models include~\cite{8954050, DBLP:conf/eccv/JiangXLQW18} and~\cite{9008826} which employed ConvLSTM and ConvGRU respectively to aggregate long-range spatiotemporal features. Compared with direct temporal fusion, these memory units can extract the temporal continuity by inputting consecutive frames in order. However, these memory units often lead to high computational and memory cost. The last type of spatiotemporal fusion employs different bypasses to explicitly extract static and motion information and then fuse features from different bypasses together. The most commonly used method to represent temporal motion is to calculate optical flow between consecutive frames. Li \emph{et al.}~\cite{9010060} proposed a novel motion attention based on optical flow to guide the spatial feature extraction and realize the fusion of spatiotemporal features. This fusion method is simple and effective, but it is highly sensitive to the accuracy of optical flow. Noticing this drawback and being aware of the complex human visual attention mechanism, we propose to dynamically fuse spatial and temporal information according to input sequences. In this discriminate way, both spatial and temporal features can show their strength without interfering with each other.

\begin{figure*}[t]
\centering
\includegraphics[width=0.98\textwidth]{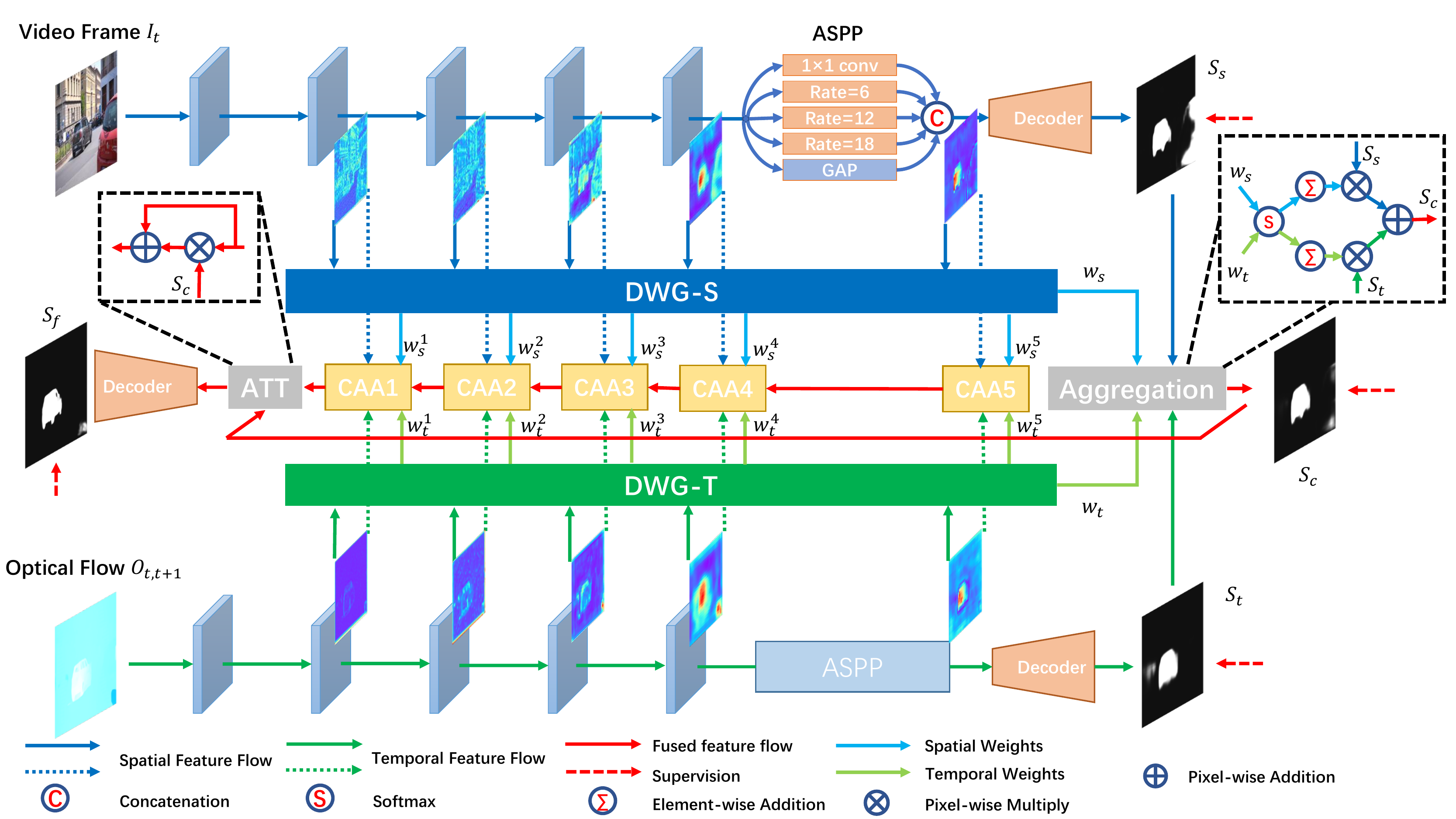} 
\caption{Overview of the proposed dynamic spatiotemporal network (DS-Net). DWG-S and DWG-T: dynamic weight generator for spatial and temporal branch. CAA: cross attentive aggregation module.}
\label{network}
\end{figure*}

\section{The Proposed Algorithm}
\label{algorithm}

The whole network structure is illustrated in Fig.~\ref{network}. We propose a dynamic spatiotemporal network which can adaptively aggregate the spatial information from static video frame and the temporal information from optical flow. Firstly, the video frame and the optical flow between the current and next frame are input into the symmetric saliency network to extract spatial and temporal motion features, respectively. After that, a dynamic weight generator is used to fuse features of different scales and generate weight vectors. The weight vectors are used to dynamically aggregate multi-scale spatial and temporal features as the final spatiotemporal features. Finally, through a top-down cross attentive feature fusion procedure, the multi-scale spatiotemporal features are progressively fused for the final saliency maps. In addition, the spatial and temporal saliency maps are also fused adaptively for a coarse saliency map based on the weight vector. And the coarse saliency map is used to suppress the background noises of spatiotemporal feature via spatial attention.

\subsection{Symmetric Spatial and Temporal Saliency Network}
\label{Symmetric network}

As shown in Fig.~\ref{network}, the DS-Net mainly includes two symmetric branches, which are used to explicitly extract static spatial features and motion temporal features respectively. Motivated by~\cite{DBLP:journals/corr/ChenPSA17}, these two branches are both composed of a ResNet34~\cite{7780459} feature extractor, an Atrous Spatial Pyramid Pooling (ASPP)~\cite{8099517} module and a decoder. We adopt the head convolution and four residual structures of classic ResNet34 as feature extractor. After that, the feature is input into the ASPP module for multi-scale information extraction. As shown in Fig.~\ref{network}, ASPP module mainly uses four convolution kernels of different receptive fields to extract multi-scale information. The convolution kernels of different receptive fields are realized by different dilation rates. Then, the four features representing multi-scale information are concatenated with the feature after global average pooling to obtain the final output feature. Finally, the feature after ASPP module is input into the decoder to generate saliency prediction map. The decoder used in this paper contains three convolution and activation layers to realize channel reduction. 

For the ease of the following descriptions, denote the current frame as $I_t$ and the next frame as $I_{t+1}$. These two images are input into FlowNet2.0~\cite{8099662} to generate the optical flow image, which is denoted as $O_{t, t+1}$. Then $I_t$ and $O_{t, t+1}$ go through the symmetric spatial and temporal saliency network to extract features and predict coarse saliency maps. The spatial features output by four residual structures of ResNet34 and ASPP module are denoted as $\mathcal{F}_s=\{f_s^i|i=1,\cdots,5\}$. Similarly, the output features of the temporal branch are denoted as $\mathcal{F}_t=\{f_t^i|i=1,\cdots,5\}$. The coarse spatial and temporal  saliency map are denoted as $S_s$ and $S_t$.

\begin{figure}[t]
\centering
\includegraphics[width=0.45\textwidth]{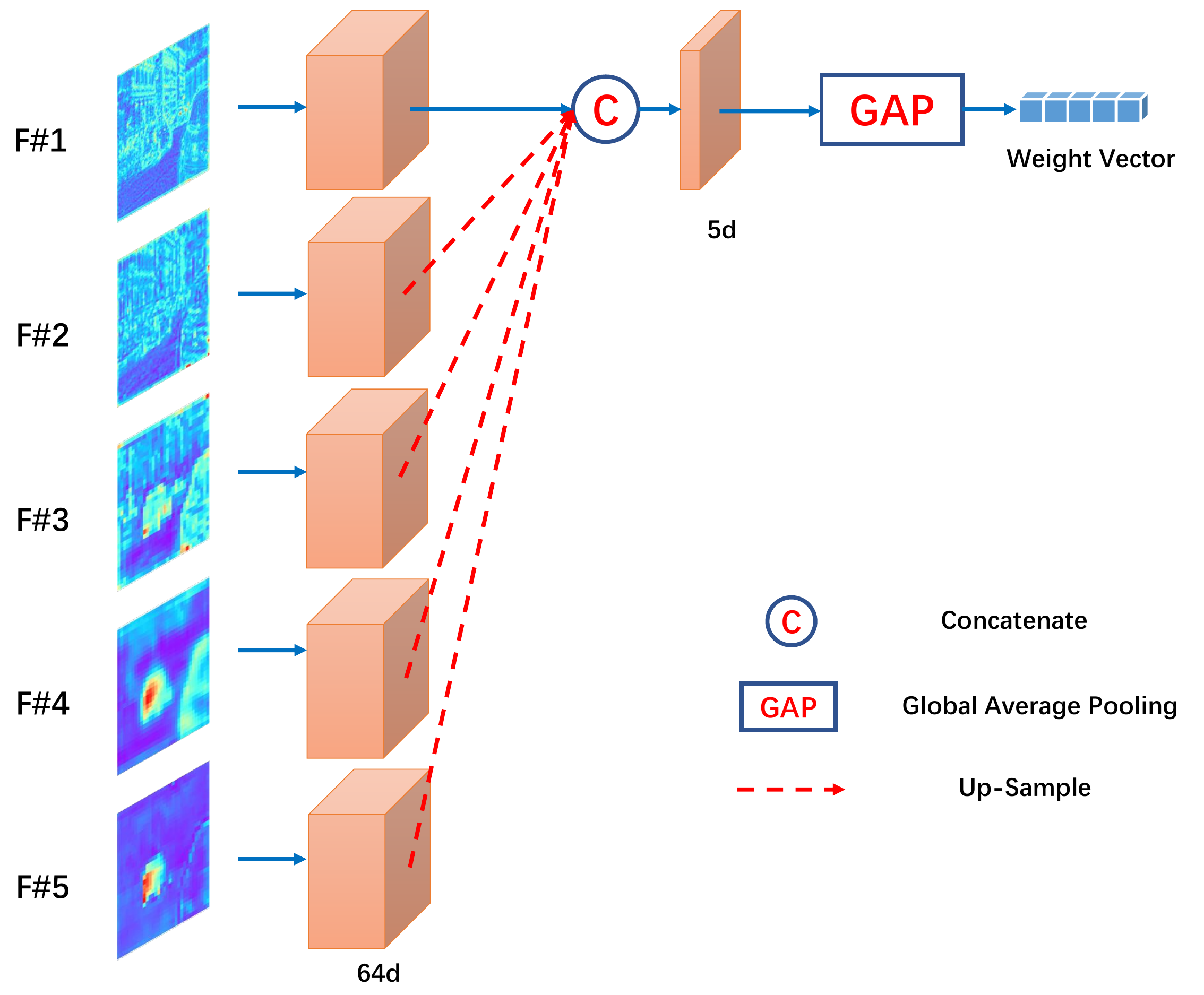}
\caption{The structure of dynamic weight generator. F\#i denote the features extracted from the i-th layer of spatial or temporal saliency branch.}
\label{DWG}
\end{figure}

\subsection{Dynamic Weight Generator}
The structure of the proposed dynamic weight generator is illustrated in Fig.~\ref{DWG}. Researches have shown that features from different layers of the network contain complementary global context and local detailed information~\cite{8315520, 8237293}. Therefore, the multi-scale features captured from spatial or temporal branch are jointly employed to generate weight vectors. In this way, the information of different scales of features are effectively fused, so that the learned weight vectors can better reflect the relative reliability of different features. 

Given the feature pyramid $\mathcal{F}_s$ or $\mathcal{F}_t$ extracted from symmetric saliency network, a convolution layer is first applied to transform all feature maps into channel of 64. Afterward, four deeper but coarser-scale features are upsampled to the finest scale and concatenated with the feature at the finest scale to get a feature with 320 channels. Finally, it goes through a channel reduction layer and global average pooling (GAP) to get a 5-d weight vector. Let $\bm{w}_s=[w_s^1,w_s^2,w_s^3,w_s^4,w_s^5]$ and $\bm{w}_t=[w_t^1,w_t^2,w_t^3,w_t^4,w_t^5]$ represent the 5-d weight vector for spatial and temporal branch respectively. As will be shown later, the 5-d weight vectors correspond to the reliability of 5 input features, which will be used for the subsequent attentive feature fusion as well as the coarse saliency maps fusion.

\subsection{Cross Attentive Aggregation}
As mentioned above, human attention is driven by both static cues and motion cues, but their contributions to salient object detection vary from sequences to sequences. Therefore, it is desired to dynamically aggregate the spatial and temporal information. Recall that the dynamic weights $w_s^i$ and $w_t^i$ are learned to stand for the relative reliability of their corresponding features $f_s^i$ and $f_t^i$. A straightforward way to get spatiotemporal feature is the weighted summation as:
\begin{equation}
\label{eqn.weightedsum}
F_{agg}^i=w_s^i\cdot f_s^i+w_t^i\cdot f_t^i.
\end{equation}
Although it seems reasonable, there is a key risk of unstable weighting. As $w_s$ and $w_t$ are derived separately from two independent networks with totally different inputs (i.e., static frame and optical flow), there is no guarantee that $w_s$ and $w_t$ can serve as the weighting factor, in other word, being comparable. Therefore, the naive aggregation approach of (\ref{eqn.weightedsum}) may result in unstable weighting of spatial and temporal information.

\begin{figure}[h]
\centering
\includegraphics[width=0.45\textwidth]{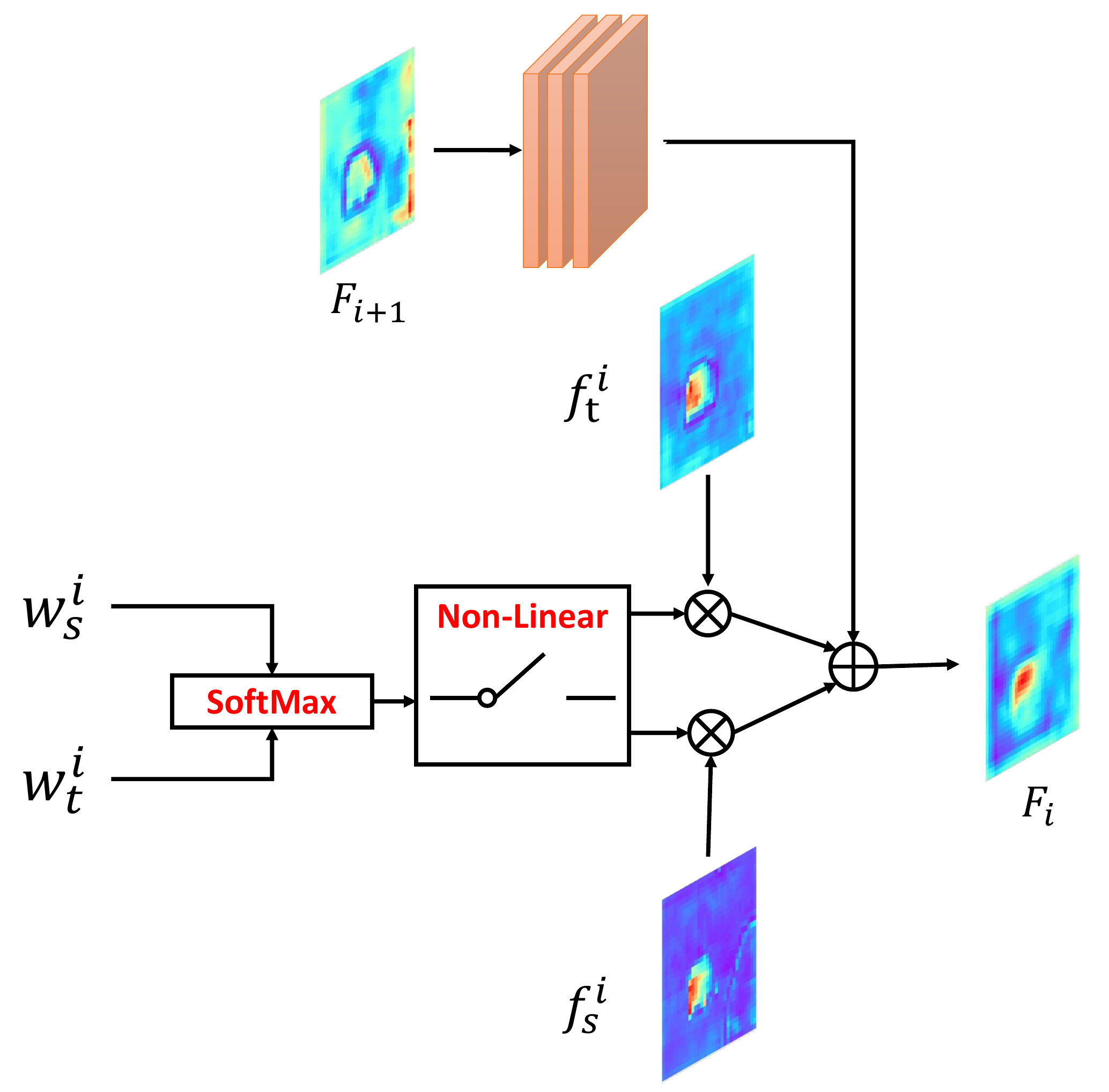}
\caption{Illustration of cross attentive aggregation.}
\label{CAA}
\end{figure}

To address this issue, instead of direct aggregation using independent weight vectors, we propose a simple yet efficient cross attentive aggregation scheme to jointly evaluate the relative reliability of spatial and temporal features. Fig.~\ref{CAA} illustrates the top-down cross attentive feature aggregation of coarser scale feature $F_{i+1}$ with spatial feature $f_s^i$ and temporal feature $f_t^i$. As shown in the figure, the spatial and temporal dynamic weights $w_s^i$ and $w_t^i$ are first normalized with `cross' softmax, making them ranging from 0 to 1 with the summation of 1. Denote the regularized dynamic weight vector as $\bm{v}_s$ and $\bm{v}_t$ for spatial and temporal branch respectively, we have:

\begin{equation}
\begin{array}{l}
v_s^i=\frac{e^{w_s^i}}{e^{w_s^i}+e^{w_t^i}}\\ \\
v_t^i=\frac{e^{w_t^i}}{e^{w_s^i}+e^{w_t^i}}.
\end{array}
\label{softmax_t}
\end{equation}


Another benefit of applying softmax is to exaggerate the gap between spatial and temporal reliability. As mentioned above, it is highly likely that one single saliency cue (spatial or temporal) is obviously much better than the other. In this case, indiscriminate aggregation of the poor saliency branch will heavily degrade the overall performance even if the other saliency cue performs excellently. To further suppress the distracting impact from the noisy or invalid saliency branch, a non-linear cross thresholding is proposed as:
\begin{equation}
(u_s^i,u_t^i)=\left\{
\begin{array}{lcl}
(\ 0\ ,v_t^i)            &   & {v_t^i-v_s^i>\tau}\\
(v_s^i,\ 0\ )            &   & {v_t^i-v_s^i<-\tau}\\
(v_s^i,v_t^i)  &   & {otherwise}
\end{array}
\right.,
\end{equation}
where $\tau$ is a threshold. The generated dynamic weights $u_s^i$ and $u_t^i$ are then used to weight the corresponding features extracted from spatial and temporal branches for complementary spatiotemporal features, which can be expressed as:
\begin{equation}
F_{agg}^i=u_s^i\cdot f_s^i+u_t^i\cdot f_t^i.
\end{equation}

It is worth noting that compared with existing cross attention modules~\cite{9157122, 8953566} where complex operations such as correlations are generally carried out on the whole feature maps, the proposed cross attentive aggregation is of extremely low complexity since the interaction between two-bypass network are carried out on the 5-d dynamic weight vector, which can also be easily obtained as described above. As will be demonstrated by the experiments, the simple yet efficient cross operations of softmax and non-linear thresholding can well determine the relative reliability of spatial and temporal saliency branch, enhancing the final performance.

\begin{table}[h] 
\caption{The parameters of three convolution kernels in five CAA modules. `ks', `pd' and `chn' denote the kernel size, padding and output channel numbers, respectively.} 
\small
\centering
\begin{tabular}{c|c|c|c|c|c|c|c|c|c} 
\hline
& \multicolumn{3}{c|}{Conv1} & \multicolumn{3}{c|}{Conv2} & \multicolumn{3}{c}{Conv3} \\ \hline
& ks & pd & chn & ks & pd & chn & ks & pd & chn \\ \hline
CAA 1 & 3      & 1      & 64       & 3      & 1      & 64       & 3      & 1      & 64       \\
CAA 2 & 3      & 1      & 128      & 3      & 1      & 128      & 3      & 1      & 64       \\
CAA 3 & 5      & 2      & 256      & 5      & 2      & 256      & 5      & 2      & 128      \\
CAA 4 & 7      & 3      & 512      & 7      & 3      & 512      & 7      & 3      & 256      \\
CAA 5 & 3      & 1      & 256      & 3      & 1      & 256      & 3      & 1      & 512      \\ 
\hline
\end{tabular}

\label{kernal}
\end{table}

For the coarse scale feature $F_{i+1}$, it first goes through three convolution layers to gradually reduce the channel number to match that of spatiotemporal feature at current scale $F_{agg}^i$. The parameters of convolution kernels in five CAA modules are listed in Table~\ref{kernal}. Then, the coarse scale feature is upsampled to match the resolution of $F_{agg}^i$. The final progressive aggregation can then be formulated as:
\begin{equation}
F_i=\mathcal{T}(F_{i+1})+F_{agg}^i, \qquad i=1,2,3,4,
\label{top-down}
\end{equation}
where $\mathcal{T}(\cdot)$ stands for three convolution layers and one upsampling layer. For the coarsest scale $i=5$, only current scale of features are available, so $F_5=F_{agg}^5$. The final output feature after the proposed top-down progressive aggregation scheme can be denoted as $F_{final}=\mathcal{T}(F_1)$.

\begin{table*}[]
\scriptsize
\caption{Comparison with state-of-the-art video salient object detection algorithms. "$\uparrow$" means larger is better and "$\downarrow$" means smaller is better. "-"  indicates the model has been trained on this dataset. "- -" represents the traditional methods. "*" denotes that this algorithm has accessed to unfixed backbones. The top two methods are marked in red and blue, respectively.}
\centering
\label{comparison}
\begin{tabular}{c|c|c|c|c|c|c|c|c|c|c|c|c|c}
\toprule[.15em]
\multicolumn{2}{c|}{Method} &
  MBD &
  MSTM &
  STBP &
  SCOM &
  SCNN &
  FCNS &
  FGRNE &
  PDB &
  SSAV &
  MGA &
  PCSA &
  ours \\ 
\midrule[.15em]
\multicolumn{2}{c|}{Backbone} &
  - - &
  - - &
  - - &
  - - &
  VGG &
  VGG &
  * &
  ResNet50 &
  ResNet50 &
  ResNet101 &
  MobileNet &
  ResNet34 \\ 
\midrule[.15em]
 &
  maxF $\uparrow$ &
  0.470 &
  0.429 &
  0.544 &
  0.783 &
  0.714 &
  0.708 &
  0.783 &
  0.855 &
  0.861 &
  {\color[HTML]{FE0000} \textbf{0.892}} &
  0.880 &
  {\color[HTML]{0000FF} \textbf{0.891}} \\ \cmidrule{2-14}
 &
  S $\uparrow$ &
  0.597 &
  0.583 &
  0.677 &
  0.832 &
  0.783 &
  0.794 &
  0.838 &
  0.882 &
  0.893 &
  {\color[HTML]{0000FF} \textbf{0.912}} &
  0.902 &
  {\color[HTML]{FE0000} \textbf{0.914}} \\
  \cmidrule{2-14} 
\multirow{-3}{*}{DAVIS} &
  MAE $\downarrow$ &
  0.177 &
  0.165 &
  0.096 &
  0.048 &
  0.064 &
  0.061 &
  0.043 &
  0.028 &
  0.028 &
  {\color[HTML]{0000FF} \textbf{0.022}} &
  {\color[HTML]{0000FF} \textbf{0.022}} &
  {\color[HTML]{FE0000} \textbf{0.018}} \\ 
\cmidrule{1-14}
 &
  maxF $\uparrow$ &
  0.487 &
  0.500 &
  0.595 &
  0.797 &
  0.762 &
  0.759 &
  0.767 &
  0.821 &
  0.865 &
  {\color[HTML]{FE0000} \textbf{0.907}} &
  0.831 &
  {\color[HTML]{0000FF} \textbf{0.875}} \\ \cmidrule{2-14} 
 &
  S $\uparrow$ &
  0.609 &
  0.613 &
  0.627 &
  0.794 &
  0.794 &
  0.794 &
  0.809 &
  0.851 &
  0.879 &
  {\color[HTML]{FE0000} \textbf{0.910}} &
  0.866 &
  {\color[HTML]{0000FF} \textbf{0.895}} \\ \cmidrule{2-14} 
\multirow{-3}{*}{FBMS} &
  MAE $\downarrow$ &
  0.206 &
  0.177 &
  0.152 &
  0.079 &
  0.095 &
  0.091 &
  0.088 &
  0.064 &
  0.040 &
  {\color[HTML]{FE0000} \textbf{0.026}} &
  0.041 &
  {\color[HTML]{0000FF} \textbf{0.034}} \\ 
\cmidrule{1-14}
 &
  maxF $\uparrow$ &
  0.692 &
  0.673 &
  0.622 &
  0.831 &
  0.831 &
  0.852 &
  0.848 &
  0.888 &
  0.939 &
  {\color[HTML]{0000FF} \textbf{0.940}} &
  {\color[HTML]{0000FF} \textbf{0.940}} &
  {\color[HTML]{FE0000} \textbf{0.950}} \\ \cmidrule{2-14}
 &
  S $\uparrow$ &
  0.726 &
  0.749 &
  0.629 &
  0.762 &
  0.847 &
  0.881 &
  0.861 &
  0.907 &
  0.939 &
  0.941 &
  {\color[HTML]{0000FF} \textbf{0.946}} &
  {\color[HTML]{FE0000} \textbf{0.949}} \\ \cmidrule{2-14}
\multirow{-3}{*}{ViSal} &
  MAE $\downarrow$ &
  0.129 &
  0.095 &
  0.163 &
  0.122 &
  0.071 &
  0.048 &
  0.045 &
  0.032 &
  0.020 &
  {\color[HTML]{0000FF} \textbf{0.016}} &
  0.017 &
  {\color[HTML]{FE0000} \textbf{0.013}} \\ 
\cmidrule{1-14}
 &
  maxF $\uparrow$ &
  0.562 &
  0.336 &
  0.403 &
  0.690 &
  0.609 &
  0.675 &
  0.669 &
  0.742 &
  0.742 &
  0.745 &
  {\color[HTML]{0000FF} \textbf{0.747}} &
  {\color[HTML]{FE0000} \textbf{0.801}} \\ \cmidrule{2-14} 
 &
  S $\uparrow$ &
  0.661 &
  0.551 &
  0.614 &
  0.712 &
  0.704 &
  0.760 &
  0.715 &
  0.818 &
  0.819 &
  0.806 &
  {\color[HTML]{0000FF} \textbf{0.827}} &
  {\color[HTML]{FE0000} \textbf{0.855}} \\ \cmidrule{2-14}
\multirow{-3}{*}{VOS} &
  MAE $\downarrow$ &
  0.158 &
  0.145 &
  0.105 &
  0.162 &
  0.109 &
  0.099 &
  0.097 &
  0.078 &
  0.073 &
  0.070 &
  {\color[HTML]{0000FF} \textbf{0.065}} &
  {\color[HTML]{FE0000} \textbf{0.060}} \\ 
\cmidrule{1-14}
 &
  maxF $\uparrow$ &
  0.554 &
  0.526 &
  0.640 &
  0.764 &
  - &
  - &
  - &
  0.800 &
  0.801 &
  {\color[HTML]{0000FF} \textbf{0.828}} &
  0.810 &
  {\color[HTML]{FE0000} \textbf{0.832}} \\ \cmidrule{2-14}
 &
  S $\uparrow$ &
  0.618 &
  0.643 &
  0.735 &
  0.815 &
  - &
  - &
  - &
  0.864 &
  0.851 &
  {\color[HTML]{FE0000} \textbf{0.885}} &
  0.865 &
  {\color[HTML]{0000FF} \textbf{0.875}} \\ \cmidrule{2-14} 
\multirow{-3}{*}{SegV2} &
  MAE $\downarrow$ &
  0.146 &
  0.114 &
  0.061 &
  0.030 &
  - &
  - &
  - &
  {\color[HTML]{0000FF} \textbf{0.024}} &
  {\color[HTML]{FE0000} \textbf{0.023}} &
  0.026 &
  0.025 &
  0.028 \\ 
\bottomrule[.15em]
\end{tabular}
\end{table*}

\subsection{Spatial Attention based on Coarse Saliency Map}
\label{coarse saliency}
In the dynamic weight generation module, the 5-d weight vectors are learned to represent the reliability of input saliency features. When the feature extracted by one branch is better than that from the other branch, larger weights are assigned to it. Otherwise, smaller weights are assigned. Under this circumstance, the 5-d vectors $u_s^i$ and $u_t^i$ as a whole can be regarded as the reliability for the corresponding spatial and temporal saliency branch. Therefore, we sum up the 5-d vector to get the dynamic reliability weight for the corresponding saliency map. Given the spatial saliency map $S_s$ and temporal saliency map $S_t$, the coarse saliency map is derived as:
\begin{equation}
S_c=\epsilon_s\cdot S_{s}+\epsilon_t\cdot S_t,
\label{weight_pre}
\end{equation}
where $\epsilon_{s}=\frac{\sum v_{s}^i}{\sum v_s^i+\sum v_t^i}$, $\epsilon_{t}=\frac{\sum v_{t}^i}{\sum v_s^i+\sum v_t^i}$ are the reliability weights for the corresponding branches.

As shown in Fig.~\ref{network}, motivated by~\cite{9010060}, the coarse prediction map is adopted as an accurate guidance to suppress background noises of spatiotemporal feature $F_{final}$. The output feature $F_{att}$ can be written as:
\begin{equation}
F_{att}=F_{final}\cdot S_c+F_{final}.
\label{attention}
\end{equation}
Finally, the feature map $F_{att}$ is passed to a decoder to get the final saliency map $S_f$.

The benefit to generate the coarse saliency map in attentive way is two-fold: 1) Through dynamic weighting the coarse spatial and temporal saliency maps, a more reliable coarse saliency map can be generated, which can facilitate more effective spatial attention. 2) As the dynamic weight for spatial and temporal coarse saliency map is the sum of weight vectors, it will promote the adaptive learning of the weight vectors with multiple supervision in the training process.

\subsection{Loss Function}
Without loss of generality, as the proposed saliency model is with multiple saliency outputs, the network is trained in a multiple supervision manner for quick convergence and superior performance. The overall loss function is composed of four components from three coarse saliency maps and the final saliency map. Cross entropy losses are adopted, which are defined as:
\begin{equation}
l=-\frac{1}{N}[\sum_{k\in Y_+}\log{}S(k)+\sum_{k\in Y_-}\log{}\left(1-S\left(k\right)\right)],
\label{cross_entropy}
\end{equation}
where $Y$ denotes the saliency ground truth, and $N$ is the total number of pixels in the input image. $S$ is the predicted saliency map. $Y_+$ and $Y_-$ denote the positive part and negative part of the ground truth respectively. Let the $l_{s}$, $l_{t}$, $l_{c}$, and $l_{f}$ represent the losses between the saliency ground truth and the three coarse saliency maps $S_s$, $S_t$, $S_c$ and final saliency map $S_f$, respectively. The total loss function of the network is calculated as:
\begin{equation}
L=l_s+l_t+l_c+l_f.
\label{loss}
\end{equation}

\section{Experimental Results}
\label{experiment}

\subsection{Implementation Details}
\begin{figure*}[t]
\centering
\begin{minipage}[b]{0.08\linewidth}
  \centering
  \includegraphics[width=1\linewidth, height=1cm]{./img_break.jpg}
\end{minipage}
\begin{minipage}[b]{0.08\linewidth}
  \centering
  \includegraphics[width=1\linewidth, height=1cm]{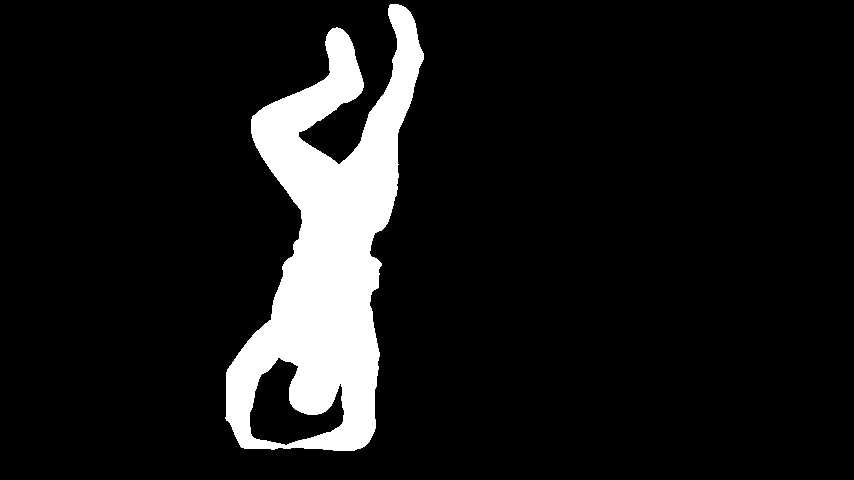}
\end{minipage}
\begin{minipage}[b]{0.08\linewidth}
  \centering
  \includegraphics[width=1\linewidth, height=1cm]{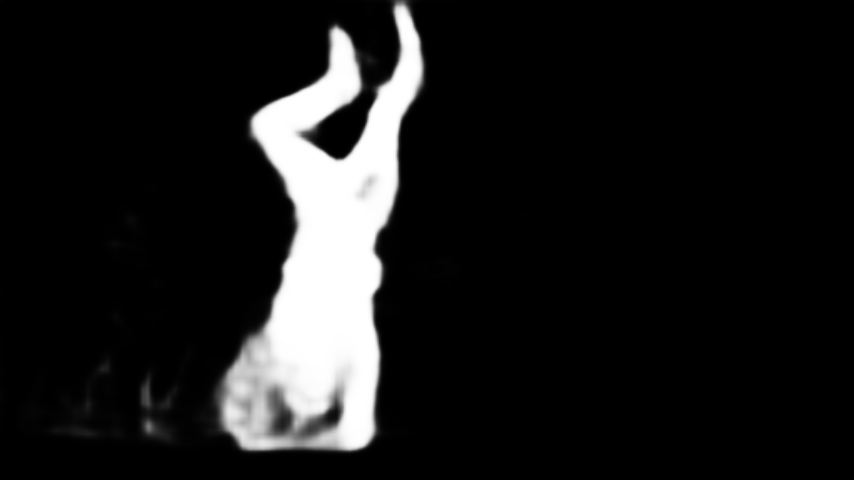}
\end{minipage}
\begin{minipage}[b]{0.08\linewidth}
  \centering
  \includegraphics[width=1\linewidth, height=1cm]{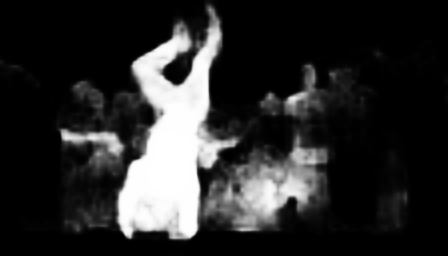}
\end{minipage}
\begin{minipage}[b]{0.08\linewidth}
  \centering
  \includegraphics[width=1\linewidth, height=1cm]{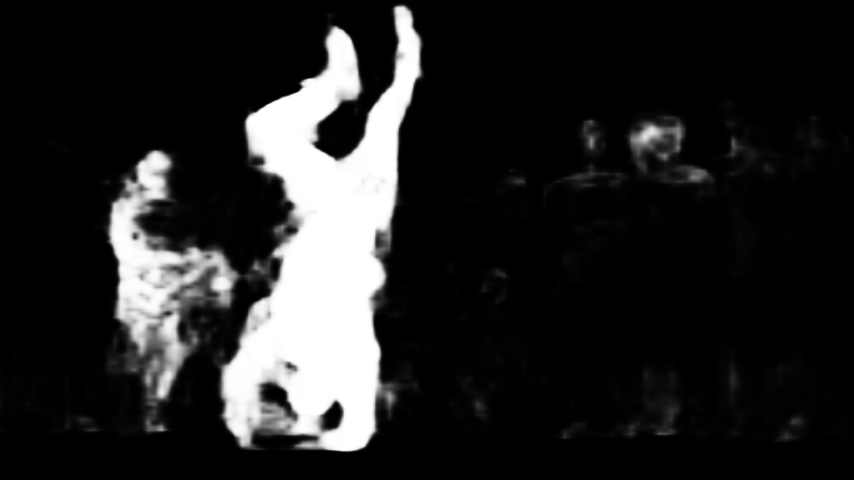}
\end{minipage}
\begin{minipage}[b]{0.08\linewidth}
  \centering
  \includegraphics[width=1\linewidth, height=1cm]{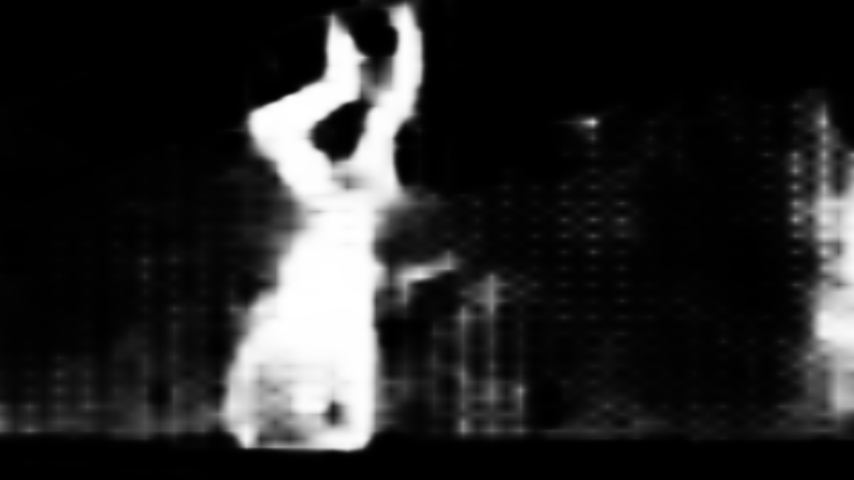}
\end{minipage}
\begin{minipage}[b]{0.08\linewidth}
  \centering
  \includegraphics[width=1\linewidth, height=1cm]{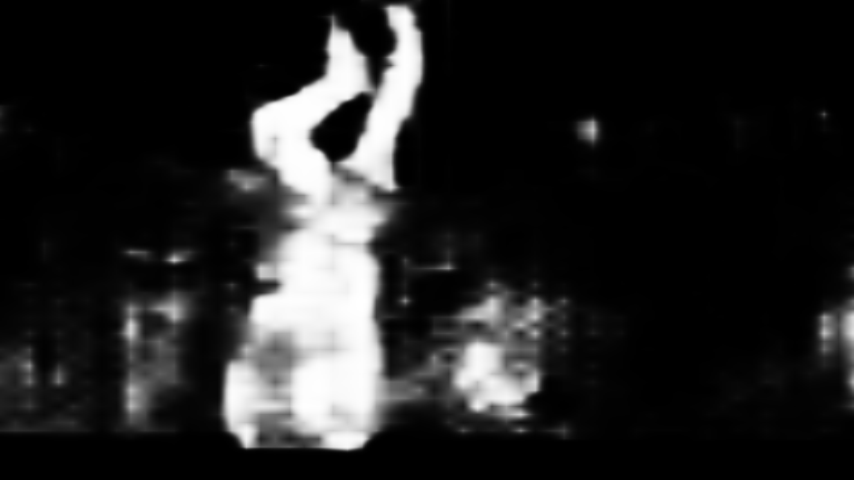}
\end{minipage}
\begin{minipage}[b]{0.08\linewidth}
  \centering
  \includegraphics[width=1\linewidth, height=1cm]{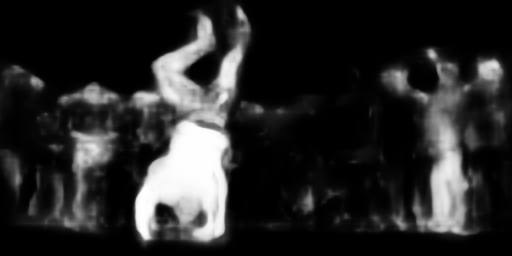}
\end{minipage}
\begin{minipage}[b]{0.08\linewidth}
  \centering
  \includegraphics[width=1\linewidth, height=1cm]{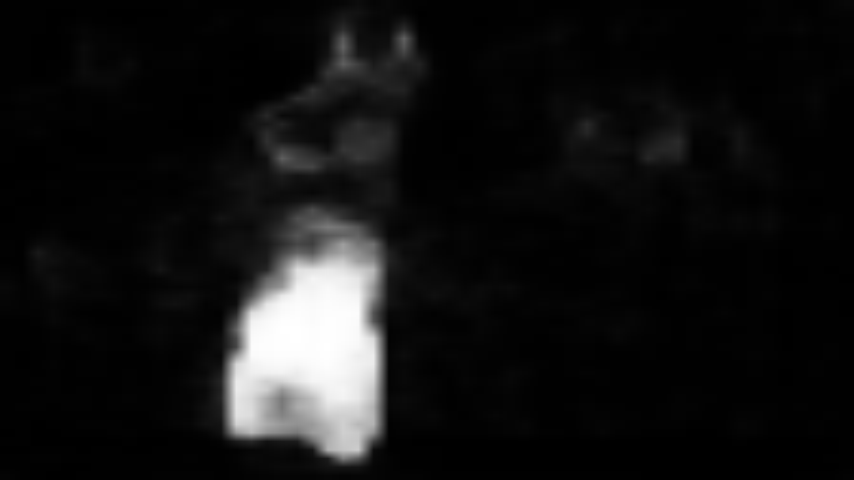}
\end{minipage}
\begin{minipage}[b]{0.08\linewidth}
  \centering
  \includegraphics[width=1\linewidth, height=1cm]{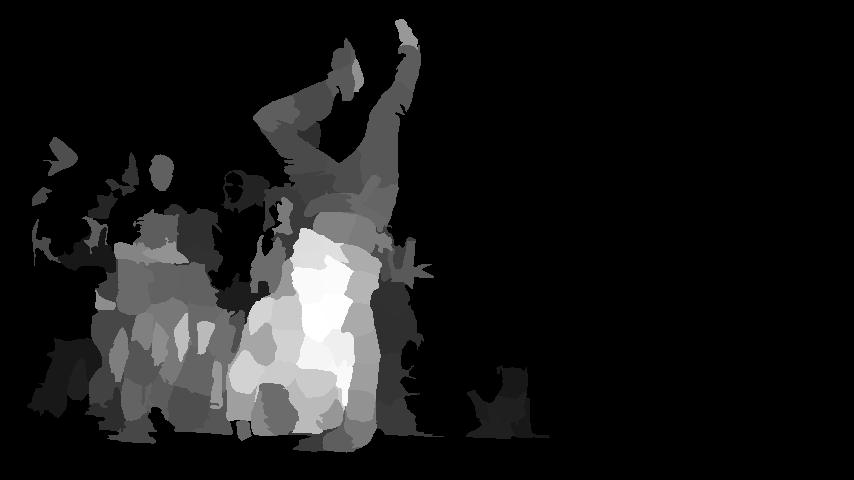}
\end{minipage}
\begin{minipage}[b]{0.08\linewidth}
  \centering
  \includegraphics[width=1\linewidth, height=1cm]{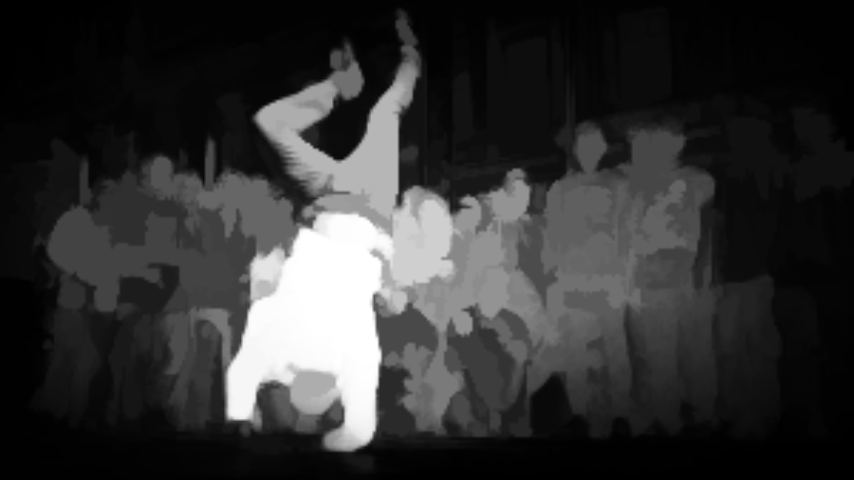}
\end{minipage}
\vspace{3pt}
\\
\begin{minipage}[b]{0.08\linewidth}
  \centering
  \includegraphics[width=1\linewidth, height=1cm]{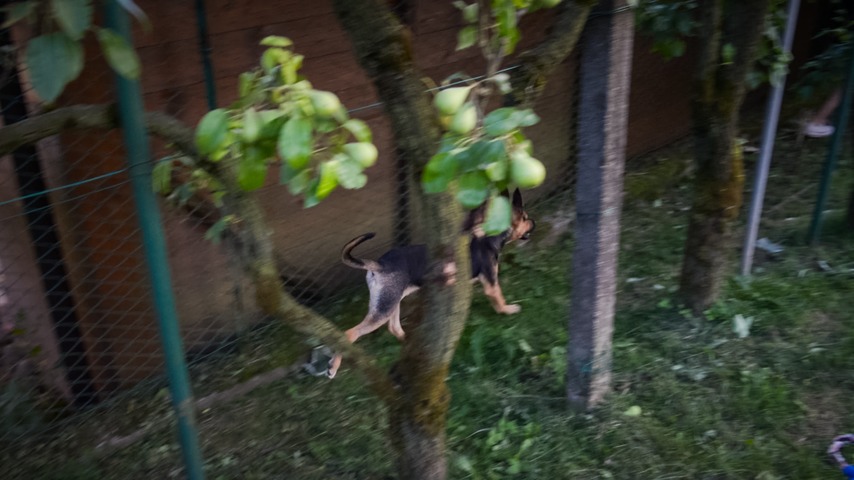}
\end{minipage}
\begin{minipage}[b]{0.08\linewidth}
  \centering
  \includegraphics[width=1\linewidth, height=1cm]{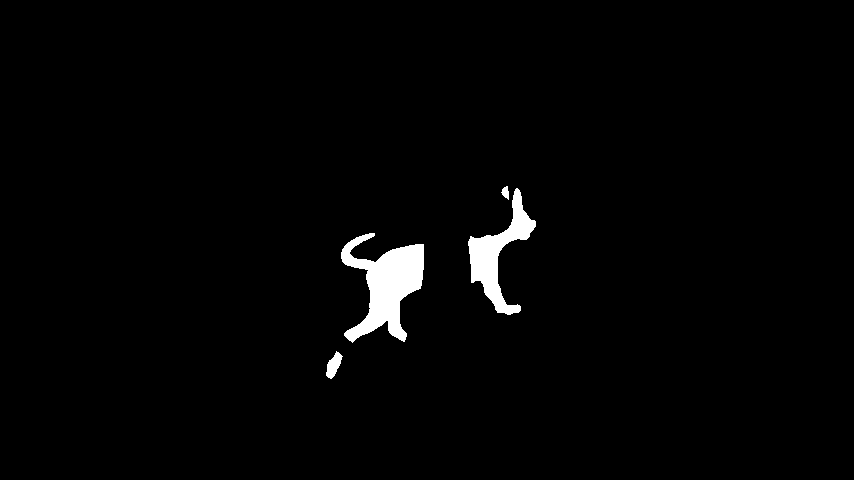}
\end{minipage}
\begin{minipage}[b]{0.08\linewidth}
  \centering
  \includegraphics[width=1\linewidth, height=1cm]{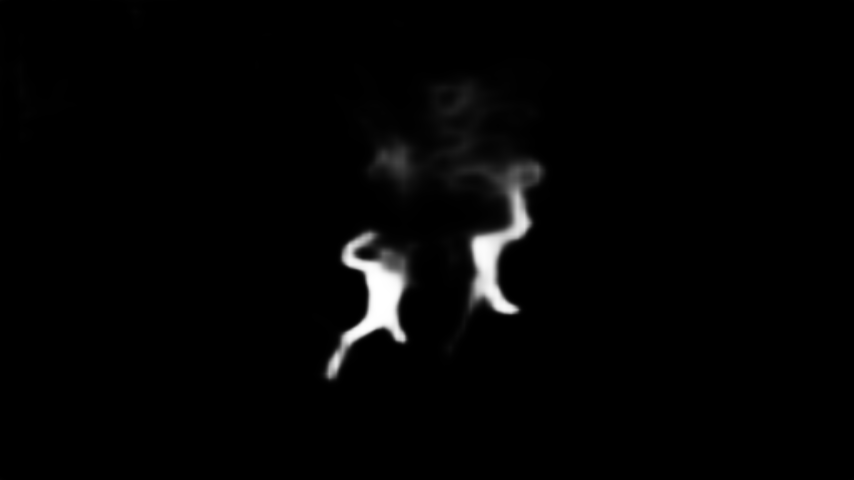}
\end{minipage}
\begin{minipage}[b]{0.08\linewidth}
  \centering
  \includegraphics[width=1\linewidth, height=1cm]{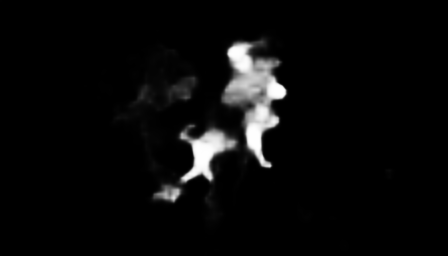}
\end{minipage}
\begin{minipage}[b]{0.08\linewidth}
  \centering
  \includegraphics[width=1\linewidth, height=1cm]{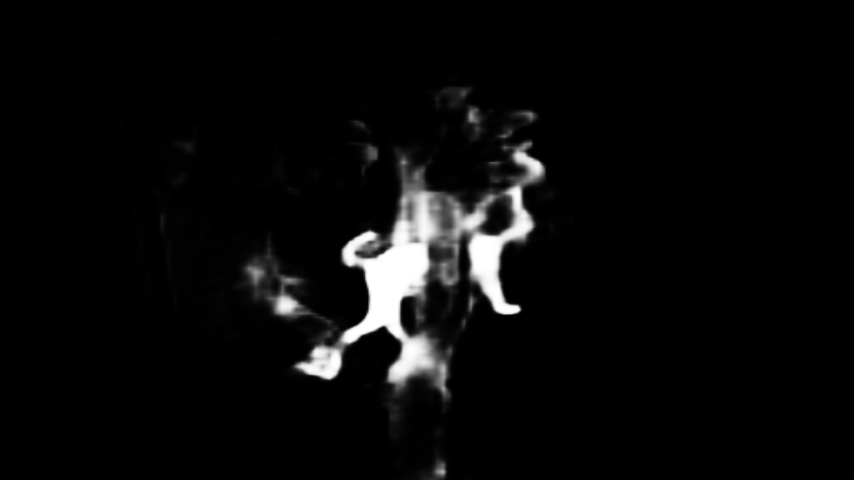}
\end{minipage}
\begin{minipage}[b]{0.08\linewidth}
  \centering
  \includegraphics[width=1\linewidth, height=1cm]{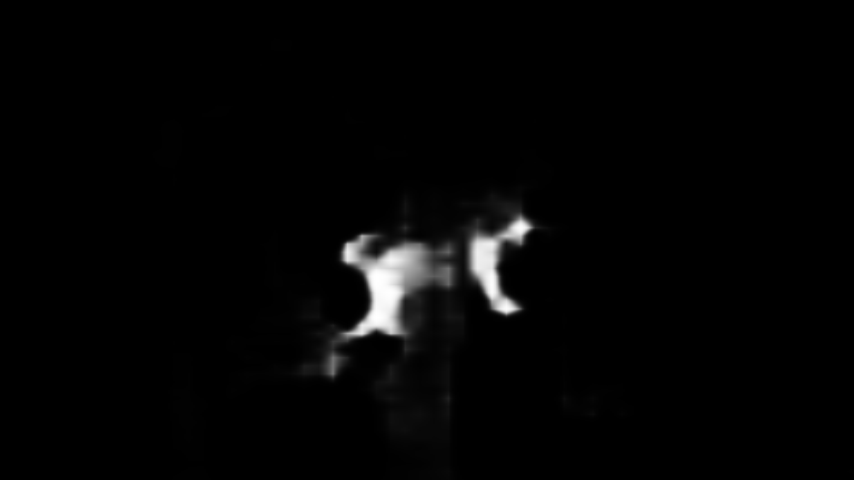}
\end{minipage}
\begin{minipage}[b]{0.08\linewidth}
  \centering
  \includegraphics[width=1\linewidth, height=1cm]{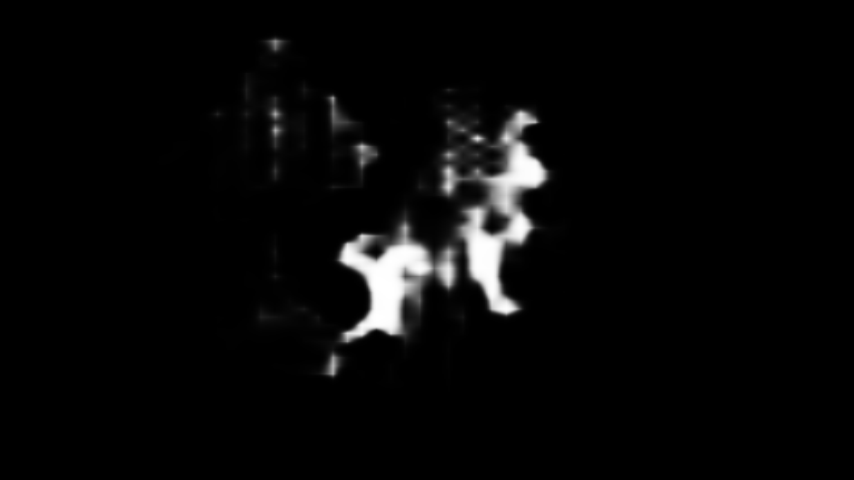}
\end{minipage}
\begin{minipage}[b]{0.08\linewidth}
  \centering
  \includegraphics[width=1\linewidth, height=1cm]{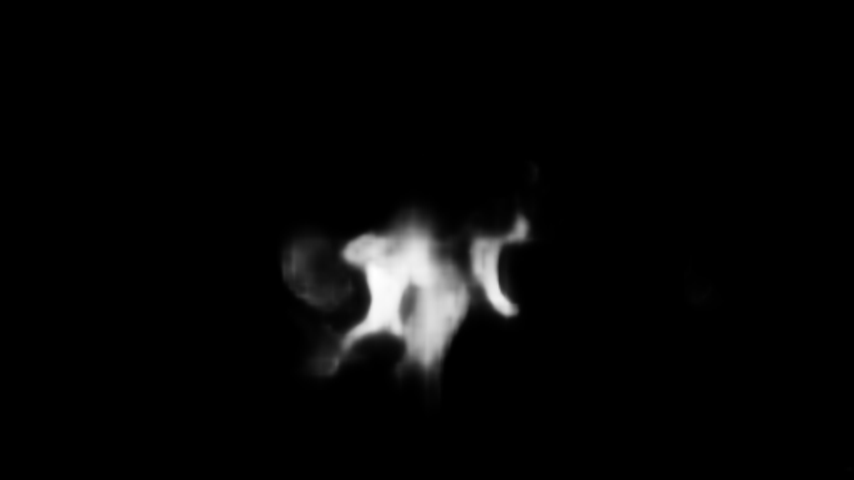}
\end{minipage}
\begin{minipage}[b]{0.08\linewidth}
  \centering
  \includegraphics[width=1\linewidth, height=1cm]{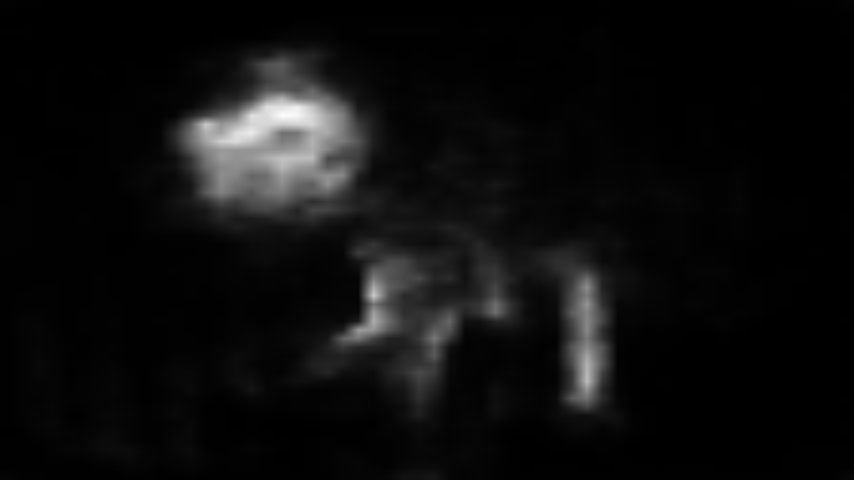}
\end{minipage}
\begin{minipage}[b]{0.08\linewidth}
  \centering
  \includegraphics[width=1\linewidth, height=1cm]{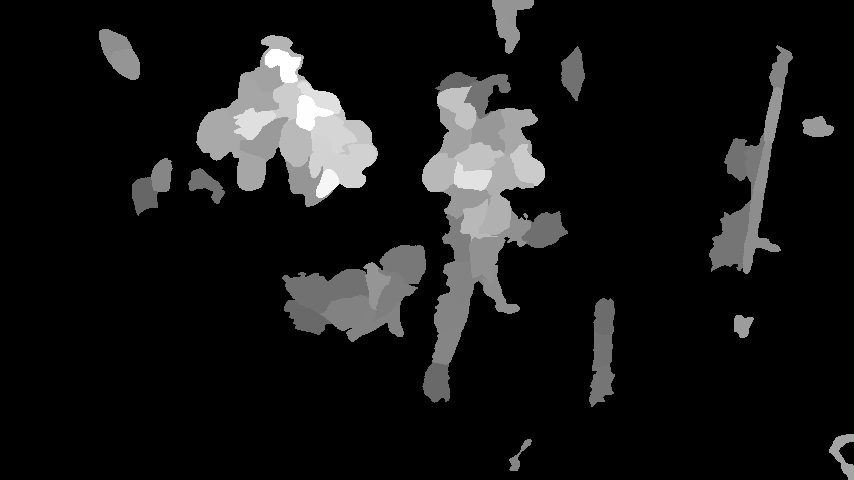}
\end{minipage}
\begin{minipage}[b]{0.08\linewidth}
  \centering
  \includegraphics[width=1\linewidth, height=1cm]{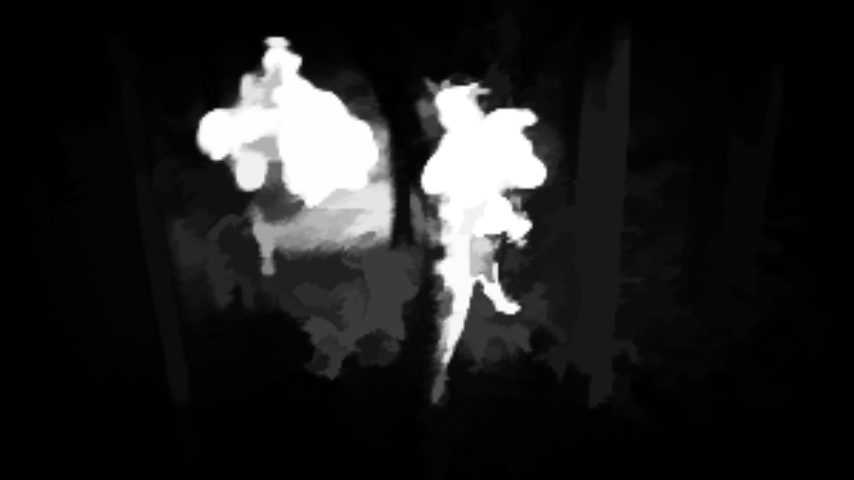}
\end{minipage}
\vspace{3pt}
\\
\begin{minipage}[b]{0.08\linewidth}
  \centering
  \includegraphics[width=1\linewidth, height=1cm]{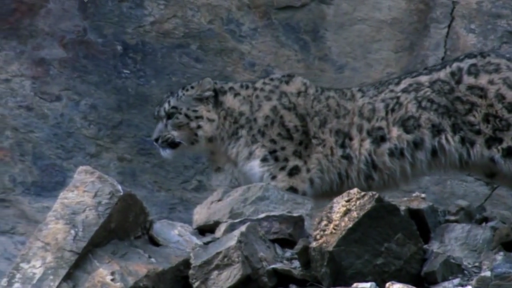}
\end{minipage}
\begin{minipage}[b]{0.08\linewidth}
  \centering
  \includegraphics[width=1\linewidth, height=1cm]{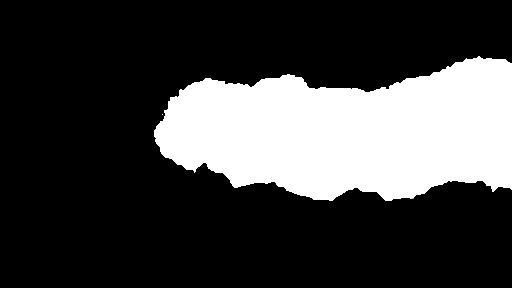}
\end{minipage}
\begin{minipage}[b]{0.08\linewidth}
  \centering
  \includegraphics[width=1\linewidth, height=1cm]{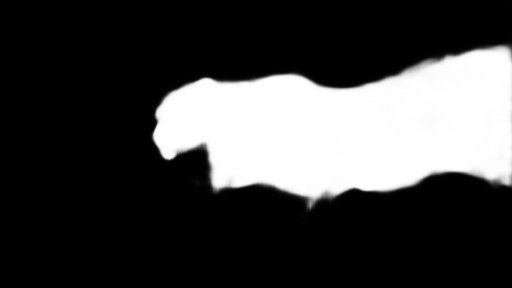}
\end{minipage}
\begin{minipage}[b]{0.08\linewidth}
  \centering
  \includegraphics[width=1\linewidth, height=1cm]{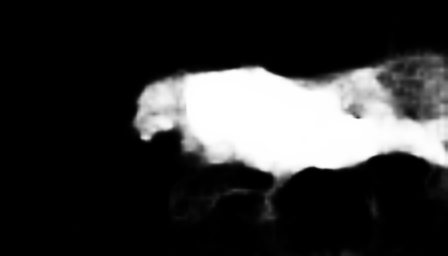}
\end{minipage}
\begin{minipage}[b]{0.08\linewidth}
  \centering
  \includegraphics[width=1\linewidth, height=1cm]{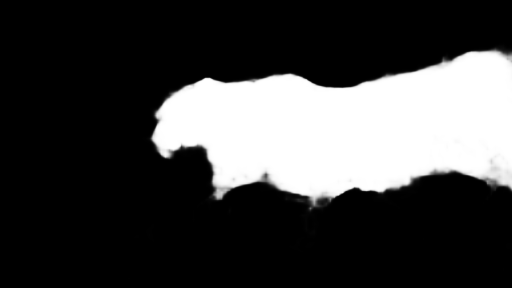}
\end{minipage}
\begin{minipage}[b]{0.08\linewidth}
  \centering
  \includegraphics[width=1\linewidth, height=1cm]{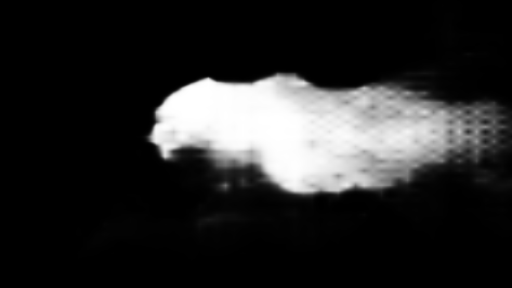}
\end{minipage}
\begin{minipage}[b]{0.08\linewidth}
  \centering
  \includegraphics[width=1\linewidth, height=1cm]{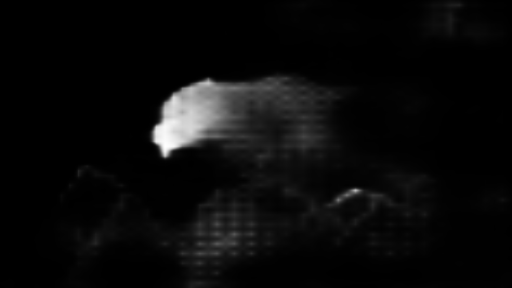}
\end{minipage}
\begin{minipage}[b]{0.08\linewidth}
  \centering
  \includegraphics[width=1\linewidth, height=1cm]{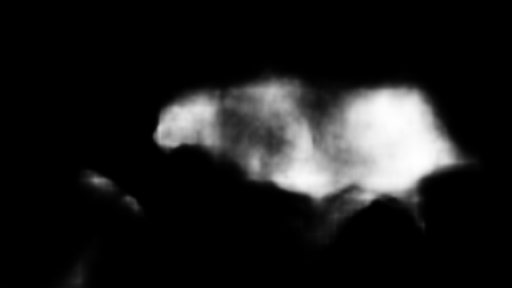}
\end{minipage}
\begin{minipage}[b]{0.08\linewidth}
  \centering
  \includegraphics[width=1\linewidth, height=1cm]{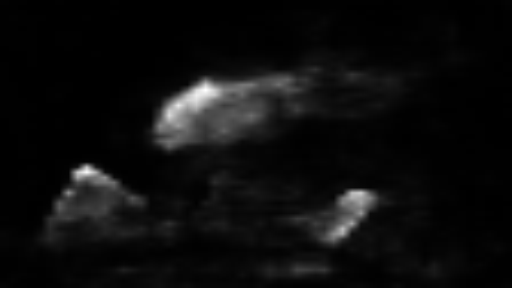}
\end{minipage}
\begin{minipage}[b]{0.08\linewidth}
  \centering
  \includegraphics[width=1\linewidth, height=1cm]{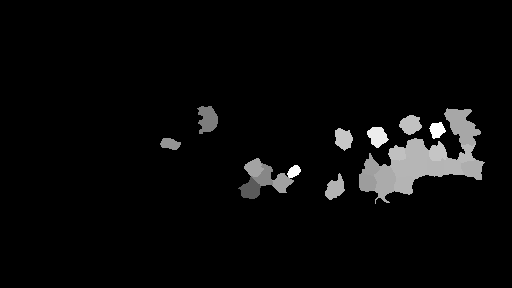}
\end{minipage}
\begin{minipage}[b]{0.08\linewidth}
  \centering
  \includegraphics[width=1\linewidth, height=1cm]{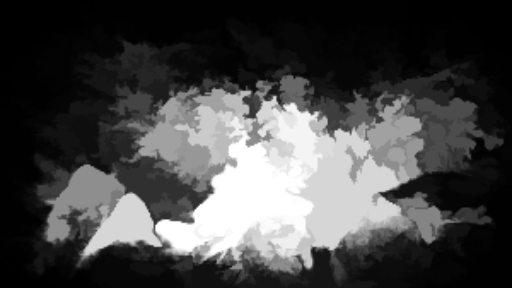}
\end{minipage}
\vspace{3pt}
\\
\begin{minipage}[b]{0.08\linewidth}
  \centering
  \includegraphics[width=1\linewidth, height=1cm]{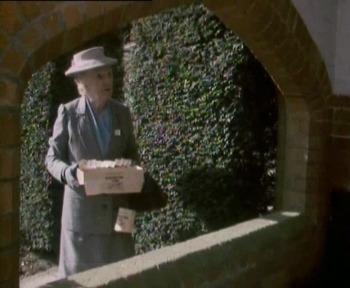}
\end{minipage}
\begin{minipage}[b]{0.08\linewidth}
  \centering
  \includegraphics[width=1\linewidth, height=1cm]{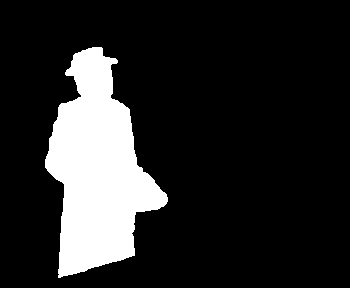}
\end{minipage}
\begin{minipage}[b]{0.08\linewidth}
  \centering
  \includegraphics[width=1\linewidth, height=1cm]{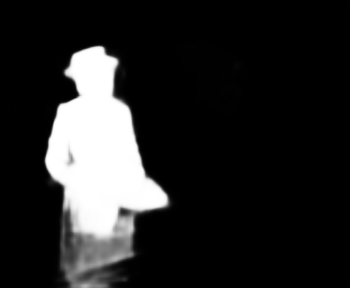}
\end{minipage}
\begin{minipage}[b]{0.08\linewidth}
  \centering
  \includegraphics[width=1\linewidth, height=1cm]{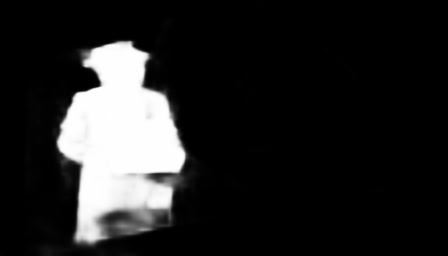}
\end{minipage}
\begin{minipage}[b]{0.08\linewidth}
  \centering
  \includegraphics[width=1\linewidth, height=1cm]{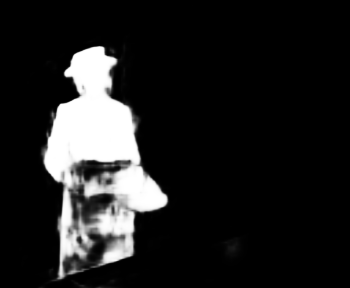}
\end{minipage}
\begin{minipage}[b]{0.08\linewidth}
  \centering
  \includegraphics[width=1\linewidth, height=1cm]{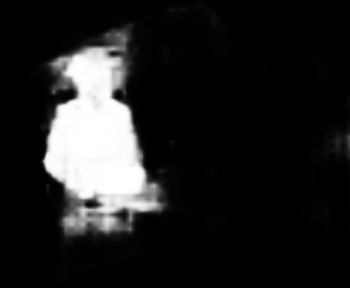}
\end{minipage}
\begin{minipage}[b]{0.08\linewidth}
  \centering
  \includegraphics[width=1\linewidth, height=1cm]{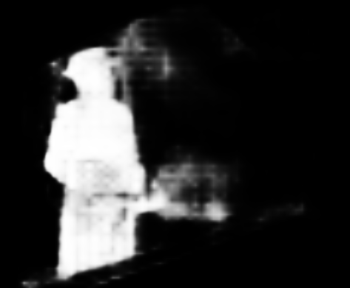}
\end{minipage}
\begin{minipage}[b]{0.08\linewidth}
  \centering
  \includegraphics[width=1\linewidth, height=1cm]{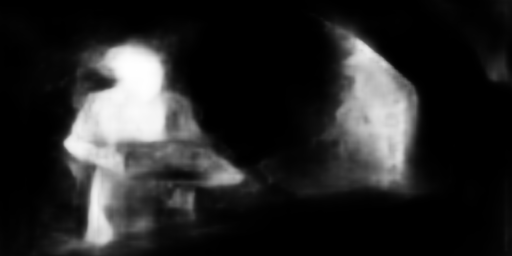}
\end{minipage}
\begin{minipage}[b]{0.08\linewidth}
  \centering
  \includegraphics[width=1\linewidth, height=1cm]{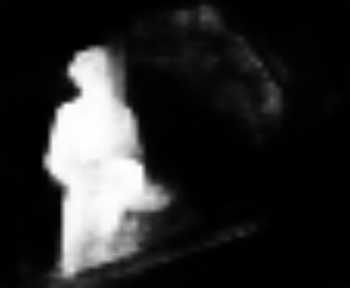}
\end{minipage}
\begin{minipage}[b]{0.08\linewidth}
  \centering
  \includegraphics[width=1\linewidth, height=1cm]{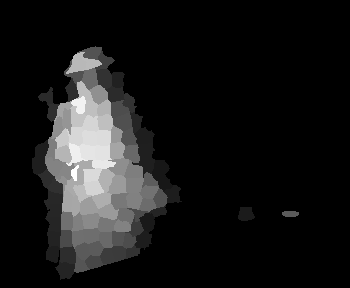}
\end{minipage}
\begin{minipage}[b]{0.08\linewidth}
  \centering
  \includegraphics[width=1\linewidth, height=1cm]{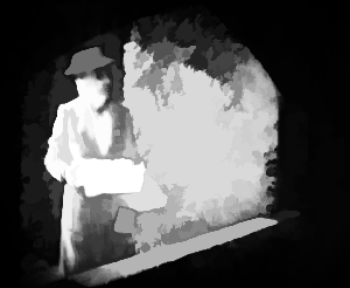}
\end{minipage}
\vspace{3pt}
\\
\begin{minipage}[b]{0.08\linewidth}
  \centering
  \includegraphics[width=1\linewidth, height=1cm]{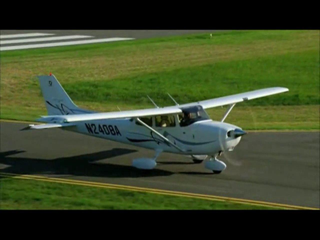}
\end{minipage}
\begin{minipage}[b]{0.08\linewidth}
  \centering
  \includegraphics[width=1\linewidth, height=1cm]{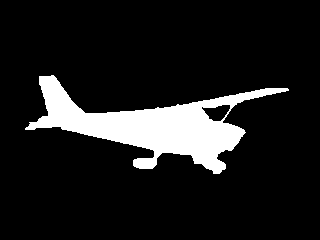}
\end{minipage}
\begin{minipage}[b]{0.08\linewidth}
  \centering
  \includegraphics[width=1\linewidth, height=1cm]{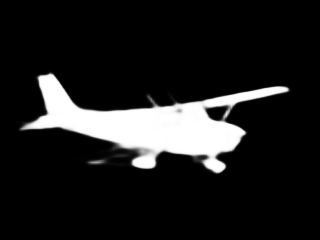}
\end{minipage}
\begin{minipage}[b]{0.08\linewidth}
  \centering
  \includegraphics[width=1\linewidth, height=1cm]{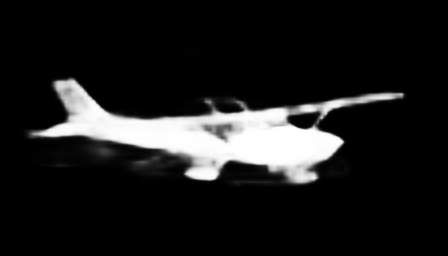}
\end{minipage}
\begin{minipage}[b]{0.08\linewidth}
  \centering
  \includegraphics[width=1\linewidth, height=1cm]{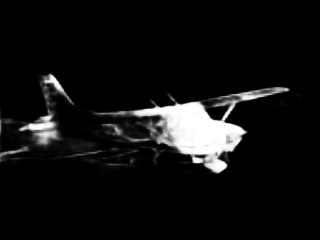}
\end{minipage}
\begin{minipage}[b]{0.08\linewidth}
  \centering
  \includegraphics[width=1\linewidth, height=1cm]{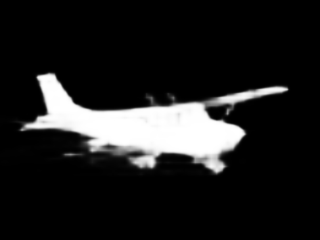}
\end{minipage}
\begin{minipage}[b]{0.08\linewidth}
  \centering
  \includegraphics[width=1\linewidth, height=1cm]{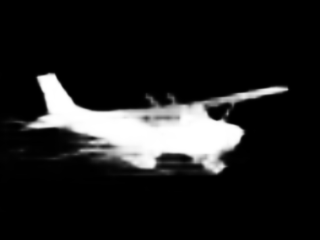}
\end{minipage}
\begin{minipage}[b]{0.08\linewidth}
  \centering
  \includegraphics[width=1\linewidth, height=1cm]{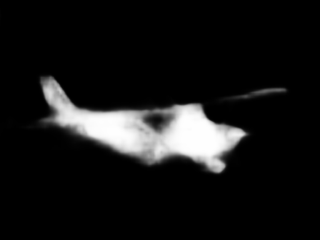}
\end{minipage}
\begin{minipage}[b]{0.08\linewidth}
  \centering
  \includegraphics[width=1\linewidth, height=1cm]{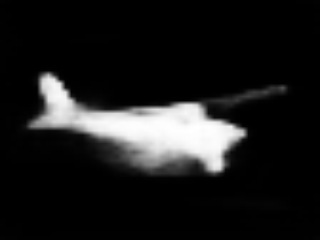}
\end{minipage}
\begin{minipage}[b]{0.08\linewidth}
  \centering
  \includegraphics[width=1\linewidth, height=1cm]{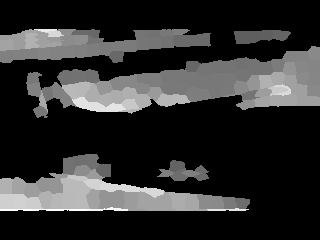}
\end{minipage}
\begin{minipage}[b]{0.08\linewidth}
  \centering
  \includegraphics[width=1\linewidth, height=1cm]{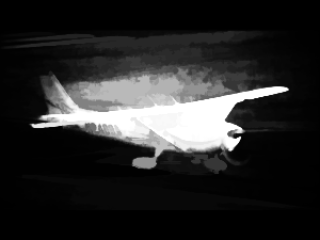}
\end{minipage}
\vspace{3pt}
\\
\begin{minipage}[b]{0.08\linewidth}
  \centering
  \includegraphics[width=1\linewidth, height=1cm]{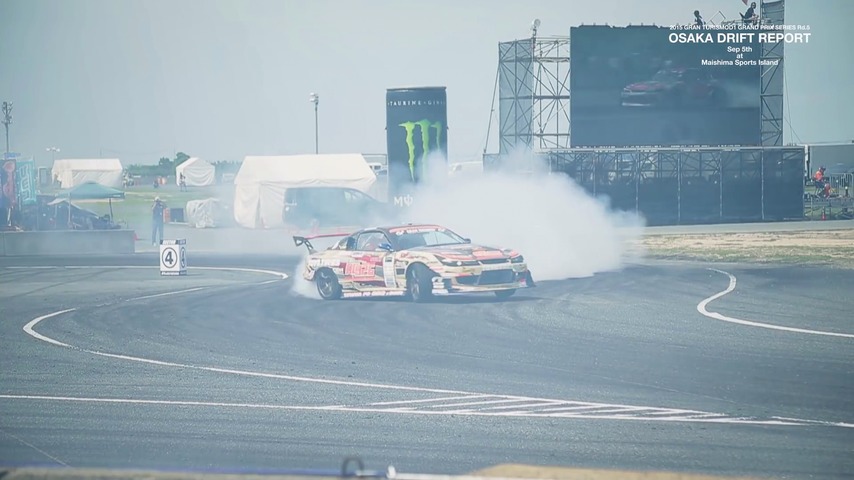}
\end{minipage}
\begin{minipage}[b]{0.08\linewidth}
  \centering
  \includegraphics[width=1\linewidth, height=1cm]{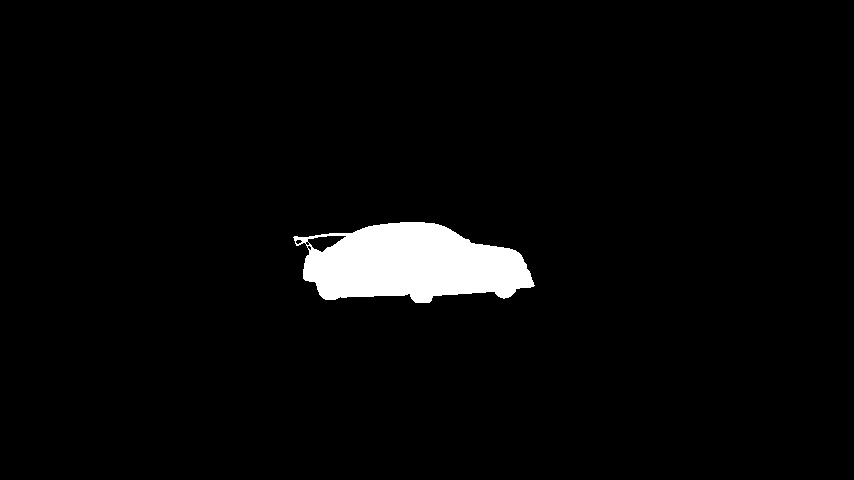}
\end{minipage}
\begin{minipage}[b]{0.08\linewidth}
  \centering
  \includegraphics[width=1\linewidth, height=1cm]{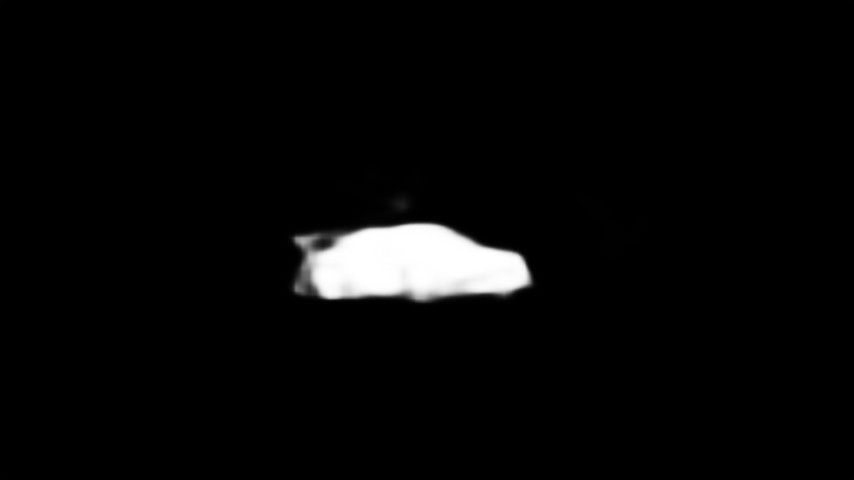}
\end{minipage}
\begin{minipage}[b]{0.08\linewidth}
  \centering
  \includegraphics[width=1\linewidth, height=1cm]{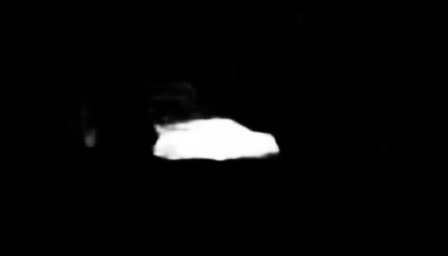}
\end{minipage}
\begin{minipage}[b]{0.08\linewidth}
  \centering
  \includegraphics[width=1\linewidth, height=1cm]{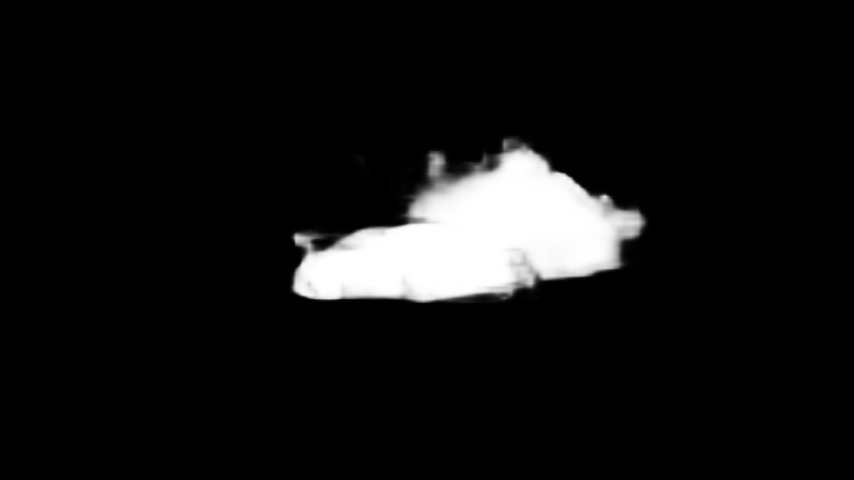}
\end{minipage}
\begin{minipage}[b]{0.08\linewidth}
  \centering
  \includegraphics[width=1\linewidth, height=1cm]{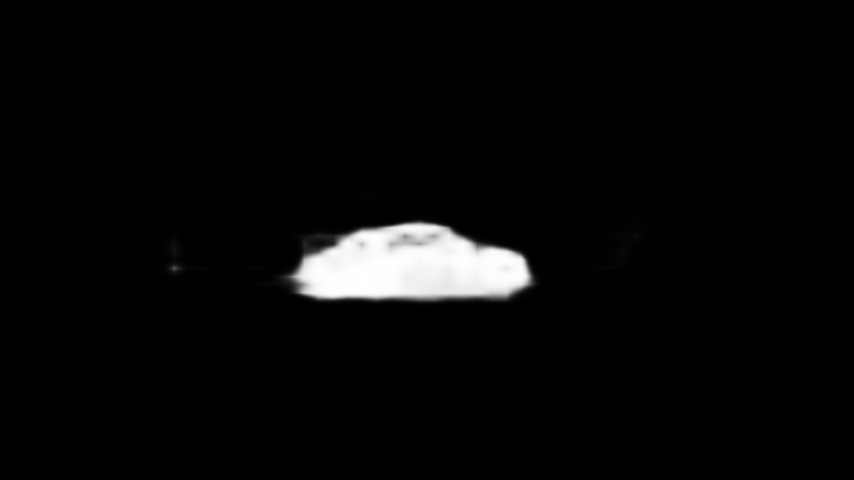}
\end{minipage}
\begin{minipage}[b]{0.08\linewidth}
  \centering
  \includegraphics[width=1\linewidth, height=1cm]{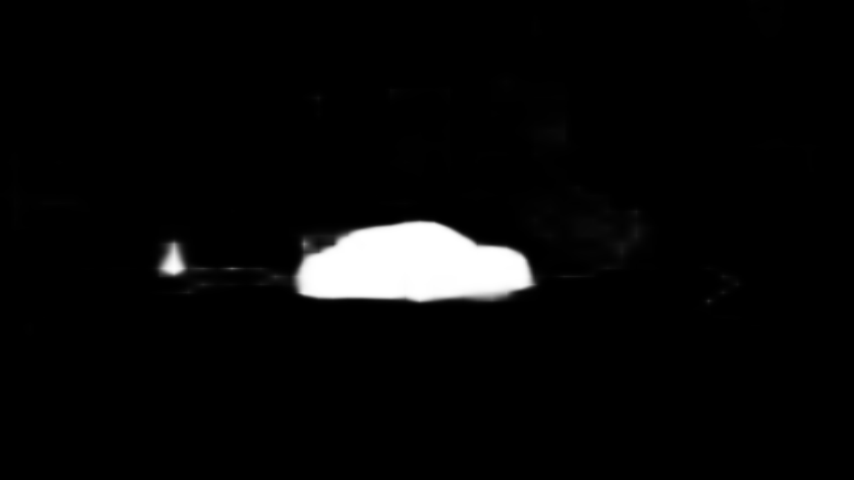}
\end{minipage}
\begin{minipage}[b]{0.08\linewidth}
  \centering
  \includegraphics[width=1\linewidth, height=1cm]{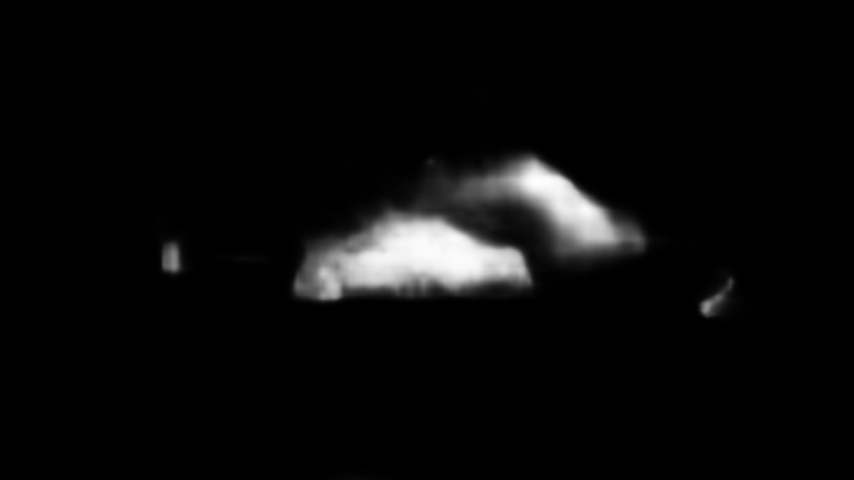}
\end{minipage}
\begin{minipage}[b]{0.08\linewidth}
  \centering
  \includegraphics[width=1\linewidth, height=1cm]{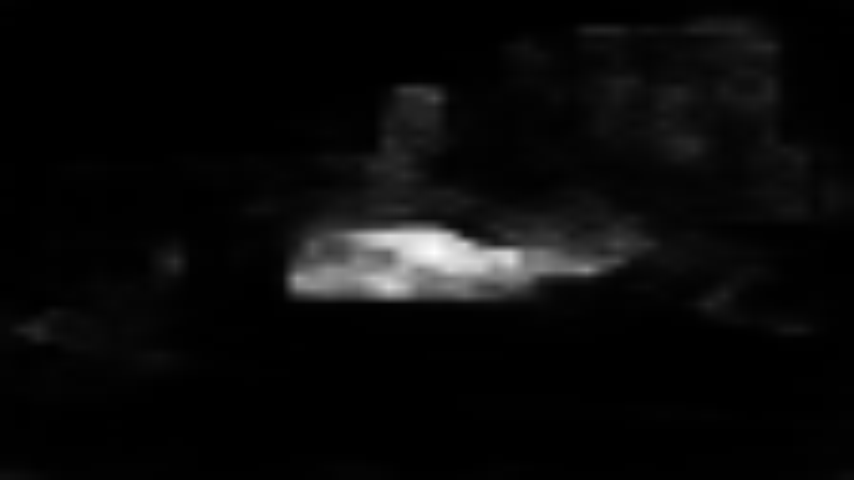}
\end{minipage}
\begin{minipage}[b]{0.08\linewidth}
  \centering
  \includegraphics[width=1\linewidth, height=1cm]{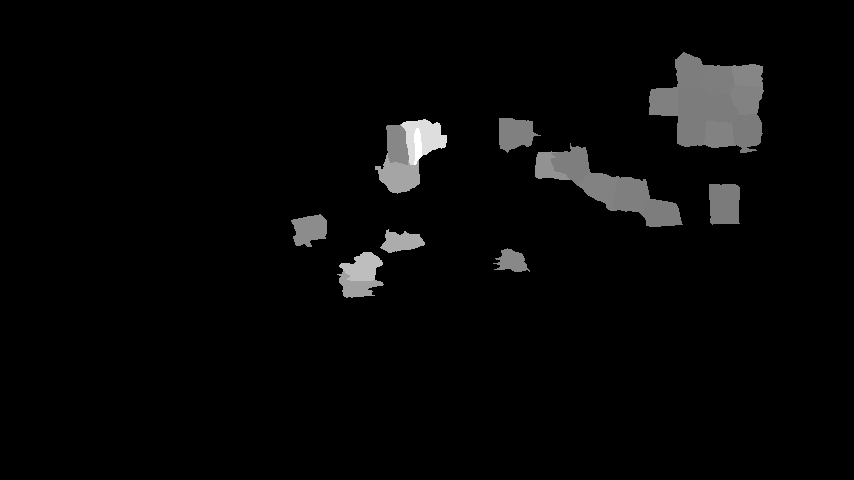}
\end{minipage}
\begin{minipage}[b]{0.08\linewidth}
  \centering
  \includegraphics[width=1\linewidth, height=1cm]{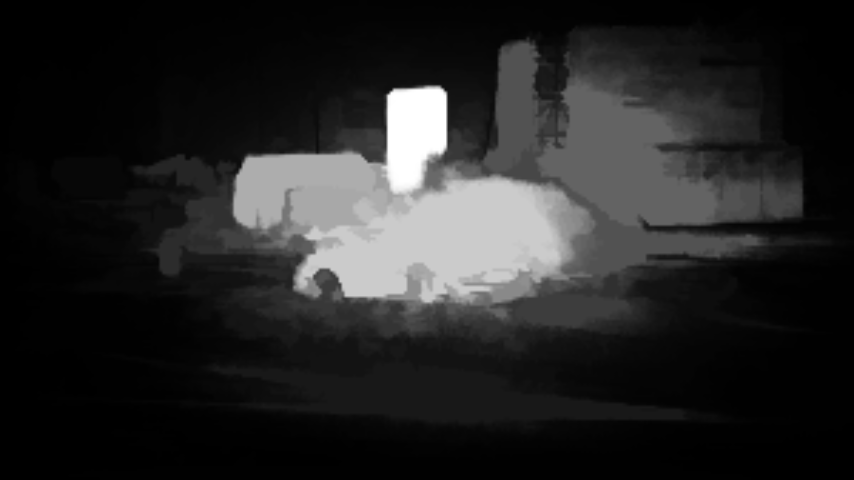}
\end{minipage}
\vspace{3pt}
\\
\begin{minipage}[b]{0.08\linewidth}
  \centering
  \includegraphics[width=1\linewidth, height=1cm]{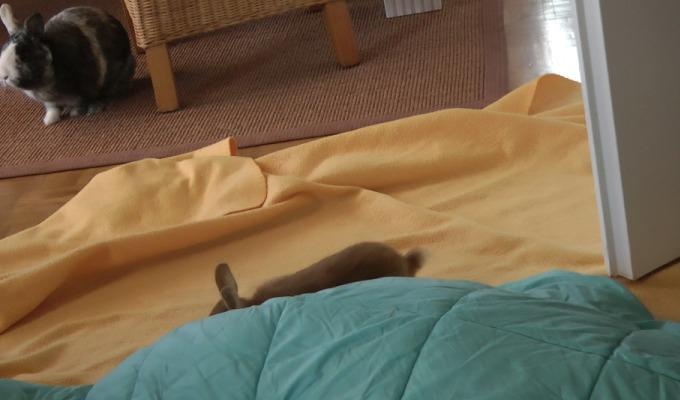}
\end{minipage}
\begin{minipage}[b]{0.08\linewidth}
  \centering
  \includegraphics[width=1\linewidth, height=1cm]{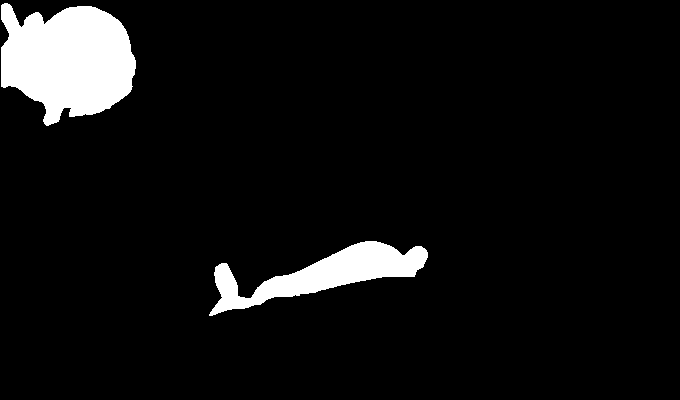}
\end{minipage}
\begin{minipage}[b]{0.08\linewidth}
  \centering
  \includegraphics[width=1\linewidth, height=1cm]{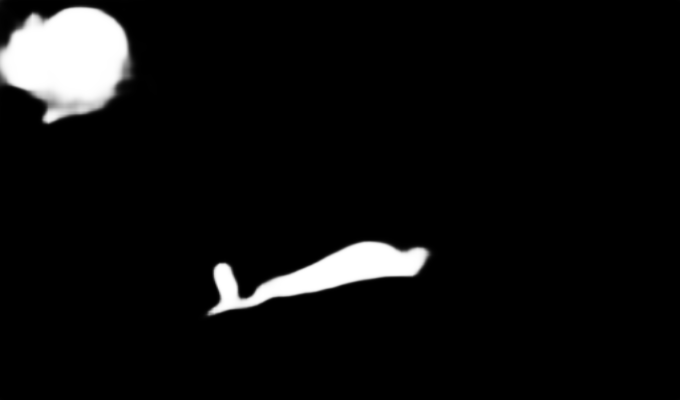}
\end{minipage}
\begin{minipage}[b]{0.08\linewidth}
  \centering
  \includegraphics[width=1\linewidth, height=1cm]{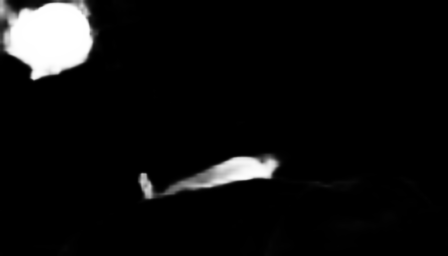}
\end{minipage}
\begin{minipage}[b]{0.08\linewidth}
  \centering
  \includegraphics[width=1\linewidth, height=1cm]{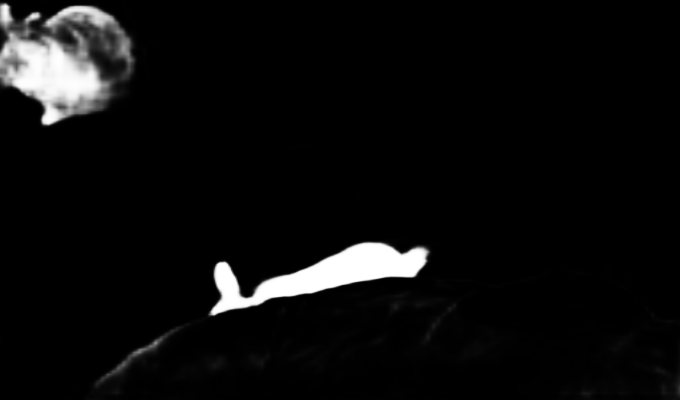}
\end{minipage}
\begin{minipage}[b]{0.08\linewidth}
  \centering
  \includegraphics[width=1\linewidth, height=1cm]{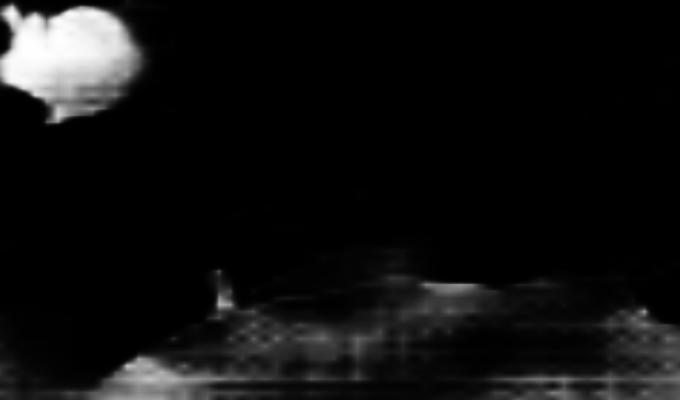}
\end{minipage}
\begin{minipage}[b]{0.08\linewidth}
  \centering
  \includegraphics[width=1\linewidth, height=1cm]{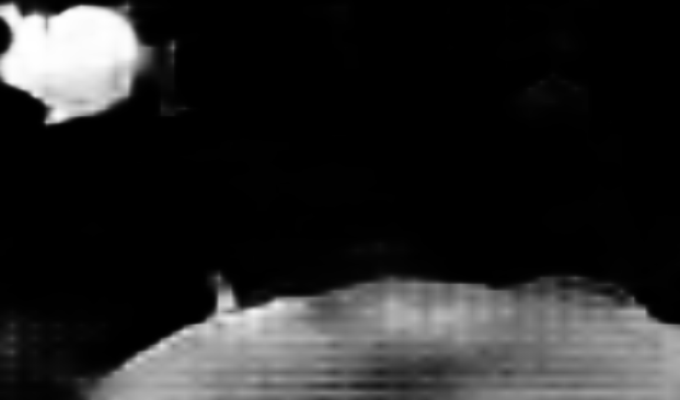}
\end{minipage}
\begin{minipage}[b]{0.08\linewidth}
  \centering
  \includegraphics[width=1\linewidth, height=1cm]{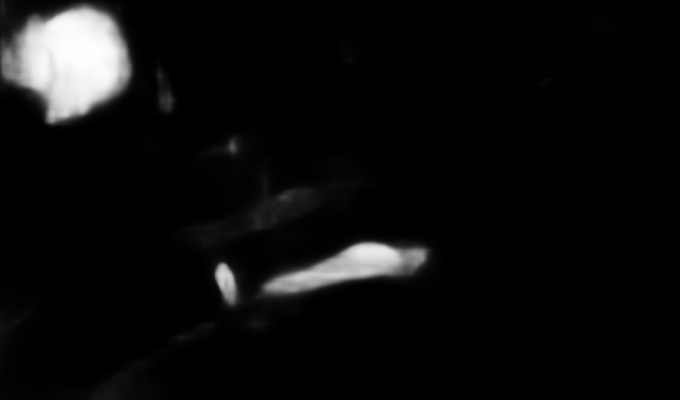}
\end{minipage}
\begin{minipage}[b]{0.08\linewidth}
  \centering
  \includegraphics[width=1\linewidth, height=1cm]{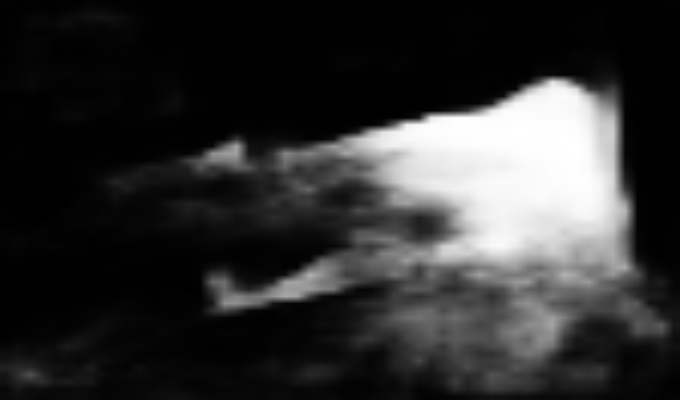}
\end{minipage}
\begin{minipage}[b]{0.08\linewidth}
  \centering
  \includegraphics[width=1\linewidth, height=1cm]{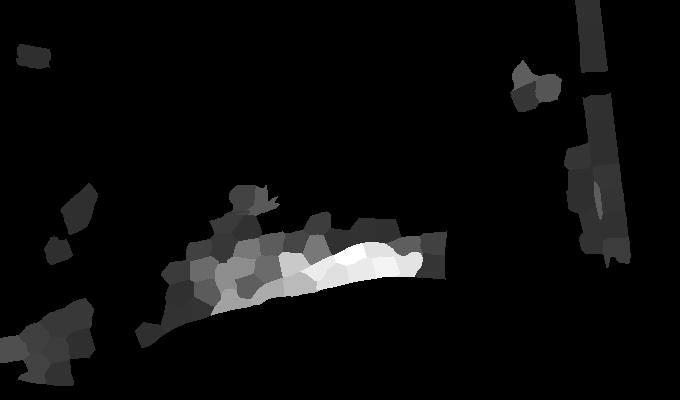}
\end{minipage}
\begin{minipage}[b]{0.08\linewidth}
  \centering
  \includegraphics[width=1\linewidth, height=1cm]{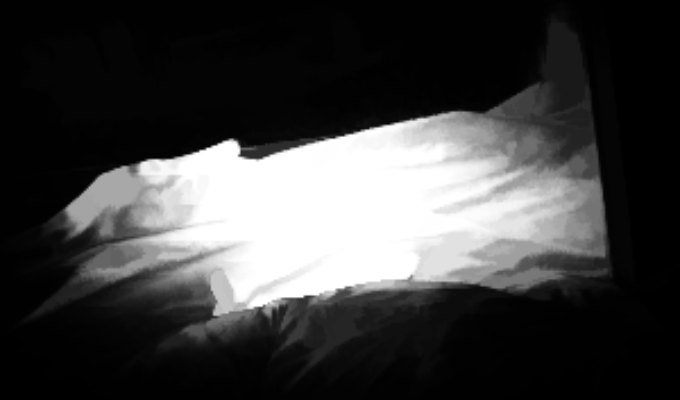}
\end{minipage}
\vspace{3pt}
\\
\begin{minipage}[b]{0.08\linewidth}
  \centering
  \includegraphics[width=1\linewidth, height=1cm]{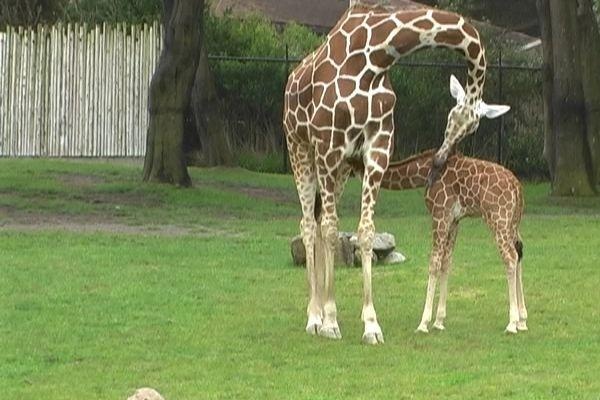}
\end{minipage}
\begin{minipage}[b]{0.08\linewidth}
  \centering
  \includegraphics[width=1\linewidth, height=1cm]{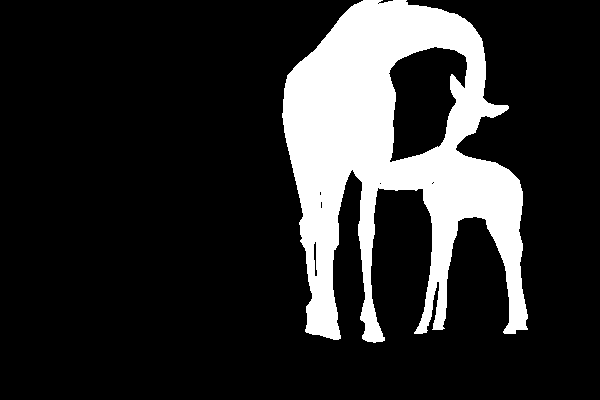}
\end{minipage}
\begin{minipage}[b]{0.08\linewidth}
  \centering
  \includegraphics[width=1\linewidth, height=1cm]{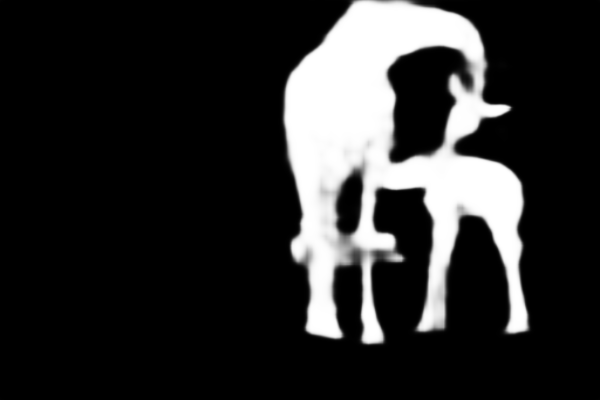}
\end{minipage}
\begin{minipage}[b]{0.08\linewidth}
  \centering
  \includegraphics[width=1\linewidth, height=1cm]{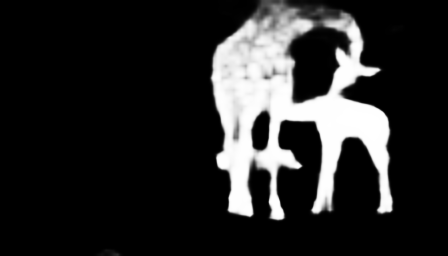}
\end{minipage}
\begin{minipage}[b]{0.08\linewidth}
  \centering
  \includegraphics[width=1\linewidth, height=1cm]{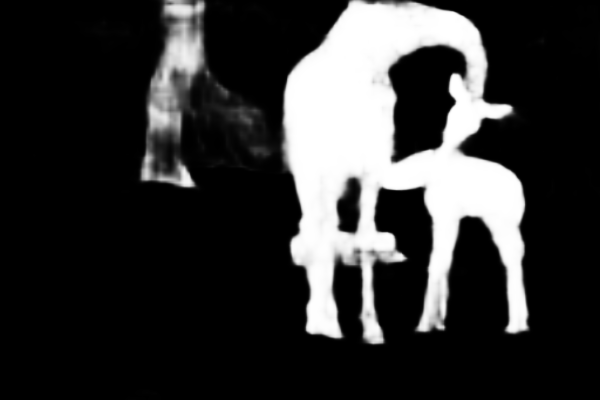}
\end{minipage}
\begin{minipage}[b]{0.08\linewidth}
  \centering
  \includegraphics[width=1\linewidth, height=1cm]{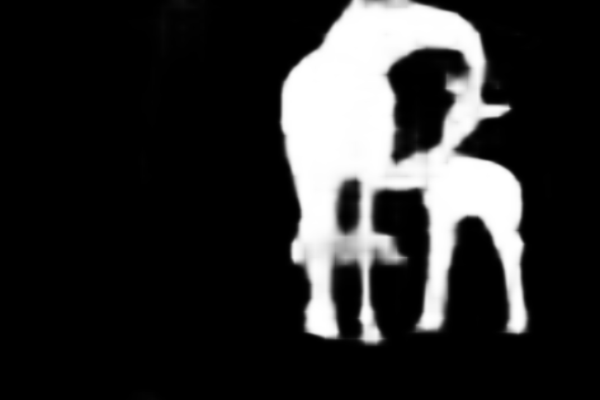}
\end{minipage}
\begin{minipage}[b]{0.08\linewidth}
  \centering
  \includegraphics[width=1\linewidth, height=1cm]{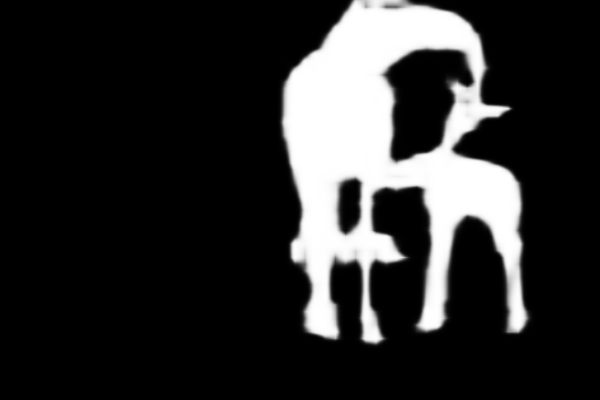}
\end{minipage}
\begin{minipage}[b]{0.08\linewidth}
  \centering
  \includegraphics[width=1\linewidth, height=1cm]{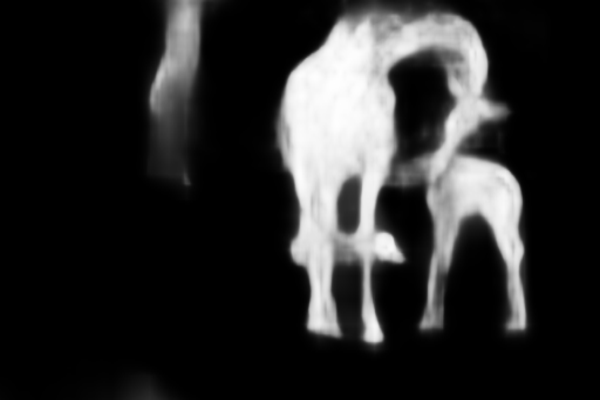}
\end{minipage}
\begin{minipage}[b]{0.08\linewidth}
  \centering
  \includegraphics[width=1\linewidth, height=1cm]{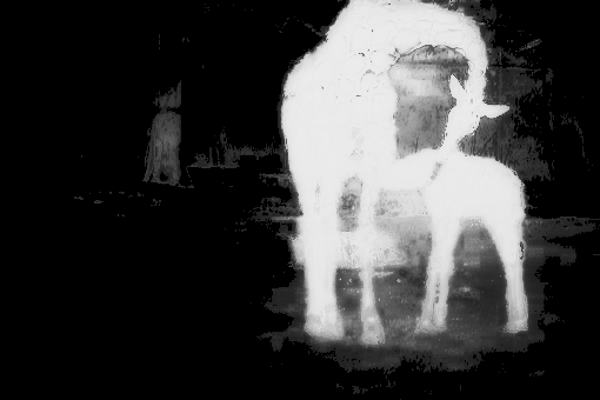}
\end{minipage}
\begin{minipage}[b]{0.08\linewidth}
  \centering
  \includegraphics[width=1\linewidth, height=1cm]{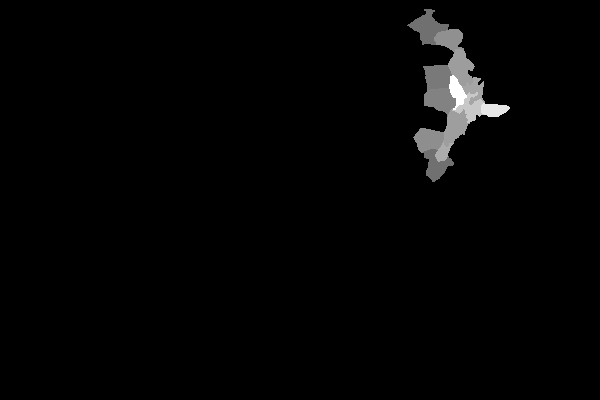}
\end{minipage}
\begin{minipage}[b]{0.08\linewidth}
  \centering
  \includegraphics[width=1\linewidth, height=1cm]{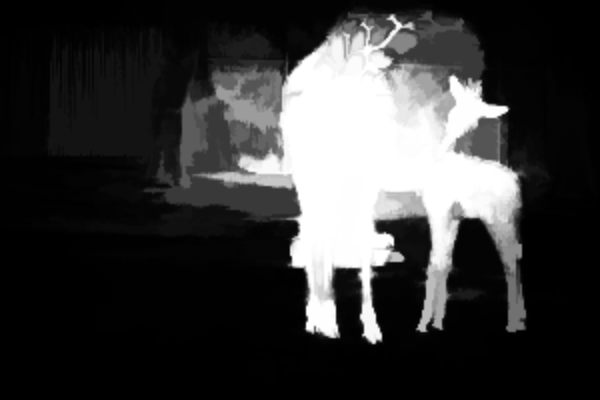}
\end{minipage}
\vspace{3pt}
\\
\begin{minipage}[b]{0.08\linewidth}
  \centering
  \includegraphics[width=1\linewidth, height=1cm]{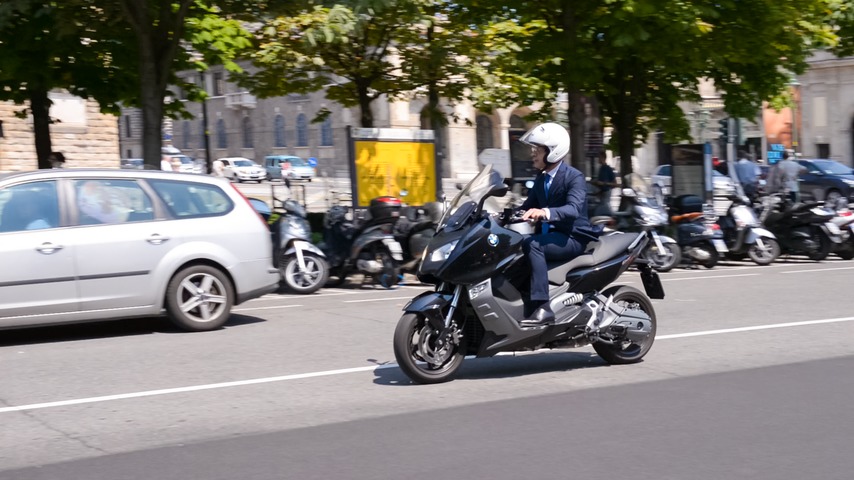}
\end{minipage}
\begin{minipage}[b]{0.08\linewidth}
  \centering
  \includegraphics[width=1\linewidth, height=1cm]{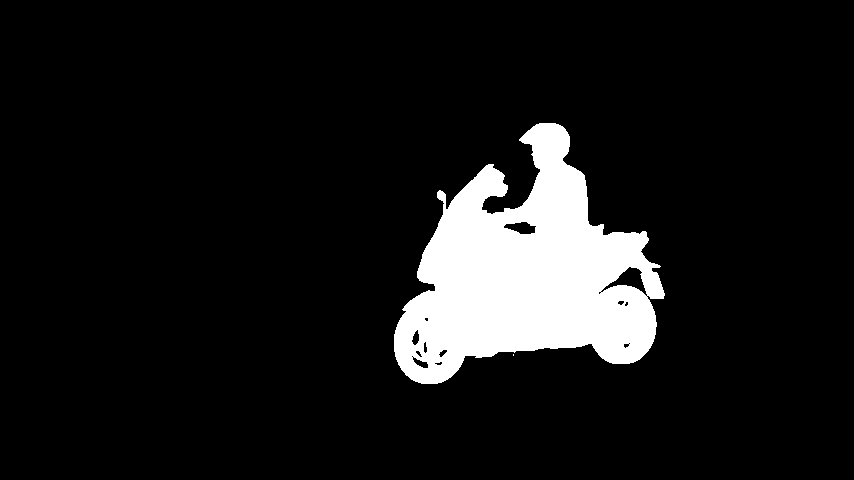}
\end{minipage}
\begin{minipage}[b]{0.08\linewidth}
  \centering
  \includegraphics[width=1\linewidth, height=1cm]{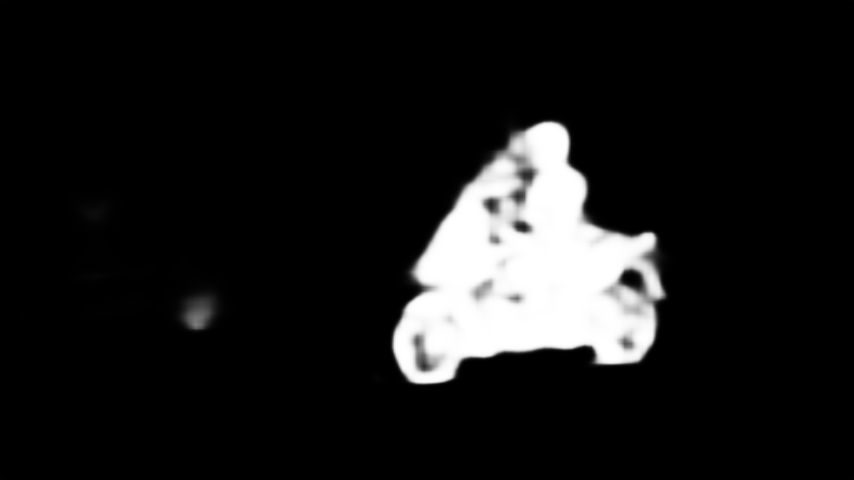}
\end{minipage}
\begin{minipage}[b]{0.08\linewidth}
  \centering
  \includegraphics[width=1\linewidth, height=1cm]{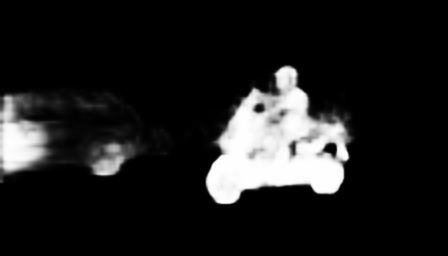}
\end{minipage}
\begin{minipage}[b]{0.08\linewidth}
  \centering
  \includegraphics[width=1\linewidth, height=1cm]{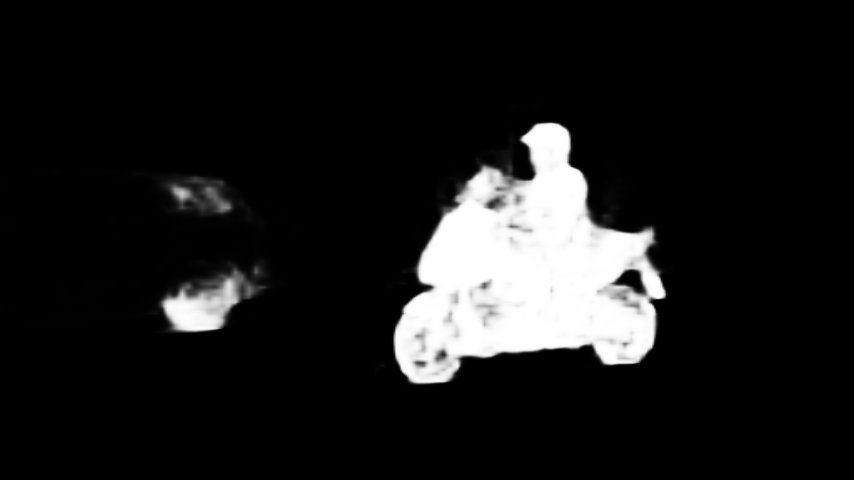}
\end{minipage}
\begin{minipage}[b]{0.08\linewidth}
  \centering
  \includegraphics[width=1\linewidth, height=1cm]{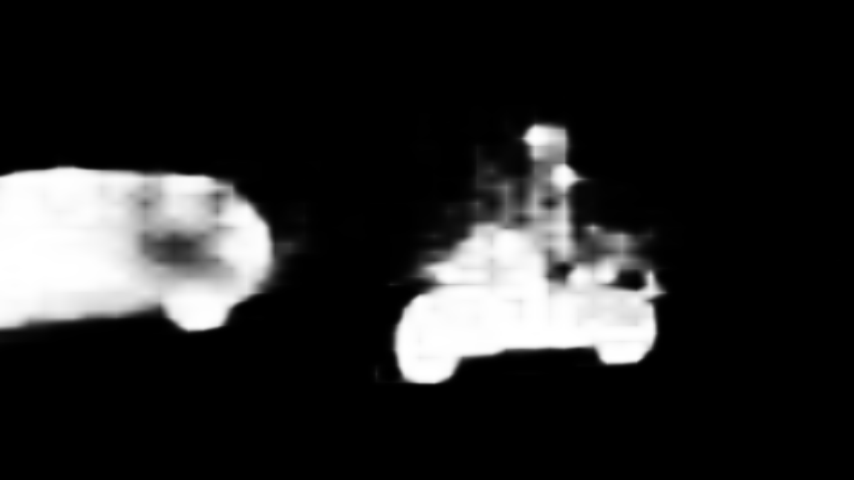}
\end{minipage}
\begin{minipage}[b]{0.08\linewidth}
  \centering
  \includegraphics[width=1\linewidth, height=1cm]{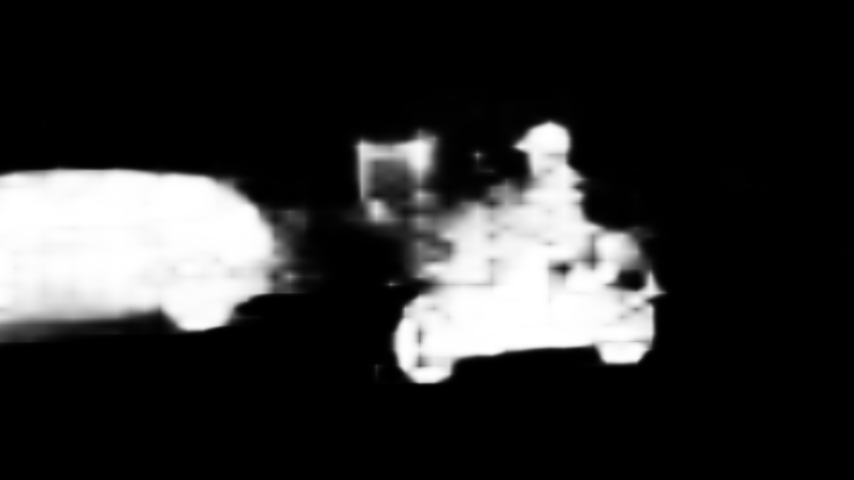}
\end{minipage}
\begin{minipage}[b]{0.08\linewidth}
  \centering
  \includegraphics[width=1\linewidth, height=1cm]{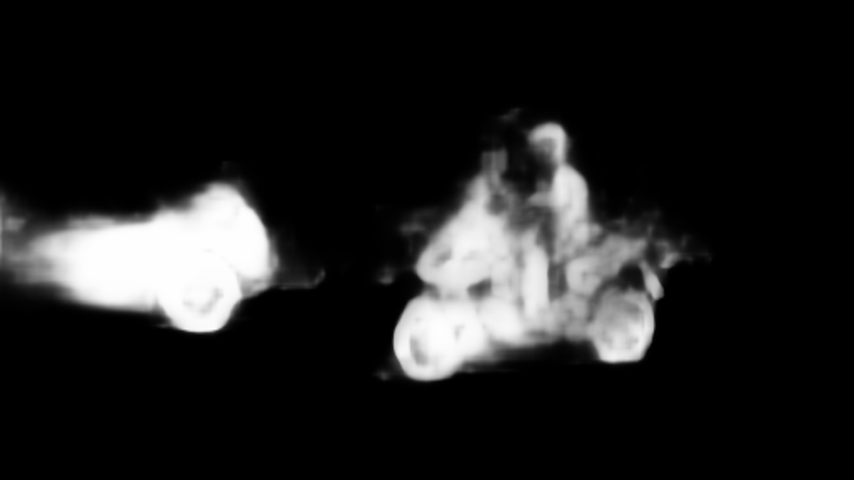}
\end{minipage}
\begin{minipage}[b]{0.08\linewidth}
  \centering
  \includegraphics[width=1\linewidth, height=1cm]{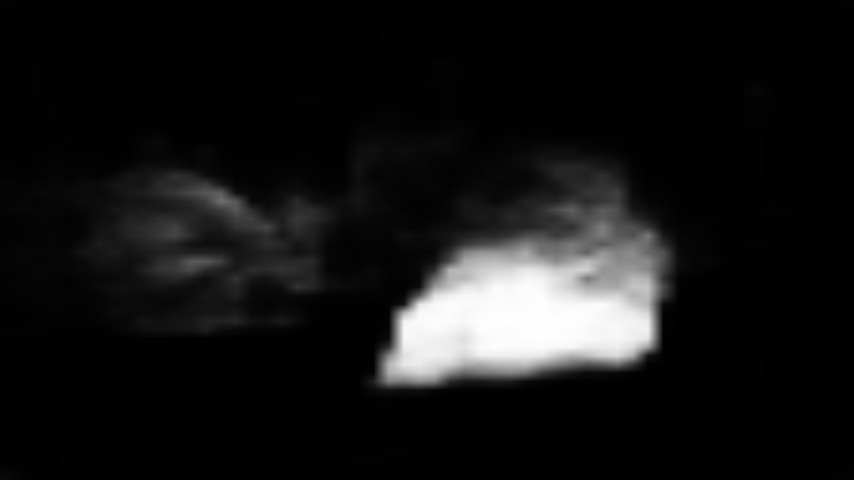}
\end{minipage}
\begin{minipage}[b]{0.08\linewidth}
  \centering
  \includegraphics[width=1\linewidth, height=1cm]{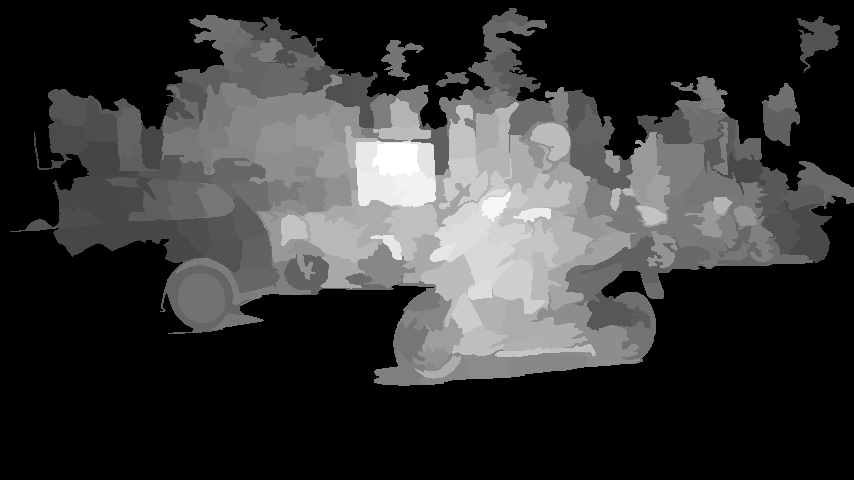}
\end{minipage}
\begin{minipage}[b]{0.08\linewidth}
  \centering
  \includegraphics[width=1\linewidth, height=1cm]{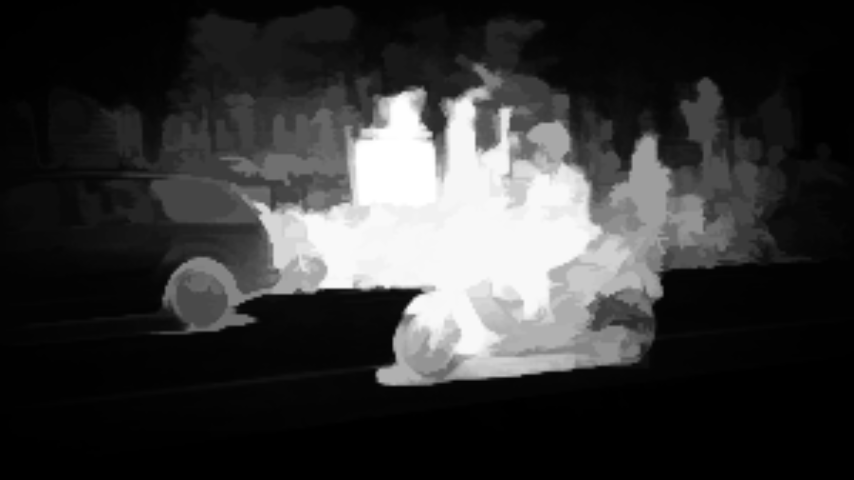}
\end{minipage}
\vspace{3pt}
\\
\begin{minipage}[b]{0.08\linewidth}
  \centering
  \includegraphics[width=1\linewidth, height=1cm]{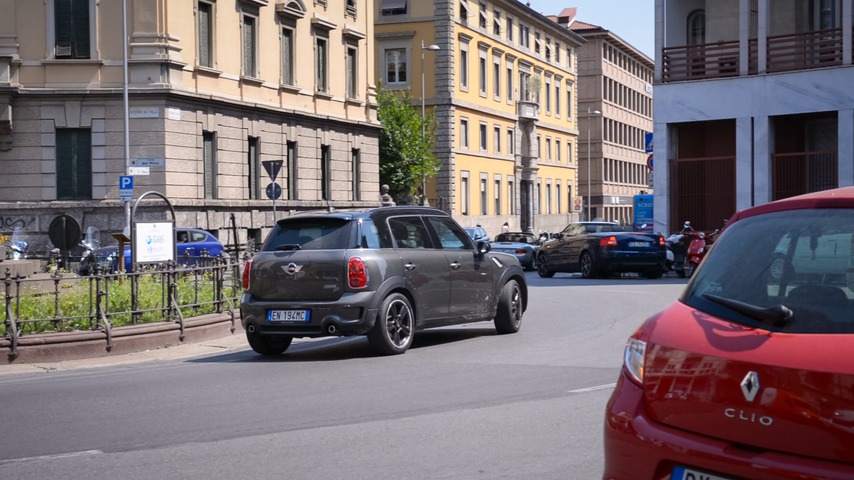}
  \centerline{Frame}
\end{minipage}
\begin{minipage}[b]{0.08\linewidth}
  \centering
  \includegraphics[width=1\linewidth, height=1cm]{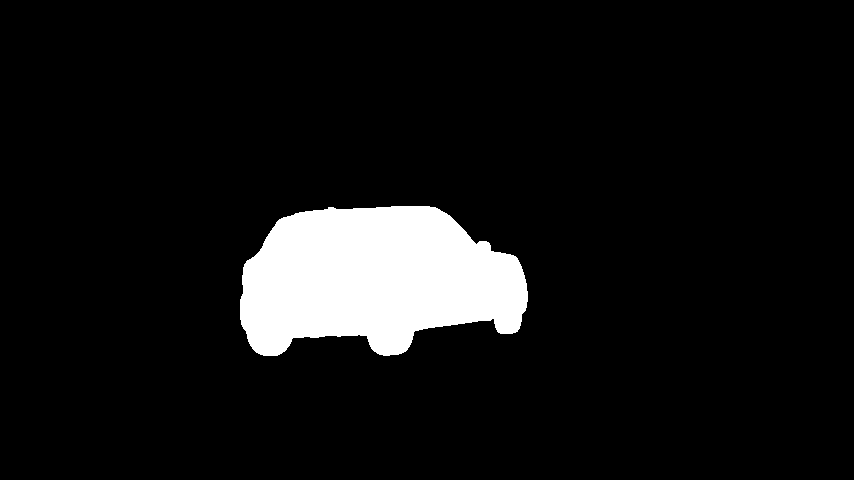}
  \centerline{GT}
\end{minipage}
\begin{minipage}[b]{0.08\linewidth}
  \centering
  \includegraphics[width=1\linewidth, height=1cm]{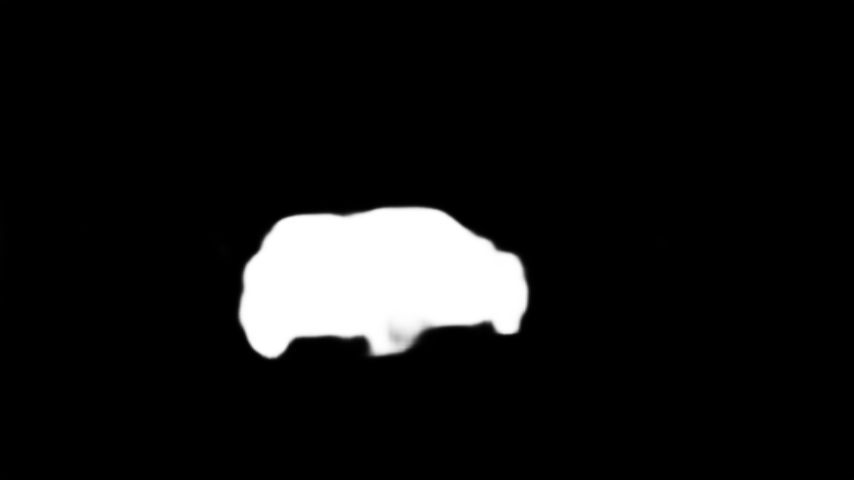}
  \centerline{Ours}
\end{minipage}
\begin{minipage}[b]{0.08\linewidth}
  \centering
  \includegraphics[width=1\linewidth, height=1cm]{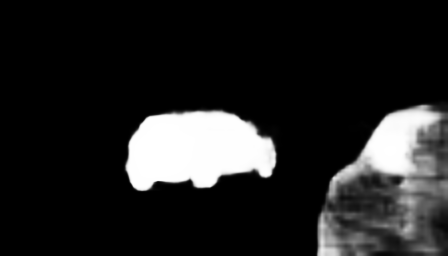}
  \centerline{PCSA}
\end{minipage}
\begin{minipage}[b]{0.08\linewidth}
  \centering
  \includegraphics[width=1\linewidth, height=1cm]{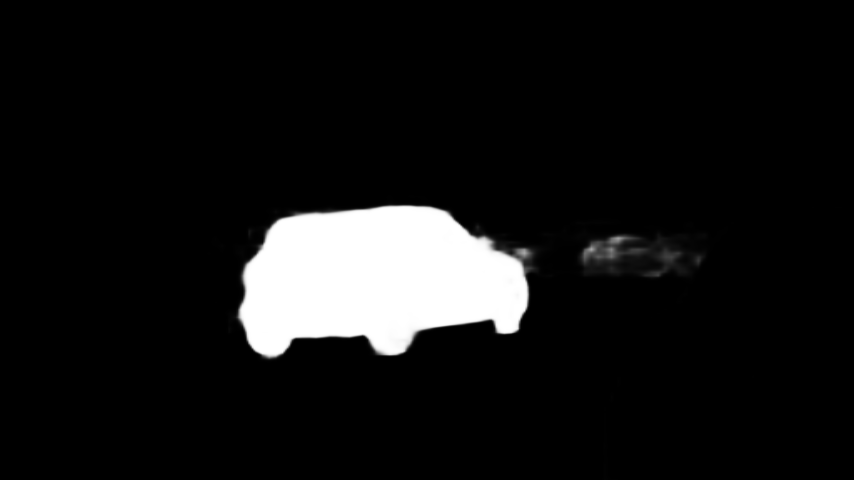}
  \centerline{MGA}
\end{minipage}
\begin{minipage}[b]{0.08\linewidth}
  \centering
  \includegraphics[width=1\linewidth, height=1cm]{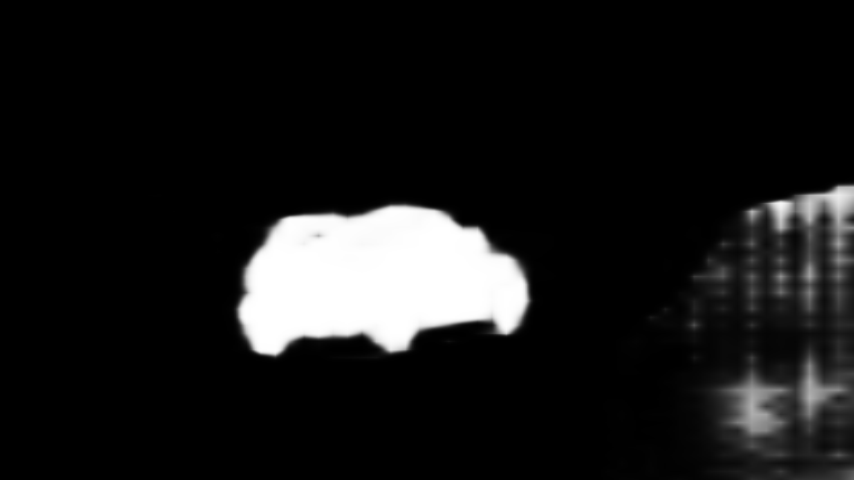}
  \centerline{SSAV}
\end{minipage}
\begin{minipage}[b]{0.08\linewidth}
  \centering
  \includegraphics[width=1\linewidth, height=1cm]{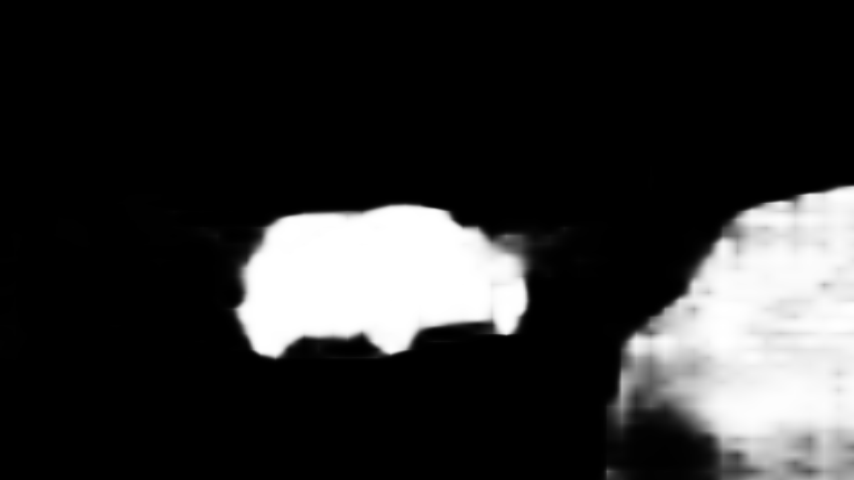}
  \centerline{PDB}
\end{minipage}
\begin{minipage}[b]{0.08\linewidth}
  \centering
  \includegraphics[width=1\linewidth, height=1cm]{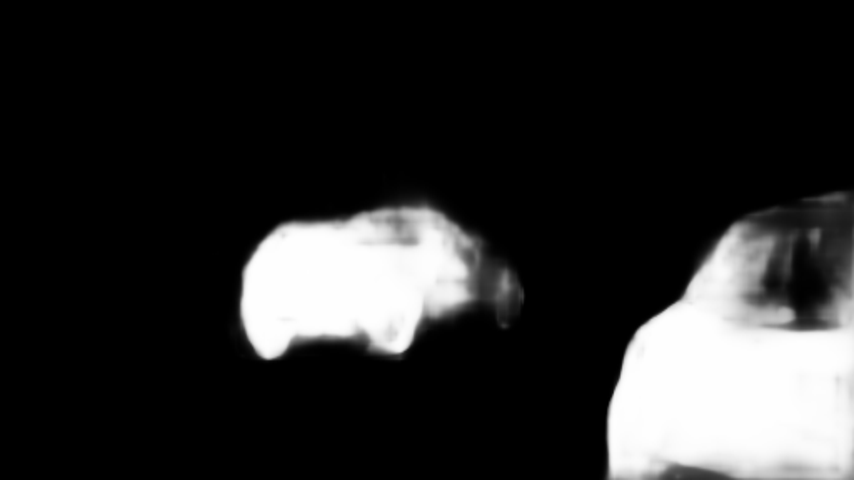}
  \centerline{FGRNE}
\end{minipage}
\begin{minipage}[b]{0.08\linewidth}
  \centering
  \includegraphics[width=1\linewidth, height=1cm]{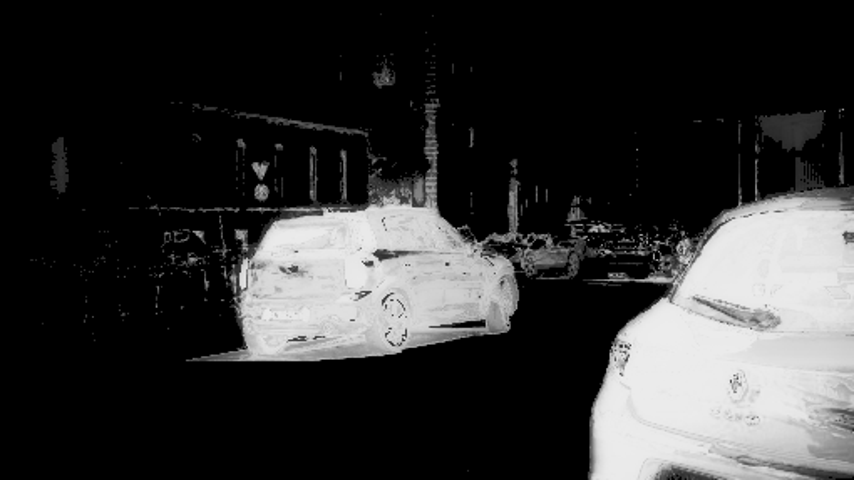}
  \centerline{FCNS}
\end{minipage}
\begin{minipage}[b]{0.08\linewidth}
  \centering
  \includegraphics[width=1\linewidth, height=1cm]{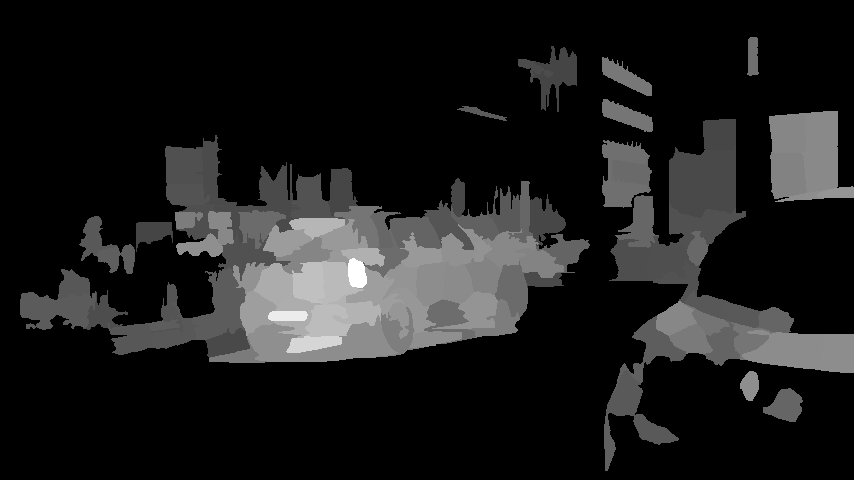}
  \centerline{STBP}
\end{minipage}
\begin{minipage}[b]{0.08\linewidth}
  \centering
  \includegraphics[width=1\linewidth, height=1cm]{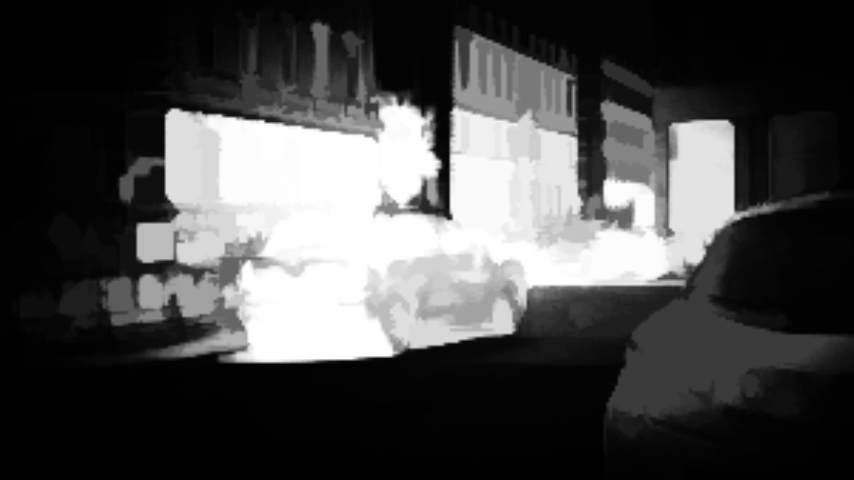}
  \centerline{MBD}
\end{minipage}
\vspace{3pt}
\\
\caption{Visual comparison with state-of-the-art video salient object detection methods.}
\label{compare_pic}
\end{figure*}
The proposed model is built based on pytorch repository. We initialize ResNet34 in symmetric spatial and temporal saliency network with the weights of ResNet34 pretrained on ImageNet~\cite{DBLP:journals/ijcv/RussakovskyDSKS15}. And we trained our model on the training set of one image dataset DUTS~\cite{8099887} and three video dataset DAVIS~\cite{7780454}, FBMS~\cite{Bro10c} and VOS~\cite{5459276}. Considering that the input of our network includes optical flow images, for the image dataset, the input optical flow images are filled with zeros, which means that these frames do not contain motion information.

During the training process, we first resize all input images to $512\times512$. The Adam~\cite{DBLP:journals/corr/KingmaB14} optimizer with initial learning rate 5e-5, a weight decay of 0.0005 and an momentum of 0.9 is used to train our model. And the training batch size is set to 8. The proposed model is easy to train where all parameters are trained by a simple one-step end-to-end strategy except for the backbone ResNet34 feature extractor. Experiments are performed on a workstation with an NVIDIA GTX 1080Ti GPU and a 2.1 GHz Intel CPU.
\begin{figure*}[t]
\centering
\begin{minipage}[b]{0.310\linewidth}
  \centering
  \includegraphics[width=1\linewidth, ]{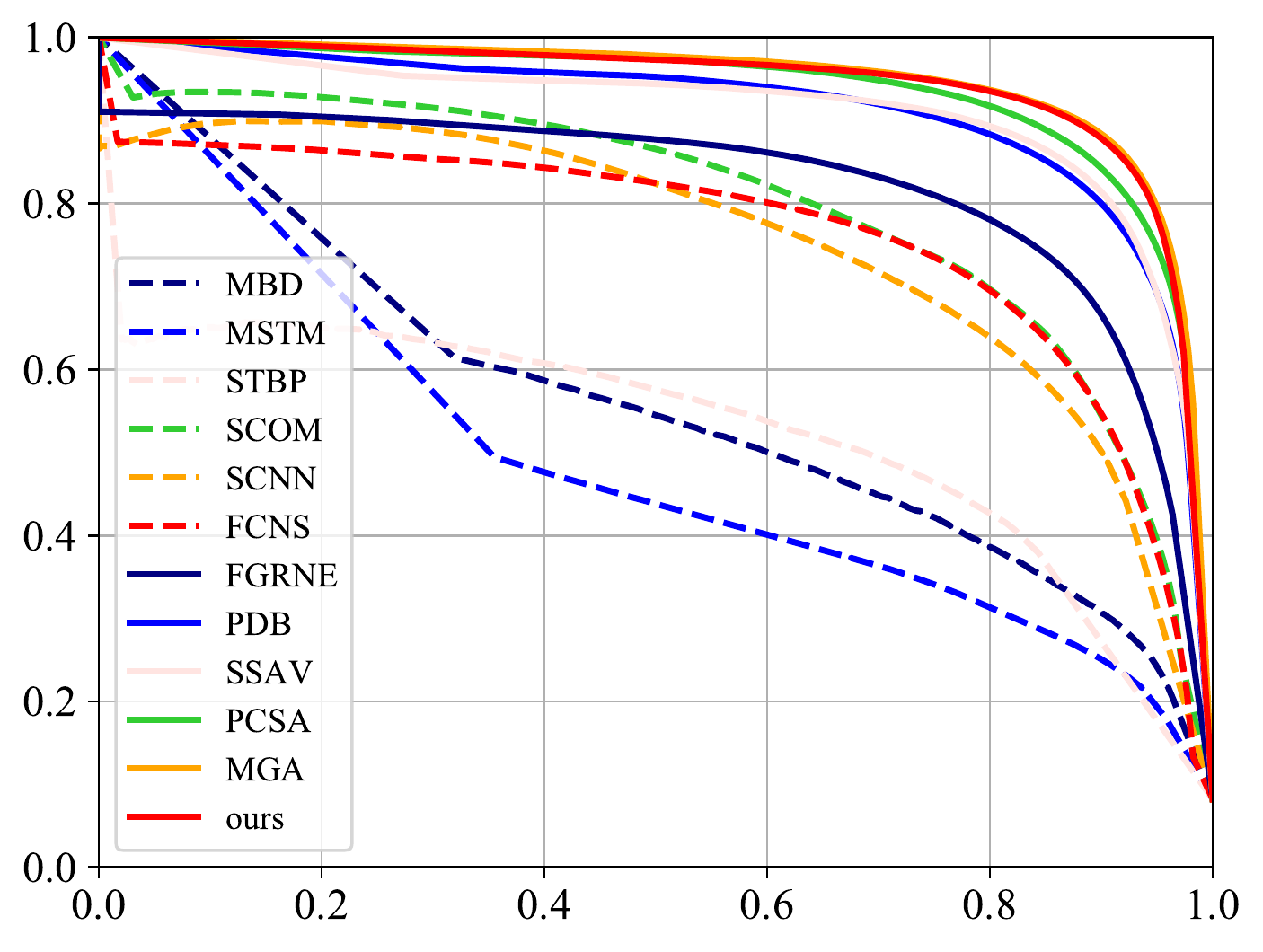}
  \centerline{DAVIS}
\end{minipage}
\begin{minipage}[b]{0.31\linewidth}
  \centering
  \includegraphics[width=1\linewidth, ]{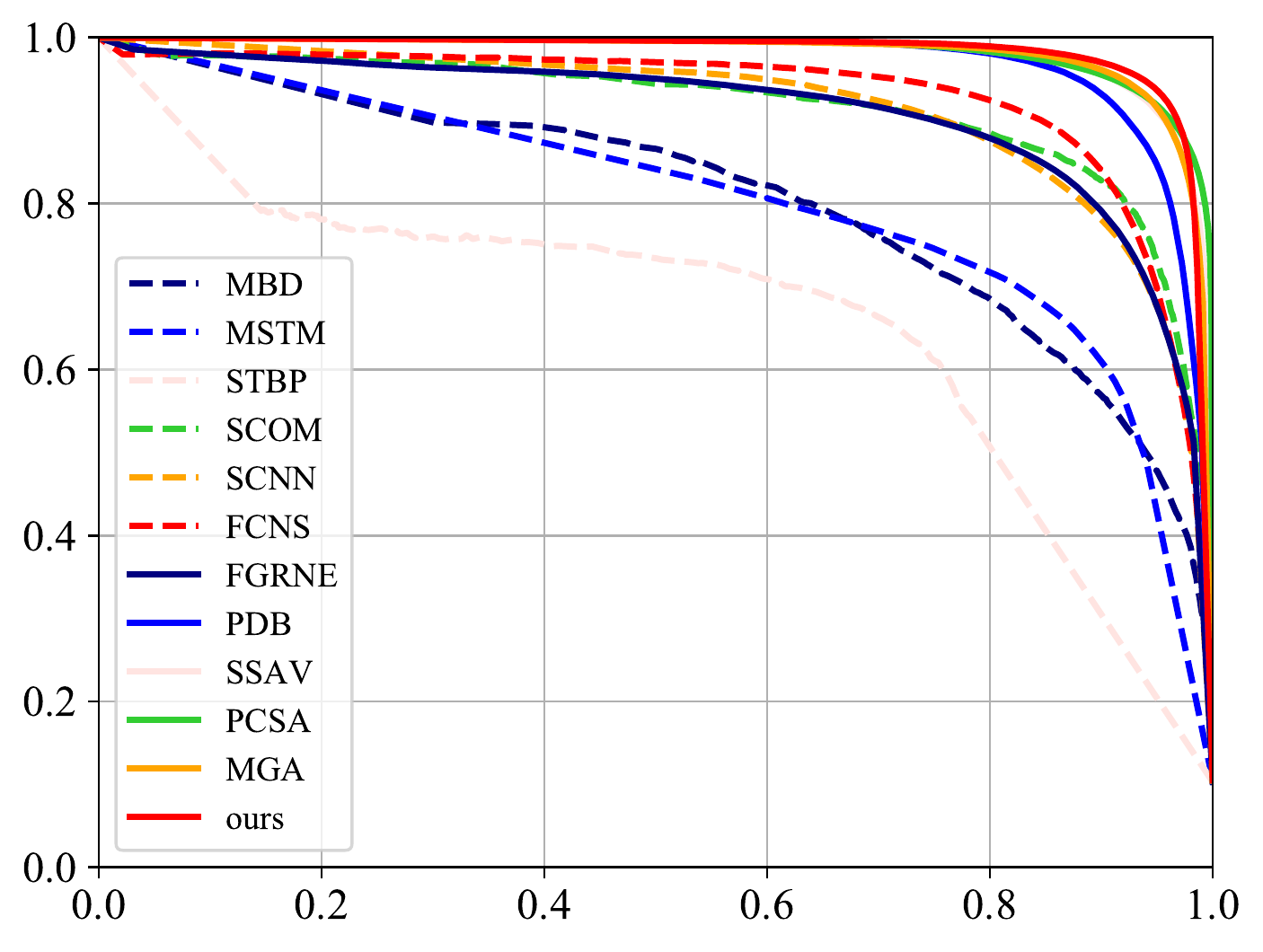}
  \centerline{ViSal}
\end{minipage}
\begin{minipage}[b]{0.31\linewidth}
  \centering
  \includegraphics[width=1\linewidth, ]{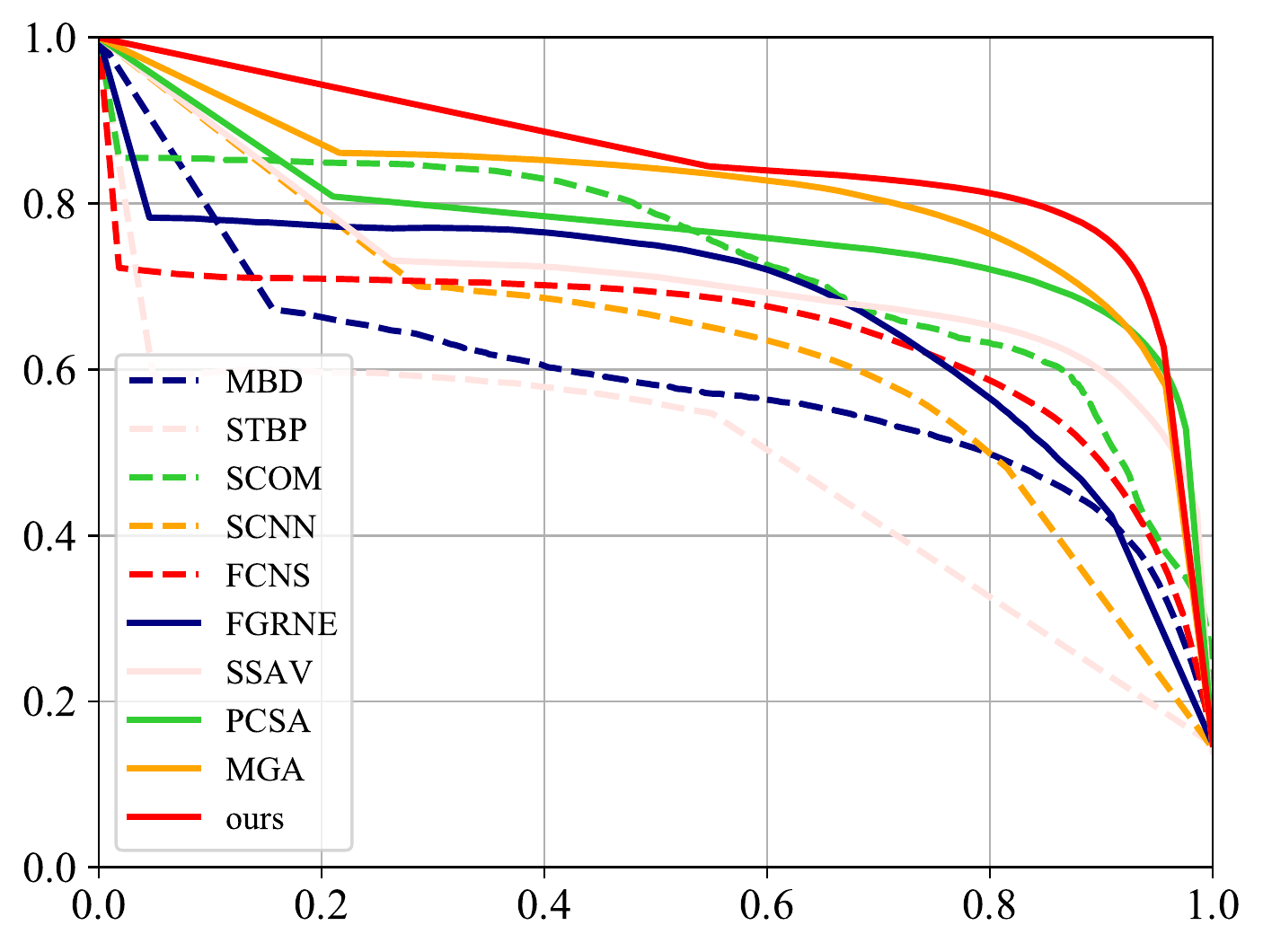}
  \centerline{VOS}
\end{minipage}
\vspace{3pt}
\\
\begin{minipage}[b]{0.31\linewidth}
  \centering
  \includegraphics[width=1\linewidth, ]{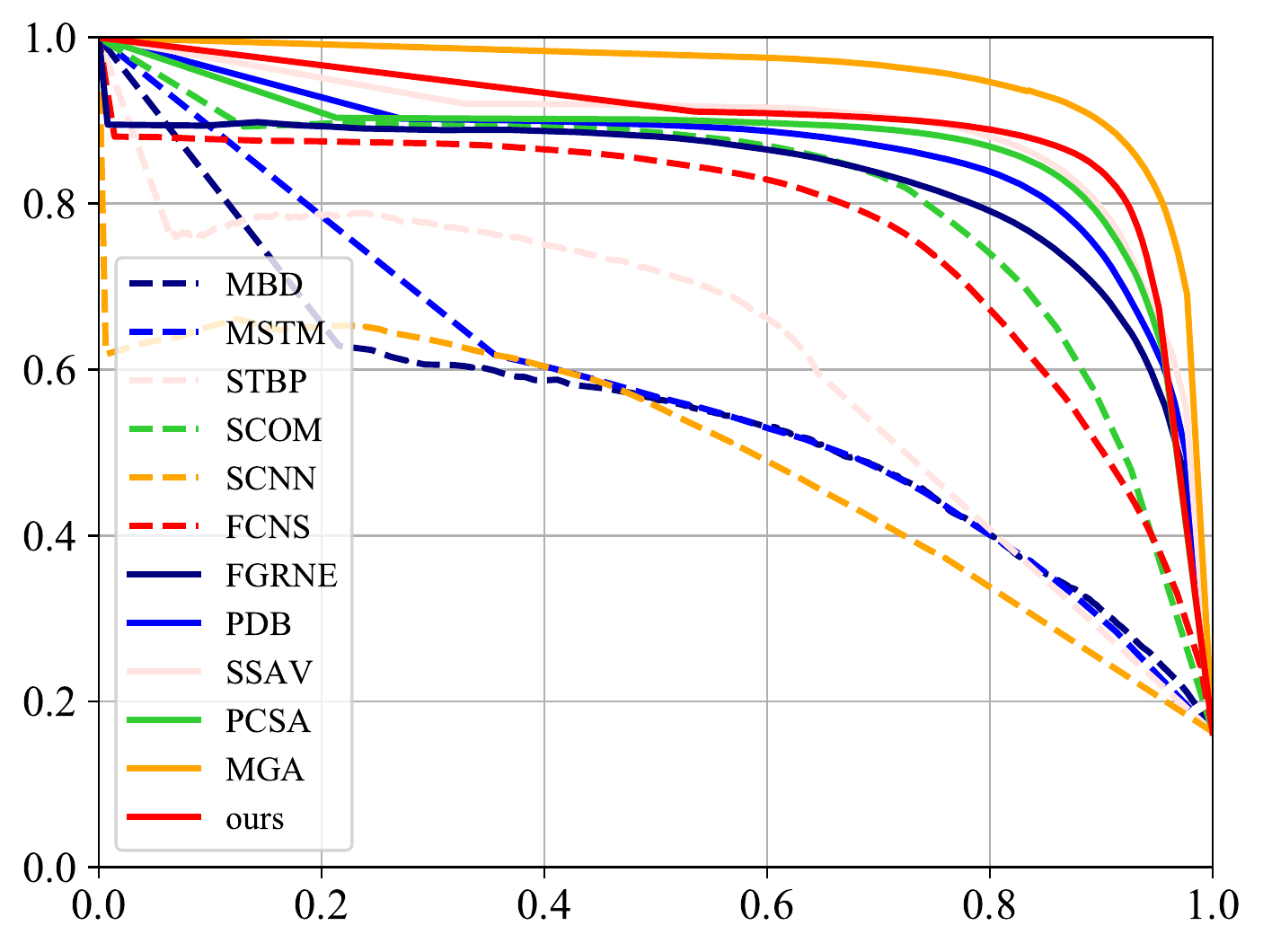}
  \centerline{FBMS}
\end{minipage}
\begin{minipage}[b]{0.31\linewidth}
  \centering
  \includegraphics[width=1\linewidth, ]{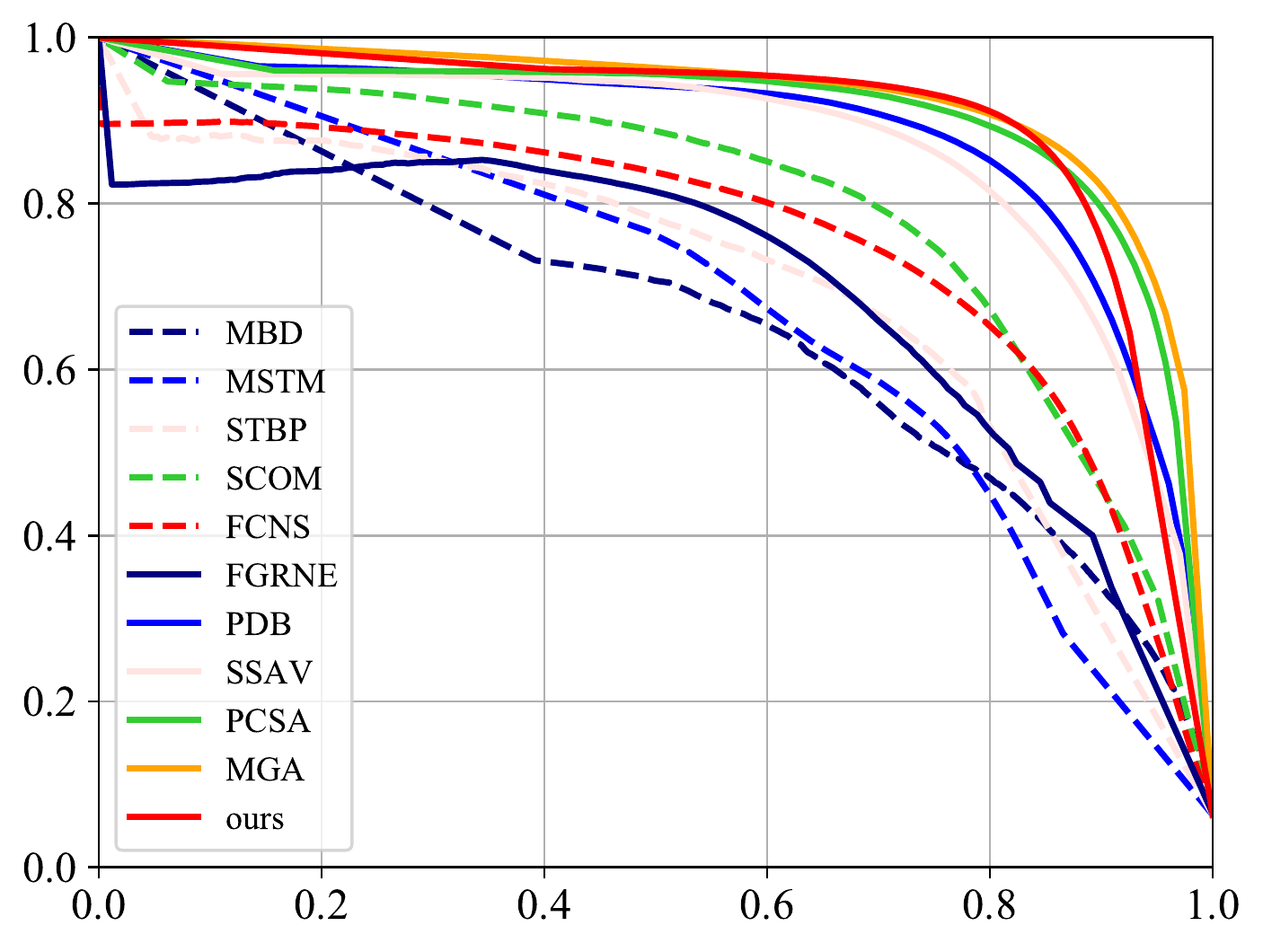}
  \centerline{SegV2}
\end{minipage}
\vspace{3pt}
\\
\caption{Precision (vertical axis) recall (horizontal axis) curves on five popular video salient object datasets.}
\label{PRcurve}
\end{figure*}

\subsection{Datasets and Evaluation Criteria}
\subsubsection{Dataset}
We evaluate our model on five public video salient object detection benchmark datasets, including DAVIS, FBMS, ViSal~\cite{7164324}, SegTrackV2~\cite{6751383}, and VOS. DAVIS is a frequently used dataset, which contains 50 videos with totally 3,455 high-quality pixel-wise annotation frames. FBMS is a dataset containing 59 videos with 720 sparsely annotated frames. ViSal is a dataset only used for test containing 19 videos with 193 pixel-wise annotation frames. VOS is a large-scale dataset with totally 200 videos and 7,467 pixel-wise annotated frames. SegTrackV2 is an early adopted dataset with 14 videos and 1,065 annotated frames. In this paper, the test is carried out on the testing part of DAVIS, FBMS, VOS and the whole datasets of ViSal and SegTrackV2.
\subsubsection{Evaluation Criteria}
Mean absolute error (MAE), structure-measure (S-m)~\cite{8237749}, max F-measure (maxF)~\cite{5206596} and precision-recall (PR) curves are adopted as the evaluation metrics. MAE is defined as the average pixel-wise difference between the binary ground truth and the saliency prediction map. F-measure is defined as
\begin{equation}
F_\beta=\frac{(1+\beta^2)\cdot Precision\cdot Recall}{\beta^2\cdot Precision+Recall},
\label{F-measure}
\end{equation}
where the $\beta^2$ is always set to 0.3 in salient object detection task. We report the maximum F-measure (maxF)  computed from the PR curve. The S-measure is a newly proposed measurement focusing on the structural similarity between saliency ground truth and prediction map. For the PR curves, we firstly threshold the output saliency map into a binary map and match it with the ground truth. And then, by applying different thresholds to the saliency map, a series of precision and recall pairs are obtained to draw the PR curve. 
\begin{figure}[]
\small
\centering
\includegraphics[width=0.4\textwidth, height=5cm]{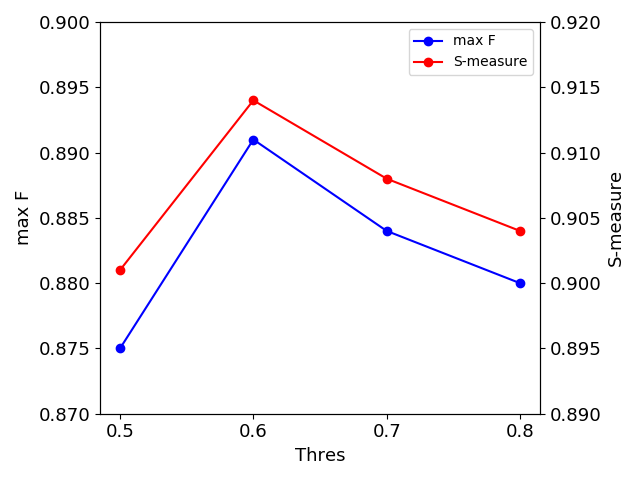}
\caption{Sensitive analysis on non-linear threshold $\tau$. The performance is tested on DAVIS dataset.}
\label{Thres}
\end{figure}

\subsection{Performance Comparison}
Our proposed method is compared with 11 state-of-the-art methods, including 4 conventional methods: MBD~\cite{7410522}, MSTM~\cite{7780625}, STBP~\cite{7752954}, SCOM~\cite{8307461} and 7 learning based methods: SCNN~\cite{8419765}, FCNS~\cite{8047320}, FGRNE~\cite{8578440}, PDB~\cite{inbook}, SSAV~\cite{8954050}, MGA~\cite{9010060}, PCSA~\cite{DBLP:conf/aaai/GuWW0CL20}. We use the evaluation code provided by~\cite{8954050} for fair comparison. The maxF, S-measure, and MAE results are listed in Table~\ref{comparison} and the PR Curves are shown in Fig~\ref{PRcurve}. As shown by Table~\ref{comparison}, our method achieves the best performance on DAVIS, ViSal, and VOS and the second best performance on FBMS and SegV2. For DAVIS, the proposed method outperforms the second best model MGA by 0.4\% in S-measure and 18.2\% in MAE. For ViSal, our algorithm surpasses the second best model PCSA by 1.1\% in maxF and 0.3\% in S-measure. As for VOS, the proposed method significantly outperforms the second best method PCSA by 7.2\% in maxF and 3.4\% in S-measure. And it achieves a significant performance gap than state-of-the-art algorithms in terms of PR curve. Our method also obtains the highest maxF and second highest S-measure on SegV2. Since we don't train our model on ViSal and SegV2, the competitive and even superior performance on these two datasets validates the generalization capability of the proposed algorithm.

Fig.~\ref{compare_pic} gives visual comparison of saliency maps with state-of-the-art algorithms. Five diverse yet challenging cases are included. From the top to bottom are cases with complex background (i.e., the crowd and partial occlusion in the 1st and 2nd rows), poor static saliency cue (i.e., low color contrast in the 3rd and 4th rows), poor dynamic saliency cue (i.e., moving camera in 5th and 6th rows), multiple salient objects (i.e., two rabbits and two giraffes in 7th and 8th rows), and distracting non-salient object (i.e., vehicles on road in 9th and 10th rows). It is obvious that our method can accurately segment salient objects in all these cases, demonstrating its efficiency and generalization capability.

As shown in Table~\ref{comparison}, Fig.~\ref{compare_pic} and Fig.~\ref{PRcurve},  MGA is the most competitive state-of-the-art algorithm, which also utilized optical flow to extract motion information and a two-branch structure to get complementary spatiotemporal features. However, MGA adopts a more complex backbone network of ResNet101 while the proposed algorithm adopts ResNet34 to extract spatial information. As deeper and more complicated structure always obtain better feature extraction at the cost of higher computational complexity and generally higher difficulty to converge, the competitive and even higher SOD performance of the proposed algorithm validates the effectiveness of the proposed dynamic fusion of spatiotemporal information.

\subsection{Sensitivity of Hyper-parameter}

Note that the proposed algorithm automatically generates the dynamic weights based on spatial and temporal features, so the only key hyper-parameter of our algorithm is the non-linear threshold $\tau$ in CAA. The non-linear cross thresholding is designed to suppress the inference from poor saliency cues where $\tau$ affects the degree of suppression. Small $\tau$ may lead to over suppression of informative saliency cue while large $\tau$ may weaken the suppression ability on distracting information of non-linear cross thresholding. We train the proposed model with different thresholds and the results are shown in Fig.~\ref{Thres}. As $\tau=0.6$ achieves the best performance, it is set to 0.6 empirically throughout this paper.

\begin{table}[]
\caption{Ablation study of the proposed DS-Net. The performance are tested on large-scale dataset DAVIS.}
\label{ablation}
\centering
\resizebox{.95\columnwidth}{!}{
\begin{tabular}{cccccccc}
\toprule[.15em]
                 & M1 & M2 & M3 & M4 & M5 & M6 & M7(Ours) \\ \midrule[.15em]
Spatial         &$\surd$    &     &     &     &   &   &   \\ 
Temporal              &    & $\surd$   &   &   &    &  &    \\ 
Symmetric              &    &    & $\surd$   & $\surd$   &  $\surd$ &  $\surd$ &  $\surd$  \\ 
\hline
DWG        &    &    &    &$\surd$    & $\surd$   & $\surd$ & $\surd$\\
CAA        &    &    &    &   &$\surd$  &$\surd$    & $\surd$   \\ 
Attention        &    &    &    &    &    &$\surd$    & $\surd$   \\ 
Multi-supervision &    &    &    &    &    &    &$\surd$    \\ \midrule[.15em]
\multicolumn{1}{c}{maxF $\uparrow$} & \multicolumn{1}{l}{0.815} & \multicolumn{1}{l}{0.772} & \multicolumn{1}{c}{0.844} & 0.861 & 0.877 & 0.882 & \textbf{0.891} \\ \cmidrule{1-8}
\multicolumn{1}{c}{S $\uparrow$}   & \multicolumn{1}{l}{0.867}  & \multicolumn{1}{l}{0.811} & \multicolumn{1}{c}{0.886} & 0.897 & 0.907 & 0.910 & \textbf{0.914} \\ \bottomrule[.15em]
\end{tabular}}
\end{table}
\subsection{Ablation Study}
In this section, we first explore the effectiveness of each key module of the propose algorithm in an incremental manner and then specifically study the effectiveness of each key module's structure in detail. For fair comparison, we propose a baseline model where the spatial and temporal features from the same scale of symmetric feature extraction branch are directly summed up and fed into finer scale. Upon this baseline model, the module of DWG, CAA, Spatial Attention, and Multi-supervision are gradually incorporated to obtain the proposed DS-Net. The maxF and S-measure performance on the DAVIS dataset is listed in Table~\ref{ablation}. It is found that the performance is gradually improved as various modules are successively added to the baseline model. So each key module of the proposed model is effective. In order to further prove the effectiveness of each module, we conduct a series of experiments for each component in our model on three  representative datasets, including the commonly used DAVIS dataset, the large-scale VOS dataset and the ViSal dataset which is used only for testing.

\begin{figure}[t]
\small
\centering
\includegraphics[width=0.48\textwidth]{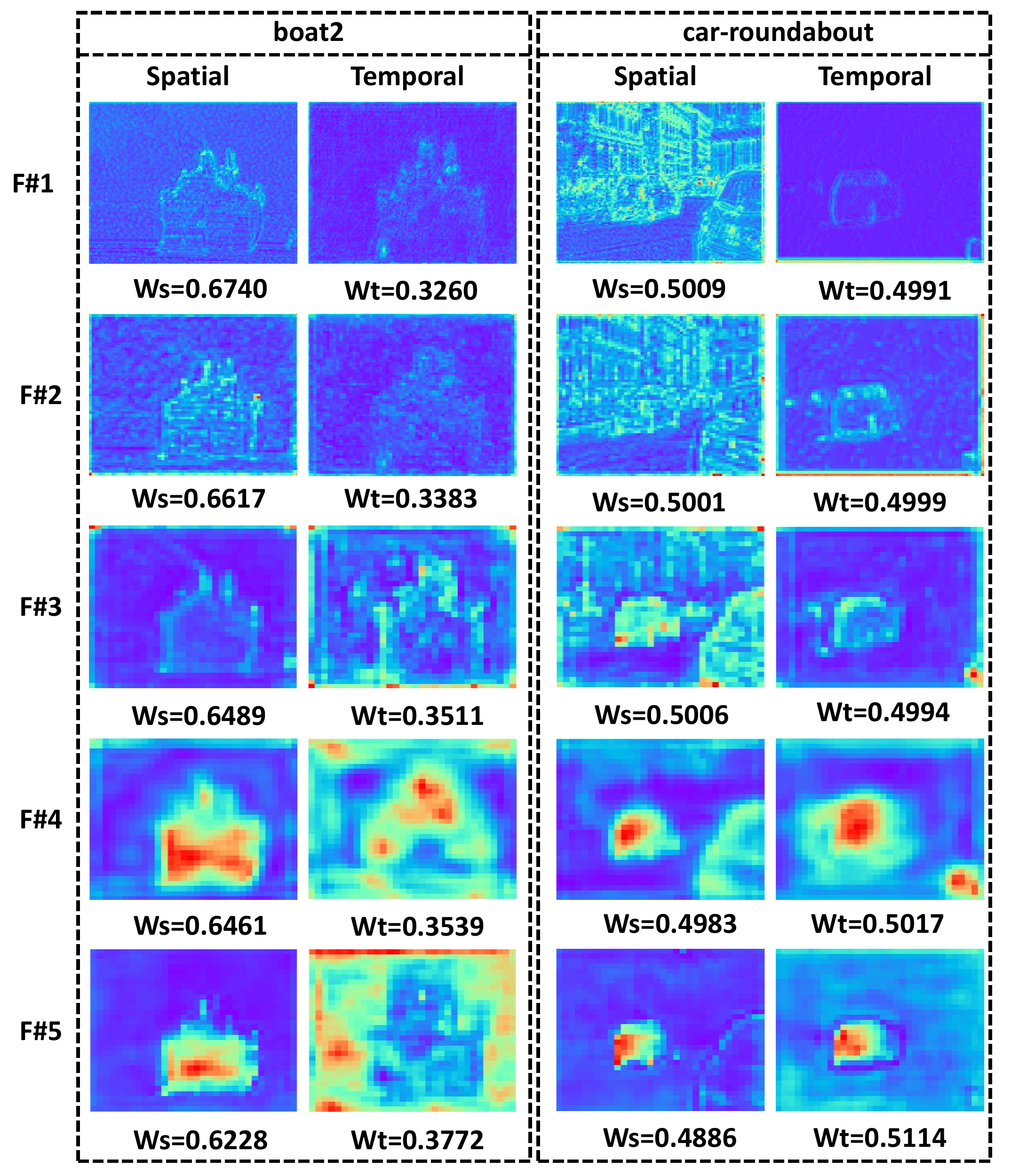}
\caption{Feature maps of different depth of layers and their corresponding weights generated by the proposed DWG. F\#i represents the feature maps of ith layers. Ws and Wt represent the weights of the feature of spatial and temporal branches respectively.}
\label{featuremap}
\end{figure}

\begin{table}[]
\caption{Ablation study of the proposed dynamic weight generator.}
\label{ablation-dwg}
\centering
\begin{tabular}{c|c|c|c|c}
\toprule[.15em]

\multicolumn{2}{c|}{}            & DWG-SEP  & DWG-FC                                  & Proposed                              \\ \midrule[.15em]
                        & maxF $\uparrow$ & 0.881 & 0.877                                 &  \textbf{0.891} \\ \cmidrule{2-5} 
                        & S $\uparrow$    & 0.906 & 0.907                                 &  \textbf{0.914} \\ \cmidrule{2-5} 
\multirow{-3}{*}{DAVIS} & MAE $\downarrow$  & 0.021 & 0.020                                 & \textbf{0.018}                                  \\ \cmidrule{1-5}
                        & maxF $\uparrow$ & 0.932 & 0.932                                 &  \textbf{0.950} \\ \cmidrule{2-5} 
                        & S $\uparrow$    & 0.935 & 0.939                                 &  \textbf{0.949} \\ \cmidrule{2-5} 
\multirow{-3}{*}{ViSal} & MAE $\downarrow$  & 0.019 & 0.018                                 & \textbf{0.013} \\ \cmidrule{1-5}
                        & maxF $\uparrow$ & 0.792 & 0.796                                 &  \textbf{0.801} \\ \cmidrule{2-5} 
                        & S $\uparrow$    & 0.848 & 0.852                                 &  \textbf{0.855} \\ \cmidrule{2-5} 
\multirow{-3}{*}{VOS}   & MAE $\downarrow$  & 0.061 & 0.064                                 &  \textbf{0.060}  \\ \bottomrule[.15em]
\end{tabular}
\end{table}

\subsubsection{Dynamic Weight Generator}
To demonstrate the effectiveness of the proposed Dynamic Weight Generator, we explore two variants of weight generation strategy. The first variant is denoted as DWG-SEP where the feature of each scale is separately processed by global average pooling and convolution to generate a scalar. The second variant is denoted as DWG-FC where the multi-scale features are concatenated but directly passed through a global average pooling layer and a fully connected layer. The performance of these two variants on three datasets are compared with the proposed model in Table~\ref{ablation-dwg}. The effectiveness of the proposed DWG for boosting the performance is clearly validated by the consistent improvement on all three datasets. The proposed DWG achieves an average performance gain of 1.40\% in maxF and 1.20\% in S-measure over DWG-SEP. As for DWG-FC, the proposed structure also brings an average increment of 1.38\% in maxF and 0.73\% in S-measure. DWG-SEP performs even worse than DWG-FC because it fails to consider the correlation among multi-scale features.


As mentioned above, the weight vector generated by DWG is later used for spatiotemporal feature fusion as it is supposed to indicate whether the corresponding features are in favor of the locating of salient objects. In order to validate this point, we illustrate the feature maps of five different scales in the spatial and temporal branches respectively as well as the corresponding weights generated by the proposed DWG module in Fig.~\ref{featuremap}. Two different situations are considered: the features extracted from flow map are quite noisy in `boat2' while the temporal features can greatly reflect the movement of foreground object in `car-roundabout'. The visualization results show that higher weights are assigned to features more highly correlated to the foreground objects. Therefore, the generated weights can effectively guided the feature fusion process by emphasizing on saliency relevant features. Moreover, it is shown that the weighting process works equally well for features from coarse scale to fine scale, hence, guiding the progressive feature fusion in depth.

\begin{table}[]
\centering
\caption{Ablation study of the proposed cross attentive aggregation module.}
\label{ablation-caa}

\resizebox{.95\columnwidth}{!}{
\begin{tabular}{c|c|c|c|c|c}
\toprule[.15em]
\multicolumn{2}{c|}{}         & CAA-woS                                  & CAA-woN                                  & CAA-BU  & Proposed                              \\ \midrule[.15em]
                        & maxF $\uparrow$ & 0.883                                 & 0.877                                 & 0.866 &  \textbf{0.891} \\ \cmidrule{2-6} 
                        & S $\uparrow$    & 0.909                                 & 0.905                                 & 0.898 &  \textbf{0.914} \\ \cmidrule{2-6} 
\multirow{-3}{*}{DAVIS} & MAE $\downarrow$  & 0.019                                 & 0.022                                 & 0.023 & \textbf{0.018}                                \\ \cmidrule{1-6}
                        & maxF $\uparrow$ & 0.949                                 & 0.947                                 & 0.933 &  \textbf{0.950} \\ \cmidrule{2-6} 
                        & S $\uparrow$    & 0.948                                 & 0.947                                 & 0.938 &  \textbf{0.949} \\ \cmidrule{2-6} 
\multirow{-3}{*}{ViSal} & MAE $\downarrow$  & 0.014                                 & 0.016                                 & 0.017 &  \textbf{0.013} \\ \cmidrule{1-6}
                        & maxF $\uparrow$ & 0.794                                 & 0.780                                 & 0.793 &  \textbf{0.801} \\ \cmidrule{2-6} 
                        & S $\uparrow$   & 0.855                                 & 0.843                                 & 0.852 &  \textbf{0.855} \\ \cmidrule{2-6} 
\multirow{-3}{*}{VOS}   & MAE $\downarrow$  & 0.062                                 & 0.061                                 & 0.060 &  \textbf{0.060}                                 \\ \bottomrule[.15em]
\end{tabular}}
\end{table}
\subsubsection{Cross Attentive Aggregation}
To demonstrate the effectiveness of the proposed Cross Attentive Aggregation module, it is compared with three different aggregation methods. The first and second aggregation modules discard the softmax and nonlinear operation of CAA respectively, which is denoted as CAA-woS and CAA-woN. The third aggregation method is denoted as CAA-BU which adopts a bottom-up fusion strategy, i.e., the spatial and temporal features at the same scale are fused and progressively fed into coarser scale. The results of these three variants and the proposed model are shown in Table~\ref{ablation-caa}. The proposed module shows significant performance gain on three datasets by average performance promotion of 0.63\%, 1.53\%, 1.91\% in maxF and 0.22\%, 0.88\%, 1.10\% in S-measure compared with CAA-woS, CAA-woN and CAA-BU respectively. Moreover, we can also find that the nonlinear operation plays a more important role for SOD task compared with softmax operation. We ascribe it to the ability of getting rid of noisy features. And the top-down fusion strategy also brings large performance gain compared with the bottom-up fusion strategy.

\begin{figure}[t]
\centering
\begin{minipage}[b]{0.20\linewidth}
  \centering
  \includegraphics[width=1\linewidth, height=1.5cm]{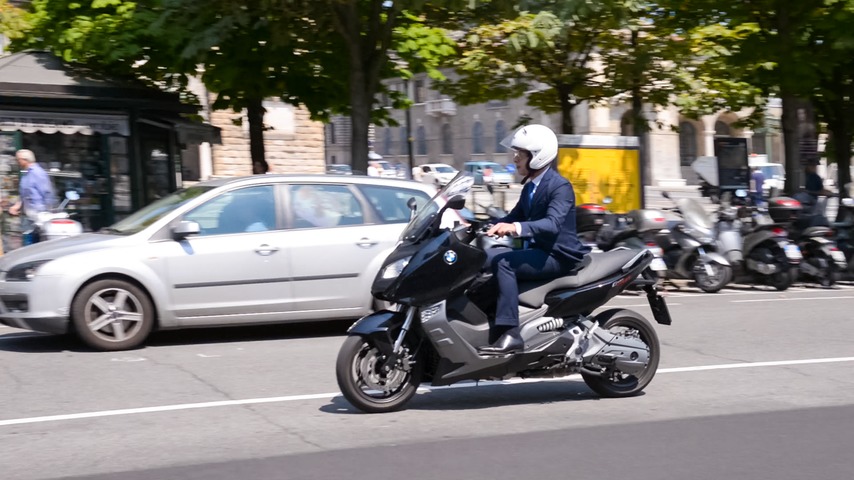}
\end{minipage}
\begin{minipage}[b]{0.20\linewidth}
  \centering
  \includegraphics[width=1\linewidth, height=1.5cm]{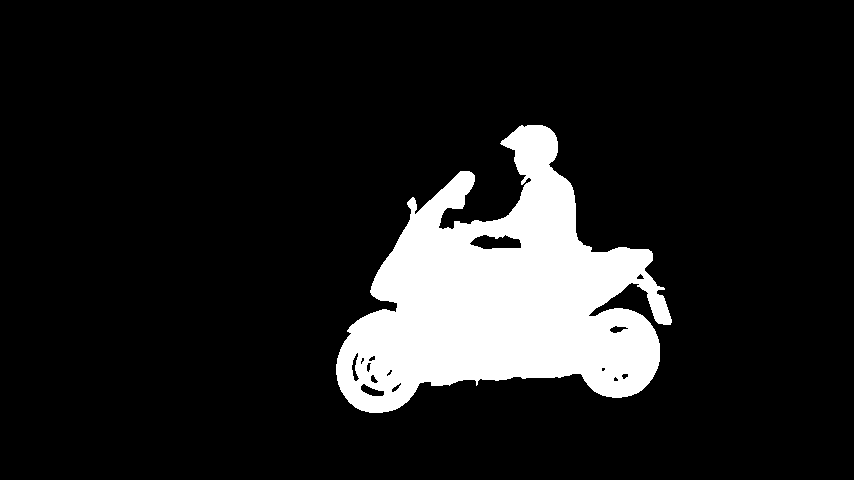}
\end{minipage}
\begin{minipage}[b]{0.20\linewidth}
  \centering
  \includegraphics[width=1\linewidth, height=1.5cm]{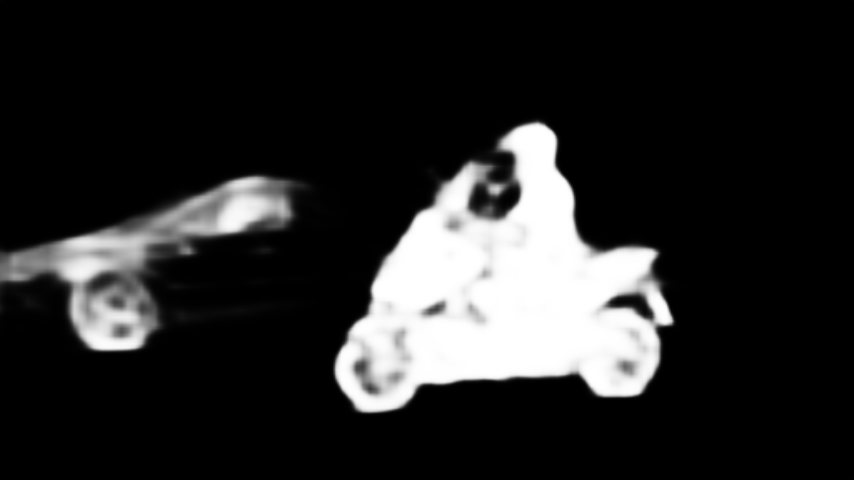}
\end{minipage}
\begin{minipage}[b]{0.20\linewidth}
  \centering
  \includegraphics[width=1\linewidth, height=1.5cm]{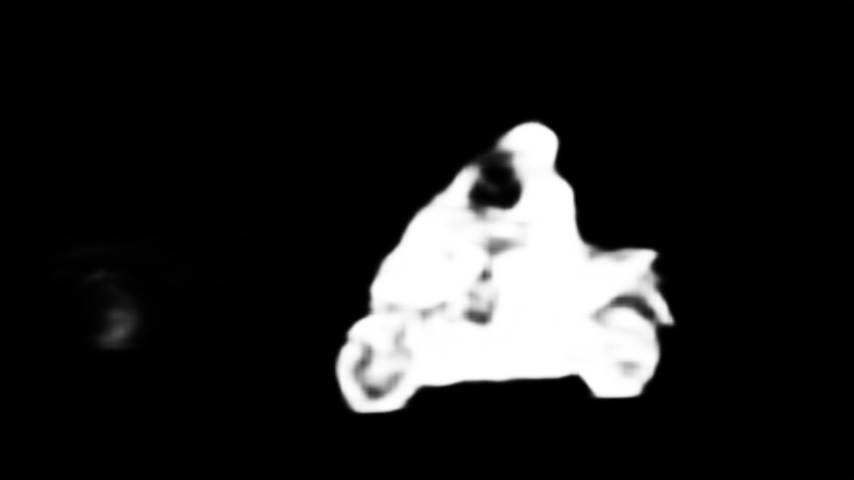}
\end{minipage}
\vspace{3pt}
\\
\begin{minipage}[b]{0.20\linewidth}
  \centering
  \includegraphics[width=1\linewidth, height=1.5cm]{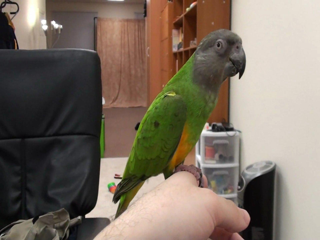}
  \centerline{Frame}
\end{minipage}
\begin{minipage}[b]{0.20\linewidth}
  \centering
  \includegraphics[width=1\linewidth, height=1.5cm]{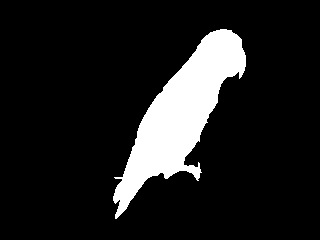}
  \centerline{GT}
\end{minipage}
\begin{minipage}[b]{0.20\linewidth}
  \centering
  \includegraphics[width=1\linewidth, height=1.5cm]{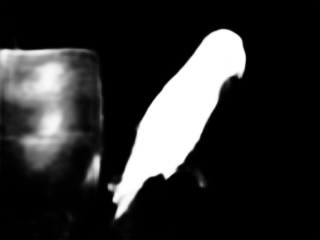}
  \centerline{M-woATT}
\end{minipage}
\begin{minipage}[b]{0.20\linewidth}
  \centering
  \includegraphics[width=1\linewidth, height=1.5cm]{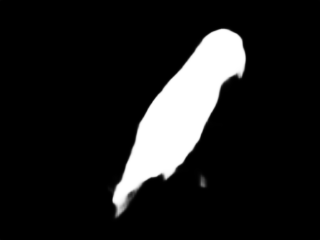}
  \centerline{Proposed}
\end{minipage}
\vspace{3pt}
\\
\caption{Visual comparison of M-woATT and the proposed method.}
\label{compare_woatt}
\end{figure}

\subsubsection{Coarse Saliency Map based Operations}

Note that the coarse saliency maps play a significant role in the proposed method. As analyzed in Section~\ref{coarse saliency}, the multiple supervision and weighted aggregation of coarse saliency maps can further guide the network to generate more effective weights. The fused spatiotemporal feature can also be enhanced by spatial attention based on aggregated coarse saliency map. To evaluate the effectiveness of these operations based on coarse saliency map, we conducted a group of ablation experiments: M-SS represents the model trained in a single supervision manner; M-AGGS denotes a simplified coarse saliency aggregation method which aggregates the spatial and temporal coarse saliency maps by direct addition; M-woATT denotes the model without spatial attention. As shown in Table~\ref{ablation-agg}, the multiple supervised training process outperforms single supervision by 1.02\% in maxF and 0.44\% in S-measure on DAVIS dataset. The weighted aggregation of coarse saliency maps brings an average increment of 1.18\% in maxF and 0.74\% in S-measure, validating its effectiveness. As for the spatial attention, it brings an average performance improvement by 0.76\% in maxF and 0.29\% in S-measure on three datasets. The final saliency map of the proposed DS-Net and M-woATT is shown in 
Fig.~\ref{compare_woatt}. The proposed saliency maps contain less background noise and enjoy higher accuracy, which also demonstrate the effectiveness of the proposed spatial attention.

\begin{table}[]
\caption{Ablation study of the coarse saliency based operations.}
\label{ablation-agg}
\centering
\resizebox{.95\columnwidth}{!}{
\begin{tabular}{c|c|c|c|c|c}
\toprule[.15em]

\multicolumn{2}{c|}{}              & M-SS &M-AGGS   & M-woATT                               & Proposed                              \\ \midrule[.15em]
                        & maxF $\uparrow$  &0.882 &0.877  & 0.884                               &  \textbf{0.891} \\ \cmidrule{2-6} 
                        & S $\uparrow$     &0.910 & 0.905 & 0.911                               &  \textbf{0.914} \\ \cmidrule{2-6} 
\multirow{-3}{*}{DAVIS} & MAE $\downarrow$   &0.022& 0.021  & 0.019                               & \textbf{0.018}                                  \\ \cmidrule{1-6}
                        & maxF $\uparrow$  &      0.950          &  0.946  & 0.943             &  \textbf{0.950} \\ \cmidrule{2-6} 
                        & S $\uparrow$     &                         0.949       & 0.942& 0.945 &  \textbf{0.949} \\ \cmidrule{2-6} 
\multirow{-3}{*}{ViSal} & MAE $\downarrow$   &   0.014    &0.016 & 0.014 &  \textbf{0.013} \\ \cmidrule{1-6}
                        & maxF $\uparrow$  &                   0.794      &  0.789   & 0.795    &  \textbf{0.801} \\ \cmidrule{2-6} 
                        & S $\uparrow$     &                0.854            & 0.851   & 0.854  &  \textbf{0.855} \\ \cmidrule{2-6} 
\multirow{-3}{*}{VOS}   & MAE $\downarrow$   &\textbf{0.056} &   0.063 & 0.061  &                      0.060  \\ \bottomrule[.15em]
\end{tabular}}
\end{table}

\section{Conclusion}
\label{conclusion}
Noticing the high variation on the reliability of static and motion saliency cues for video salient object detection, in this paper, we propose a dynamic spatiotemporal network (DS-Net) to complementarily aggregate spatial and temporal information. Unlike existing methods that treat saliency cues in an indiscriminate way, the proposed algorithm is able to automatically learn the relative reliability of spatial and temporal information and then carry out dynamic fusion of features as well as coarse saliency maps. Specifically, a dynamic weight generator and a top-down cross attentive aggregation approach are designed to obtain dynamic weights and facilitate complementary aggregation. Extensive experimental results on five benchmark datasets show that our DS-Net achieves superior performance than state-of-the-art video salient object detection algorithms.











\bibliographystyle{elsarticle-num}

\bibliography{cas-refs}

\clearpage
\bio{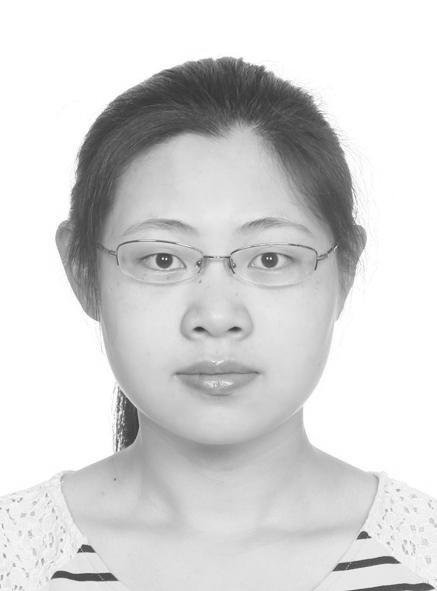}
\textbf{Jing Liu} received the B.E. and Ph.D. degree from Shanghai Jiao Tong University, Shanghai, China, in 2011 and 2017, respectively. She is currently an Associate Professor with the Multimedia Institute of Tianjin University. From 2014 to 2015, she was a visiting student with the Department of Computer Science and Engineering, State University of New York at Buffalo, U.S. Her research interests include multimedia signal processing and perceptual visual processing.

\endbio

\bio{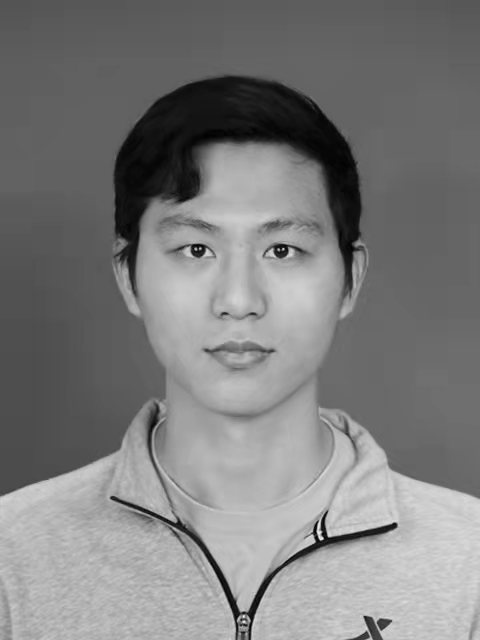}
\textbf{Jiaxiang Wang} is currently pursuing his M.S degree with the School of Information and Communication Engineering, Tianjin University, Tianjin, China. His current research interests are Salient Object Detection and deep learning.
\\
\\
\\
\\
\endbio

\bio{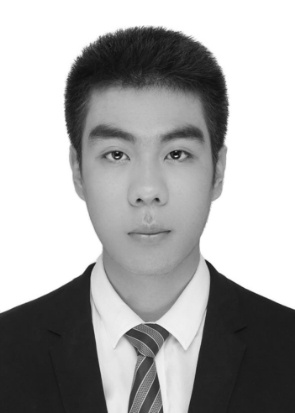}
\textbf{Weikang Wang} is currently pursuing a Ph.D. at the School of Information and Communication Engineering, Tianjin University. His current research interests are salient object detection, video understanding, and deep learning. 
\\
\\
\\
\\
\endbio

\bio{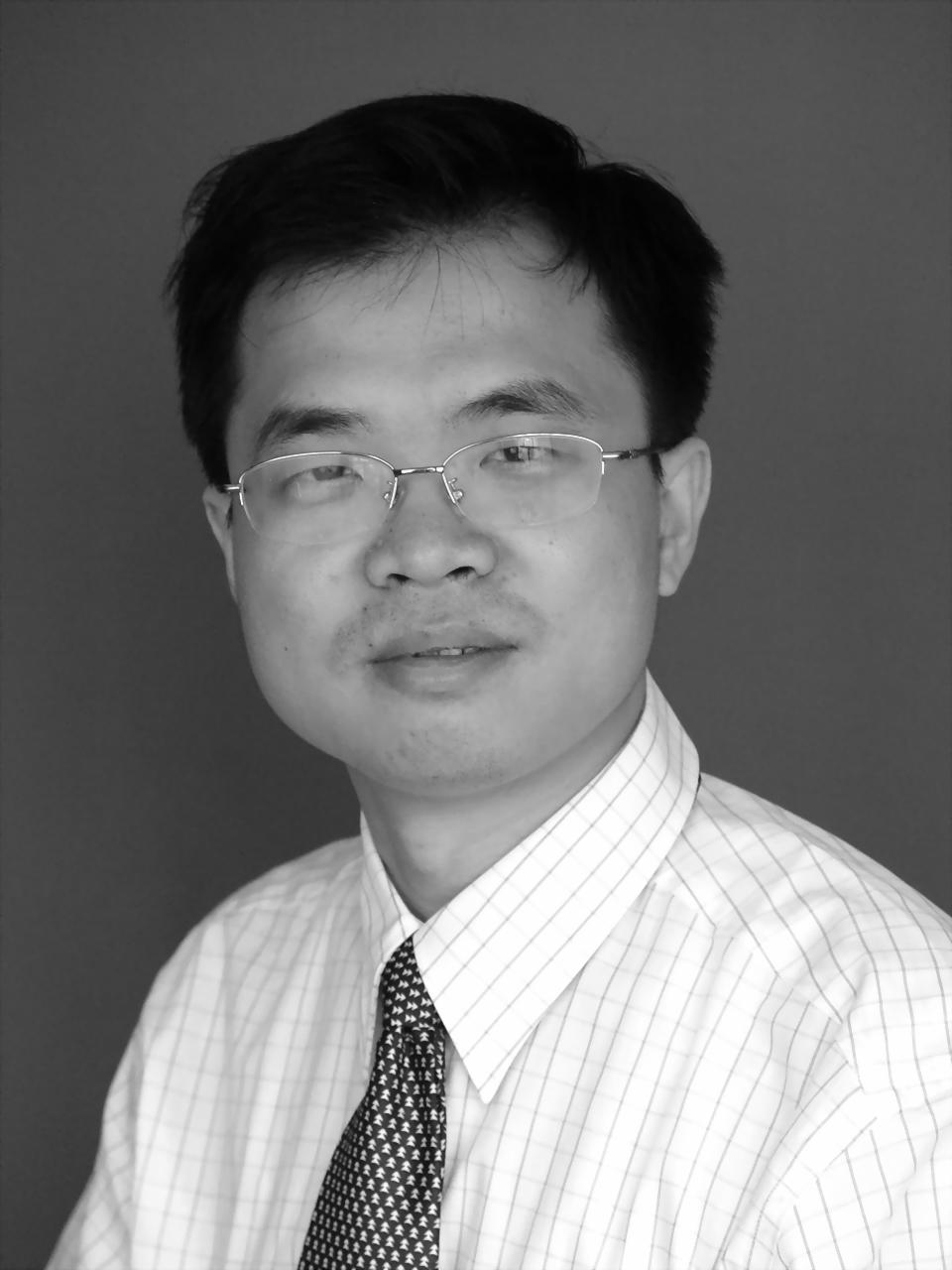}
\textbf{Yuting Su} received the B.S. degree, the M.S. degree and the Ph.D. degree from Tianjin university, Tianjin, China, in 1995, 1998 and 2001 respectively, both in electrical engineering. He is currently a Professor with the School of Electronic Engineering, Tianjin University, Tianjin, China. His research interests include multimedia content analysis and security.
\\

\endbio

\end{document}